%% file: arxiv.tex
\documentclass{fairmeta}

\usepackage{multirow}
\usepackage{tabularx, booktabs} 
\usepackage{amsmath}
\usepackage[normalem]{ulem}
\usepackage{xspace}
\usepackage{pifont}

\input{preamble}

\usepackage{amssymb}
\usepackage{xifthen}
\usepackage{wrapfig}
\usepackage{makecell}
\usepackage[percent]{overpic}
\usepackage{multicol}
\usepackage{hyperref}

\title{Perception Encoder: The best visual embeddings are not at the output of the network}

\author[1,*]{Daniel Bolya}
\author[1,*]{Po-Yao Huang}
\author[1,*]{Peize Sun}
\author[1,2,*,\dagger]{Jang Hyun Cho}
\author[1,*]{Andrea Madotto}
\author[1]{Chen Wei}
\author[1]{Tengyu Ma}
\author[1]{Jiale Zhi}
\author[1]{Jathushan Rajasegaran}
\author[3,\dagger]{Hanoona Rasheed}
\author[4,\dagger]{Junke Wang}
\author[1]{Marco Monteiro}
\author[1]{Hu Xu}
\author[5]{Shiyu Dong}
\author[1]{Nikhila Ravi}
\author[1]{Daniel Li}
\author[1]{Piotr Doll{\'a}r}
\author[1]{Christoph Feichtenhofer}

\affiliation[1]{Meta FAIR}
\affiliation[2]{UT Austin}
\affiliation[3]{MBZUAI}
\affiliation[4]{Fudan University}
\affiliation[5]{Meta Reality Labs}

\contribution[*]{Joint first author}
\contribution[\dagger]{Work done during internships at Meta}

\abstract{
We introduce Perception Encoder (PE), a state-of-the-art vision encoder for image and video understanding trained via simple vision-language learning. 
Traditionally, vision encoders have relied on a variety of pretraining objectives, each tailored to specific downstream tasks such as classification, captioning, or localization. 
Surprisingly, after scaling our carefully tuned image pretraining recipe and refining with our robust video data engine, we find that contrastive vision-language training \textit{alone} can produce strong, general embeddings for all of these downstream tasks. There is only one caveat: \textit{these embeddings are hidden within the intermediate layers of the network}.
To draw them out, we introduce two alignment methods: language alignment for multimodal language modeling, and spatial alignment for dense prediction.
Together, our PE family of models achieves best-in-class results on a wide variety of tasks, including (1) zero-shot image and video classification and retrieval, simultaneously obtaining 86.6 average zero-shot ImageNet robustness and 76.9 zero-shot Kinetics-400 video classification; (2) document, image, and video Q\&A, enabling 94.6 DocVQA, 80.9 InfographicVQA, and 82.7 PerceptionTest with an 8B LLM; and (3) spatial tasks such as detection, tracking, and depth estimation, setting a new COCO state-of-the-art of 66.0 box mAP.
To foster further research, we release our models, code, and novel dataset of synthetically and human-annotated videos.%
}

\metadata[Code]{\url{https://github.com/facebookresearch/perception_models}}
\metadata[Dataset]{\url{https://ai.meta.com/datasets/pe-video/}}

\begin{document}

\maketitle


\vspace{-3pt}
\section{Introduction} \label{sec:intro}
\vspace{-2pt}
For the last decade in computer vision, pretrained vision encoders have been the core building block for most applications requiring \textit{perception}.
From million-scale ImageNet~\cite{imagenet} pretrained convolutional networks~\cite{alexnet,vgg,resnet,efficientnet,convnext} to billion-scale web-pretrained transformers~\cite{vit,align,basic,coca,vit22b,dfn,metaclip,internvl,eva18b}, the dominant strategy in vision has consistently been to adapt large-scale pretrained encoders to downstream tasks.

There are many pretraining objectives today, each with distinct characteristics and each yielding representations better suited for specific tasks:
vision-language contrastive losses~\cite{clip,siglip} learn a global vision and language embedding well-suited for zero-shot classification and retrieval as well as provide vision-language alignment for open-world~\cite{glip,owlv1} and generative tasks~\cite{ldm,dalle2};
captioning losses~\cite{cappa,aimv2} learn to predict image descriptions using a language decoder, which transfers well to downstream multimodal language model (MLLM) tasks;
and spatially self-supervised losses~\cite{mae,dinov2} learn dense spatial correspondences without language supervision, making them useful for tasks requiring precise localization like object detection. 

Many works are now attempting to combine two or more of these techniques in different ways~\cite{coca,aimv2,internvl,eva,eva2,ranzinger2023radio,heinrich2024radio2.5,maninis2024tips}.
While many have been successful, the complexity of these strategies grows exponentially with number of use cases, which can make scaling difficult.
There has not yet been shown a \textit{single, simple, and easily scalable} pretraining technique that can learn state-of-the-art features for all downstream tasks.

\begin{figure}[t!]
    \centering
    \begin{overpic}[width=\linewidth, trim=8.7in 0in 0in 17.48in, clip]{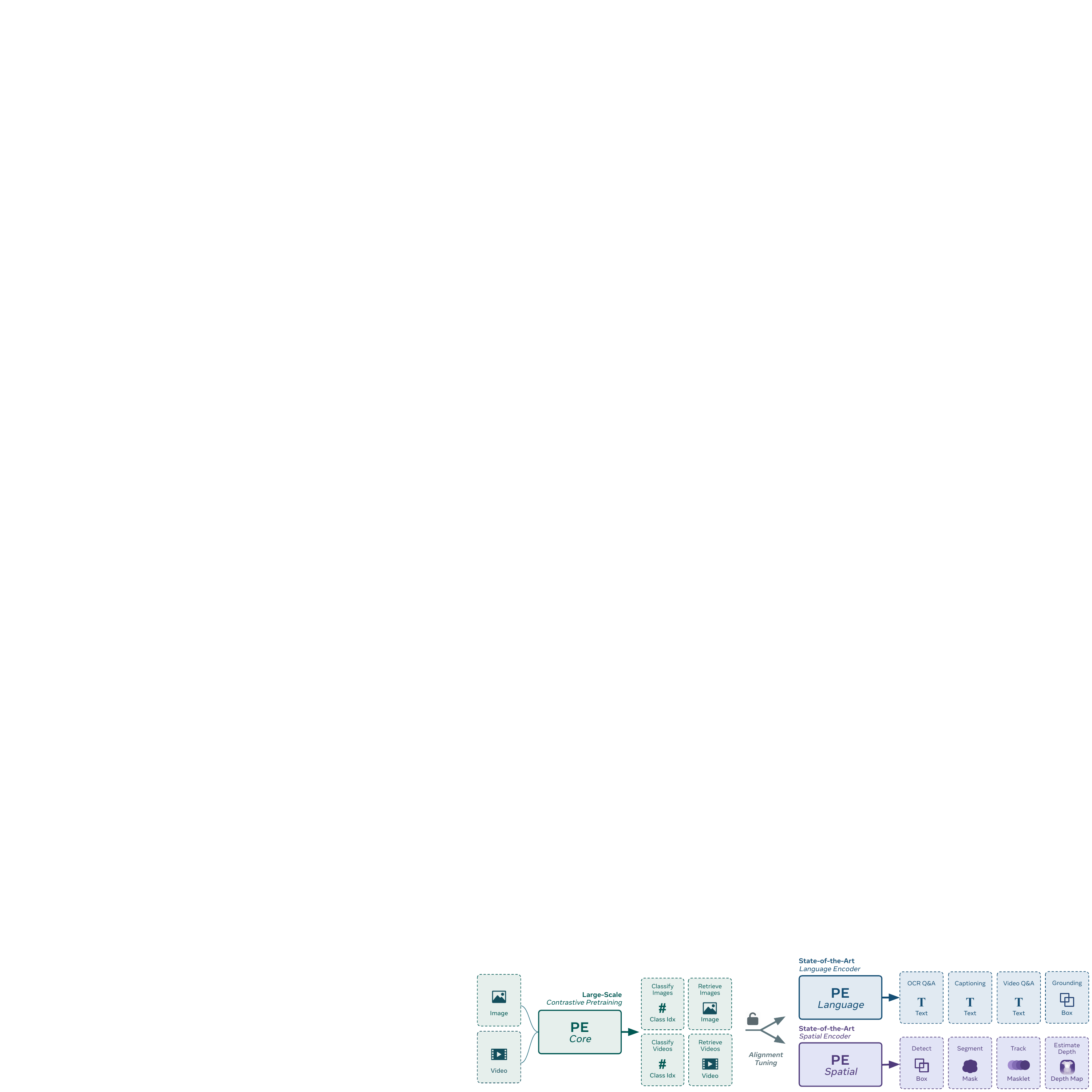}
        \put(21.7,11.75){\fontsize{7}{6.2}\selectfont \hyperref[sec:core]{\bf \textcolor{c1-title-text}{\S}\textcolor{citecolor}{2}}}
        \put(46.2,12){\fontsize{7}{6.2}\selectfont \hyperref[sec:layerfinder]{\bf \textcolor[HTML]{5e757c}{\S}\textcolor{citecolor}{3}}}
        \put(63.9,17.35){\fontsize{7}{6.2}\selectfont \hyperref[sec:la]{\bf \textcolor{c4-title-text}{\S}\textcolor{citecolor}{4}}}
        \put(63.9,6.5){\fontsize{7}{6.2}\selectfont \hyperref[sec:sa]{\bf \textcolor{c6-title-text}{\S}\textcolor{citecolor}{5}}}
    \end{overpic}
    \caption{{\bf Perception Encoder (PE)} is a family of large-scale vision encoder models with state-of-the-art performance on a large variety of vision tasks. 
    By using a robust contrastive pretraining recipe and finetuning on synthetically aligned videos, PE not only outperforms all existing models on classification and retrieval (\S\ref{sec:core}), but it also internally produces strong, general features that \textit{scale} for downstream tasks (\S\ref{sec:layerfinder}).
    PE unlocks the ability for large-scale contrastive pretraining to transfer to downstream tasks with alignment tuning to capitalize on those general features (\S\ref{sec:la}, \S\ref{sec:sa}).
    }
    \label{fig:teaser}
\end{figure}
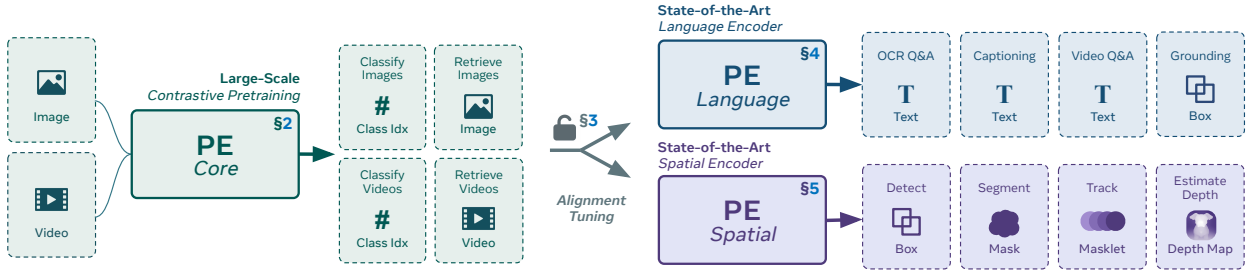

In this work we discover that \textit{global vision-language contrastive learning alone} can be one such approach. After building a state-of-the-art contrastive model for image and video, we found a surprising result: \textit{inside the model were specific features aligned to OCR, VQA, grounding, detection, depth estimation, and tracking}. 
Compared to the state-of-the-art models with captioning~\cite{aimv2} and spatially self-supervised~\cite{dinov2} pretraining, our contrastive encoder has specific layers that, when used as frozen features, matches or exceeds the performance of the other two pretraining techniques \textit{on tasks they should be the best at}. The only problem is---these features exist at \textit{different layers} for each task. By exploiting this phenomenon with \textit{alignment tuning}, we show it is possible to align these features to the end of the network in order to create state-of-the-art encoders for downstream MLLM and spatial tasks---all following the same easily scalable contrastive pretraining.

We begin by building \PEcore{} (Fig.~\ref{fig:teaser}, left), a large-scale contrastively pretrained model with state-of-the-art zero-shot performance on \textit{both} images and video (\S\ref{sec:core}).
To accomplish this, we first focus on developing a strong \textit{image-only} contrastive pretraining recipe to extract general knowledge from billion-scale image-text data.
Keeping the data and training FLOPs fixed, this recipe significantly improves upon vanilla CLIP in both absolute performance and robustness (\S\ref{sec:core_image_pt}).
We then use the resulting model as a frame-based encoder to develop a \textit{video} data engine for generating well-aligned video captions.
Finetuning on this synthetic video-text data substantially improves performance on \textit{both image and video} classification and retrieval tasks (\S\ref{sec:video_data_engine}).
Motivated by this success, we release a large portion of the data used to train the engine: PE Video Dataset (PVD), consisting of 1M diverse videos with 120K human-refined annotations (\S\ref{sec:pvd}).
Finally, we scale our robust image pretraining and well-aligned video finetuning strategy to 2B parameters to produce \PEcore{G} (\S\ref{sec:unified-encoder}), a single unified encoder that outperforms SigLIP2~\cite{siglip2} on zero-shot image tasks and InternVideo2~\cite{internvideo2} on most zero-shot video tasks. We further transfer this power to smaller model scales through distillation.

With the strongest image and video recognition model in hand, we shift our focus to downstream tasks.
Remarkably, despite being pretrained with CLIP loss, we find that the \textit{intermediate layers} of \PEcore{G} can rival AIMv2-3B \cite{aimv2} on language tasks and DINOv2-g \cite{dinov2} on spatial tasks, both of which among the strongest pretrained models in their respective domains. Upon investigation, we attribute this capability to our robust image pretraining strategy, which appears to have unlocked the potential of contrastive pretraining to scale effectively for downstream tasks (\S\ref{sec:layerfinder}). However, a challenge remains: the model does not naturally output these features, keeping them hidden internally. To address this, we introduce two \textit{alignment tuning} methods (Fig.~\ref{fig:teaser}, right) to extract these strong, general features. 

First, in \S\ref{sec:la}, we investigate the most effective technique to align features to the end of the network by adapting to a large language model. This \textit{language alignment} enables us to construct \PElang{G}, which individually outperforms all other popular vision encoders for MLLM tasks. Moreover, when paired with our Perception Language Model (PLM) \cite{PLM}, the combination rivals the latest state-of-the-art MLLMs, like InternVL3 \cite{internvl3}.

Second, in \S\ref{sec:sa}, we identify a dichotomy in the layers optimal for spatial tasks. By visualizing the features and pinpointing the explicit reason for this dichotomy, we develop a straightforward \textit{spatial alignment} approach: distilling \textit{from the model's own frozen features} to achieve most of the alignment, complemented by a novel use of SAM 2 \cite{sam2} for \textit{spatial correspondence} distillation to refine the process. The resulting \PEspat{G} not only outperforms other popular models in depth estimation, tracking, and semantic segmentation, but also sets a new absolute state-of-the-art on COCO \cite{coco} detection with a much simpler decoder.

With this family of checkpoints, Perception Encoder unlocks the potential to scale one simple pretraining method to solve many downstream vision tasks. We are releasing our models, code, and PE Video Dataset.


\section{Perception Encoder: \textit{Core}} \label{sec:core}
\vspace{-5pt}
To build Perception Encoder (PE), we start by training a large-scale, robust, and highly performant vision-language contrastive model for image \textit{and video}.
We have two objectives: first, to enhance the scalability and data efficiency of contrastive training; and second, to create a unified model effective on both image and video.

These goals are somewhat conflicting: image-text data is plentiful and training on images is efficient, but video-text data is scarce and video training is expensive. Thus, we decouple image and video training into two stages. We first develop a strong \textit{image} pretraining recipe (\S\ref{sec:core_image_pt}) with several regularization techniques to create a robust starting point. Then we use the resulting image model as a frame encoder to develop a \textit{video data engine} (\S\ref{sec:video_data_engine}) supported by our novel human-refined video-text dataset (\S\ref{sec:pvd}) to generate aligned captions for video clips. Finally, we finetune the image encoder on the resulting aligned video data (\S\ref{sec:unified-encoder}). Using our data engine design, this short finetuning step substantially improves \textit{both} image and video performance.

\vspace{-5pt}
\subsection{Robust Image Pretraining} \label{sec:core_image_pt}
In the first stage of pretraining, we want to learn as much visual information as possible from a large set of image-text data.
Notably, a unique quirk of contrastive training is the loss for a given sample depends on the other samples in the batch.
Because each batch is different, there is potential to learn new information every time an example is sampled, even if that sample has been seen before.
Thus, we find contrastive learning to benefit from a long training schedule. To exploit this, we design our pretraining recipe with high regularization, stability, and training efficiency in mind.

\begin{wrapfigure}{r}{0.545\textwidth}
\vspace{-27pt}
  \begin{center}
      \begin{tabular}{c}
        \includegraphics[width=1\linewidth, trim = 14.4in 0in 0in 7.6in, clip]{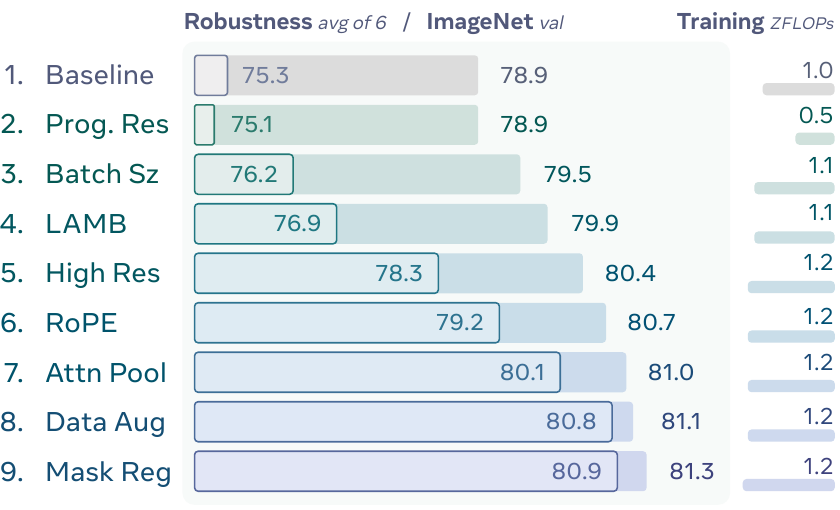}
    \end{tabular}
    \end{center}
    \caption{{\bf Robust Image Pretraining.} We tune our pretraining recipe (\S\ref{sec:core_image_pt}) to maximize performance on a fixed set of data, starting with an OpenCLIP~\cite{openclip} ViT-L/14 model.
    We report cumulative zero-shot classification results for each modification.
    The inner bars show robustness evaluation, calculated as the average of 6 robustness benchmarks~\cite{imagenet,imagenetv2,objectnet,imagenet-a,imagenet-r,imagenet-sketch}, and the outer bars show ImageNet val~\cite{imagenet} alone. Several changes significantly improve robustness, indicating that ImageNet val scales more with data, while robustness can scale with refined training techniques.
    }
    \label{fig:core_pt_ablations}
\vspace{-20pt}
\end{wrapfigure}

\paragraph{Setup.}\,(Fig.~\expandref{fig:core_pt_ablations}{.1})
We track our changes on a vanilla CLIP model using an OpenCLIP~\cite{openclip} ViT-L/14 model at 224 resolution as a baseline. We keep the training budget fixed to around 1T GFLOPs (\emph{i.e.}, a ZFLOP), and train on a fixed 2.3B image-text dataset curated using the MetaCLIP~\cite{metaclip} text-only curation pipeline. For the baseline, we use a global batch size of 32K, class token, AdamW~\cite{adamw}, and train for 12B samples seen.
To assess the \textit{generality} of the information learned during pretraining, we report not only zero-shot ImageNet val~\cite{imagenet} results but also the average performance across a range of robustness metrics, including ImageNet val~\cite{imagenet}, ImageNet v2~\cite{imagenetv2}, ObjectNet~\cite{objectnet}, ImageNet Adversarial~\cite{imagenet-a}, ImageNet Rendition~\cite{imagenet-r}, and ImageNet Sketch~\cite{imagenet-sketch}. As observed with other pure CLIP models~\cite{clip,dfn,metaclip}, the average robustness metric performance of this vanilla recipe is much lower than ImageNet val alone.

\paragraph{Progressive Resolution.}\,(Fig.~\expandref{fig:core_pt_ablations}{.2})
To enable longer training, we first improve training efficiency. As shown in many works~\cite{efficientnet,li2023clipa,swin, touvron2022deit,li2023clipav2}, vision encoders work well with a \textit{progressively increasing} resolution schedule. Thus, we \textit{halve} the training FLOPs while maintaining performance by evenly splitting the baseline 12B-sample run into 98, 154, and 224 resolution stages, with 4B samples per stage.

\paragraph{Increasing Batch Size.}\,(Fig.~\expandref{fig:core_pt_ablations}{.3})
We use the extra budget to double the batch size from 32K to 64K, increasing the total samples seen from 12B to 24B.
Larger batch size means a higher likelihood for there to be a non-trivially novel pair of samples, \emph{i.e.}, hard negatives.
This is akin to increasing the ``task difficulty'' of CLIP and improves ImageNet val by +0.6\% and robustness by double of that, +1.1\%.

\paragraph{LAMB Optimizer.}\,(Fig.~\expandref{fig:core_pt_ablations}{.4})
We switch from AdamW to LAMB~\cite{lamb}, which is known to stabilize large batch training.
More importantly, LAMB allows us to train stably with a higher learning rate of $2$\,$\times$\,10$^{-3}$ compared to the original $5$\,$\times$\,10$^{-4}$. We observe that starting with a high learning rate is important to allow the model to adapt to different resolutions.
These factors combine for +0.4\% on ImageNet val and +0.7\% on robustness.

\paragraph{Increasing Final Resolution.}\,(Fig.~\expandref{fig:core_pt_ablations}{.5})
A classic finding is that parameters and resolution should be scaled together~\cite{efficientnet,feichtenhofer2020x3d}. Thus, we add a fourth 336 resolution stage at the end of training. To keep the training FLOPs the same, we adjust the training schedule to 10B samples at 98 resolution, 8B at 154, 4B at 224, and 2B at 336. While ImageNet val only increases by +0.5\%, robustness improves threefold, rising by +1.4\%.

\paragraph{RoPE.}\,(Fig.~\expandref{fig:core_pt_ablations}{.6})
We add 2D RoPE~\cite{rope} to each attention layer to improve extrapolation, keeping the original position embedding. 2D RoPE only improves ImageNet val by +0.3\% but enhances robustness by +0.9\%.

\paragraph{Attention Pooling.}\,(Fig.~\expandref{fig:core_pt_ablations}{.7})
We follow \cite{siglip} in constructing the CLIP embedding using an attention probing transformer block. Surprisingly, we found keeping the class token as an input to this block is important for small model performance. Together, this improves ImageNet val by +0.3\% and robustness by +0.9\%.

\paragraph{Tuned Data Augmentation.}\,(Fig.~\expandref{fig:core_pt_ablations}{.8})
Despite training on billions of samples, we find data augmentation still important---especially for transfer to unlikely scenarios like in ObjectNet~\cite{objectnet}. We add heavy random cropping, brightness/saturation jitter, and horizontal flip. Random cropping encourages using the entire caption, as not everything is in frame. Jitter helps low-light settings and documents. Horizontal flip improves natural images and does not hurt OCR (see \S\ref{sec:core_results}). These improve robustness by +0.7\%, notably, ObjectNet by +2.4\%. 

\paragraph{Mask Regularization.}\,(Fig.~\expandref{fig:core_pt_ablations}{.9})
As regularization, we want the model to produce the same features if some patches are not visible.
However, passing the CLIP gradients through masked images may negatively alter behavior on unmasked images. Thus, we convert MaskFeat~\cite{maskfeat} into a regularization loss by duplicating and masking 1/16th of the batch. At the output, the masked tokens are aligned to their unmasked counterparts by maximizing cosine similarity. Care is taken to ensure that the CLIP and masked gradients are disjoint.

\begin{figure}[h!]
    \centering
        \includegraphics[width=1.0\linewidth,trim=2.05in 0in 0in 7.55in, clip]{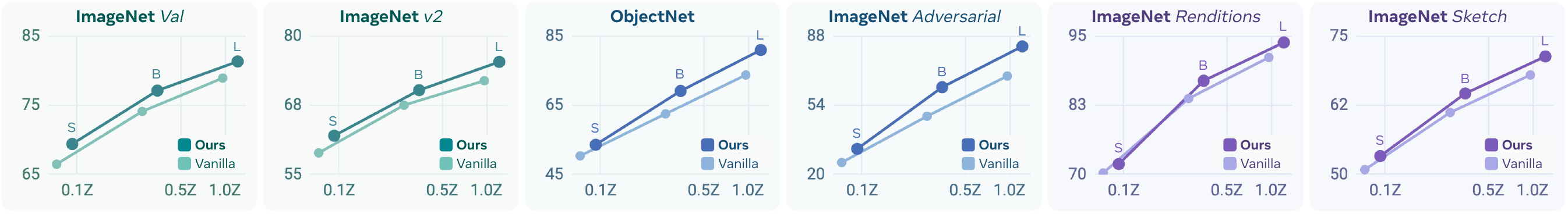}
    \caption{\textbf{Scaling Behavior {(Model Size)}.} Results before and after our recipe changes (Fig.~\ref{fig:core_pt_ablations}) for S/14, B/14, and L/14 models. Our recipe improves scaling for difficult metrics like ObjectNet~\cite{objectnet} and ImageNet Adeversarial~\cite{imagenet-a}. 
    }
    \label{fig:core_pt_scaling}
\end{figure}

\begin{figure}[h!]
    \centering
        \includegraphics[width=1.0\linewidth,trim=2.05in 0in 0in 7.55in, clip]{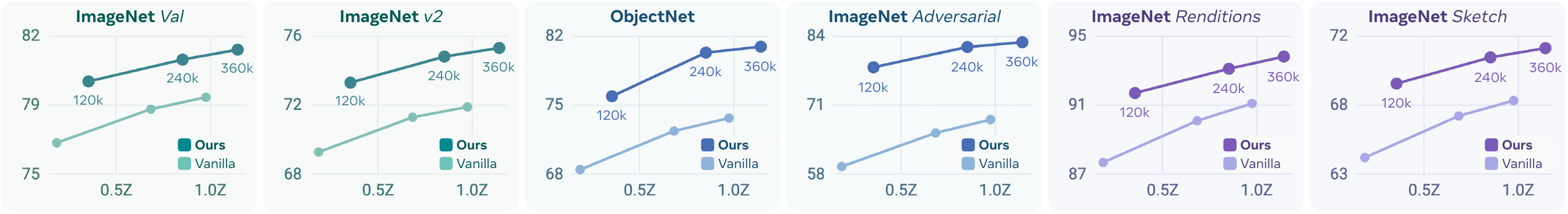}
    \caption{\textbf{Scaling Behavior {(Training Steps)}.} Results before and after our recipe changes for an L/14 model trained with 120K, 240K, and 360K steps, adjusting the learning rate and progressive resolution schedules accordingly. Despite our recipe being much stronger than the original, there is still room for further improvement by training longer. 
    }
    \label{fig:core_pt_scaling2}
\end{figure}

\paragraph{Scaling Behavior.}\,(Figs.~\ref{fig:core_pt_scaling} and \ref{fig:core_pt_scaling2})
In Fig.~\ref{fig:core_pt_scaling}, we show the performance of our recipe (Fig.~\expandref{fig:core_pt_ablations}{.9}) \vs the original CLIP recipe (Fig.~\expandref{fig:core_pt_ablations}{.1}) across S/14, B/14, and L/14 models. For each benchmark, our recipe scales around the same rate or better than the original CLIP recipe. On some difficult datasets like ObjectNet~\cite{objectnet} and ImageNet Adversarial~\cite{imagenet-a}, our recipe shows distinctly better scaling. This indicates that the improvements in performance were not at the cost of scalability, meaning we can further benefit from scaling the model size.

In Fig.~\ref{fig:core_pt_scaling2}, we additionally show the performance of our recipe \vs the original CLIP recipe across L/14 models trained with 120K steps (one-third schedule), 240K steps (two-thirds schedule), and 360K steps (full ablation schedule). All models are their own training runs with full learning rate annealing and the progressive resolution schedule adjusted proportionally. We see nearly linear trends for our recipe on most datasets. This suggests we can train longer for more performance, even at L scale and with 24B samples seen already.

\newpage
\subsection{Bootstrapping a Video Data Engine with Perception Encoder}
\label{sec:video_data_engine}

\begin{wrapfigure}{r}{0.65\textwidth}
\vspace{5pt}
    \centering
    \includegraphics[width=\linewidth, trim=2.1in 0in 0in 5.1in, clip]{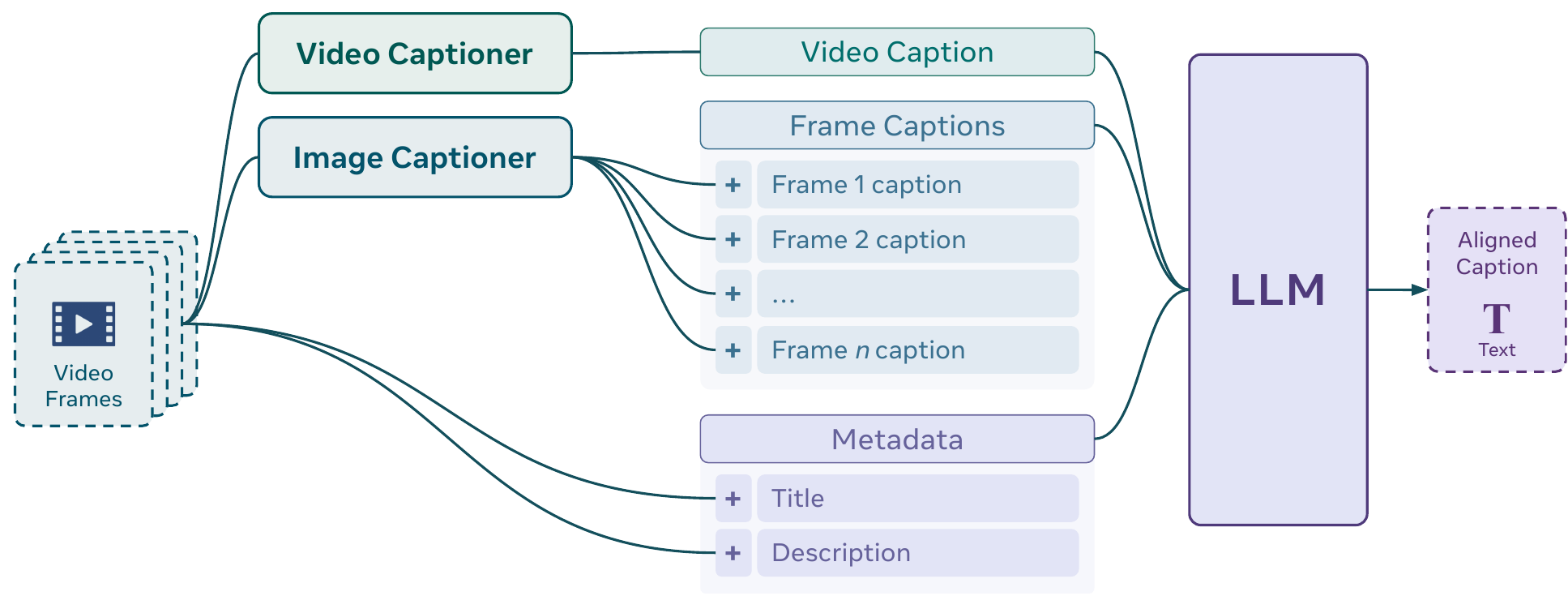}
    \caption{{\bf Video Data Engine.}
    To create aligned video-text data for contrastive training, we use a PE-based video captioner~\cite{PLM} to generate a holistic video caption and an image-level captioner~\cite{llama3} on sampled frames.
    We then provide those captions as well as the original video metadata to text-only LLM~\cite{llama3} to synthesize a single short, aligned caption optimal for contrastive training.
    }
    \label{fig:video_caption_pipeline}
\vspace{-5pt}
\end{wrapfigure}

\label{sec:core_video_ft}
With a robust image pretraining recipe settled and its scaling behavior confirmed, our next step is to extend the image-only encoder to accommodate video and build a unified image-video model.
Unlike web-scale image-text data, which comes in many cases with human-generated descriptive alt-text information, videos with aligned language annotation are inherently scarce. High-quality human-annotated captions for videos are even rarer. This scarcity presents a unique and significant challenge in training encoders capable of effectively processing video inputs.
Inspired by the recent success of image data engines \cite{sam, sam2, veclip, Nguyen2023recap, altogether}, we extend this concept to develop a robust video data engine that generates well-aligned synthetic captions for a diverse set of videos, facilitating the training of a video encoder. This innovative approach represents the first large-scale exploration of its kind. In the following sections, we introduce the process of building our video data engine.

To bootstrap our contrastive video finetuning, we focus on synthesizing video captions. We build our data engine in three stages: (1) we create a strong baseline video captioner, which we call the Perception Language Model (PLM), described in~\cite{PLM}; (2) we add additional high quality video data with human-refined captions to further enhance the captioner's quality; (3) we refine and summarize the generated video captions with an LLM to construct a large video dataset to use for the contrastive video finetuning of our Perception Encoder.

\paragraph{Phase 1: Base Video Captioner (PLM).}
We build our data engine on an early version of PLM~\cite{PLM}, a multimodal large language model with PE as the vision encoder and Llama~\cite{llama3} as the language decoder. 
We train PLM on a large-scale collection of open-access image and video datasets~\cite{PLM}. In total, the training dataset consists of 64.7M images and videos covering natural images, charts, documents, exocentric and egocentric videos. 


\begin{wraptable}{r}{0.5\textwidth}
\centering
\vspace{-13pt}
{
\tablestyle{0pt}{1.05} 
\begin{tabular}{y{90} x{21}x{21}x{25}x{25}x{45}}
    \shline
    & \multicolumn{2}{c}{\ct[c3]{AuroraCap~\cite{auroracap}}} & \multicolumn{2}{c}{\ct[c4]{VCG Diverse~\cite{Maaz2024VideoGPT+}}} & \ct[c5]{VCG Bench~\cite{Maaz2023VideoChatGPT}} \\
    Captioner & \cc[c3]{Score} & \cc[c3]{Acc} & \cc[c4]{Score} & \cc[c4]{Acc} & \cc[c5]{Score} \\
    \hline
    \addpadding
    PLM & {2.2} & 51.9 & 3.1 & 65.1 & 34.3 \\
    PLM + Human-Refined Data & \textbf{3.4} & \textbf{71.1} & \textbf{3.6} & \textbf{79.4} & \textbf{35.2} \\
    \shline
\end{tabular}
}
\caption{{\bf Video Captioning.}
We use an early version of PLM-8B~\cite{PLM}, consisting of our image-only PE encoder and a Llama decoder, for captioning. Adding human-refined data greatly boosts captioning performance (higher is better).
}
\vspace{-15pt}
\label{tab:plm_ablation_for_caption}
\end{wraptable}

\paragraph{Phase 2: PLM\,+\,Refined Data.} 
To further boost captioning performance, we collect a set of 265K videos (105K from PVD which we release, see \S\ref{sec:pvd}), caption them with our base PLM model, and ask human raters to refine the captions\footnote{The annotators are instructed to remove, correct, and add information from the captions.}. We then finetune our base PLM model with this data, significantly improving captioning quality (see Tab.~\ref{tab:plm_ablation_for_caption}).

\paragraph{Phase 3: LLM Summarization.} 
We synthesize the final aligned video captions by incorporating the PLM video captions, Llama 3.2~\cite{llama3} image-only frame captions, and the existing video metadata of video titles and descriptions (Fig.~\ref{fig:video_caption_pipeline}).
Similar to image alt-text, video metadata contains knowledge often not covered by the image and video captioning models. Thus, combining the two leads to more comprehensive captions.
We summarize video captions, frame captions, and video metadata together using the Llama 3.3 70B model to provide the final captions. The prompt used to generate the summary can be found in Appendix~\ref{sec:appx_video_caption}.

\paragraph{Using the Engine.}
Finally, we use the resulting data engine bootstrapped with an image-only checkpoint of PE to generate well-aligned, information-dense captions for a diverse set of 22M videos for contrastive finetuning.

\paragraph{Training with Recaptioned Videos}.
Our goal is to develop a unified image \textit{and} video encoder.
To encode videos using our existing image encoder, we uniformly sample $N$\,$=$\,$8$ frames from video clips and extract frame-level embeddings with the image encoder.
We then apply average pooling over these frame embeddings to obtain video embeddings, which are used for contrastive learning with encoded video captions by the text encoder.
Despite being extremely simple, we find this technique surprisingly effective in producing a strong joint image-video encoder.
We share this finding with previous studies~\cite{clip4clip,internvl}, which note that simple average pooling outperforms more complex pooling strategies like attention-based compression for video.

\begin{wraptable}{r}{0.55\textwidth}
    \vspace{-10pt}
    \centering
    { 
    \tablestyle{0pt}{1.05} 
    \begin{tabular}{x{15}x{15}x{15}x{15}awwwww awwwww}
        \shline
        &&&  & \multicolumn{6}{c}{\ct[c1]{\it Image Zero-Shot}} & \multicolumn{5}{c}{\ct[c3]{\it Video Zero-Shot}} \\
              \cb{Title}{}
            & \cb{Description}{}
            & \cb{Video Caption}{}
            & \cb{Frame Caption}{}
            & \cb[c1]{\textit{\textbf{Average Image}}}{}
            & \cb[c1]{ImageNet}{val~\cite{imagenet}}
            & \cb[c1]{ImageNet}{v2~\cite{imagenetv2}}
            & \cb[c1]{ObjectNet}{IN Classes~\cite{objectnet}}
            & \cb[c2]{MS-COCO}{txt$\rightarrow$img~\cite{coco}}
            & \cb[c2]{MS-COCO}{img$\rightarrow$txt~\cite{coco}}
            & \cb[c3]{\textit{\textbf{Average Video}}}{}
            & \cb[c3]{Kinetics}{400~\cite{kay2017kinetics}}
            & \cb[c3]{Kinetics}{600~\cite{kay2017kinetics}}
            & \cb[c3]{MSR-VTT}{txt$\rightarrow$vid~\cite{vtt}}
            & \cb[c3]{MSR-VTT}{vid$\rightarrow$txt~\cite{vtt}}
            \\
            
        \hline
        &  &           &               
        & \cat{72.6} & 83.3 & 77.8 & 85.8  & 49.4 & 66.8
        & \cat{50.9} & 69.7 & 68.4  & 38.0 & 27.3
        \\
        
        $\checkmark$ & $\checkmark$ &           &               
        & \ca{75.4} & 83.2 & 78.2 & 87.1  & 47.3 & 66.0
        & \ca{56.0} & 74.1 & 73.5  &	39.0 &	37.3
        \\
        
        $\checkmark$ & $\checkmark$ & $\checkmark$ &            
        & \ca{78.2} & 83.5 & 78.4 & 86.8  & 56.0& 74.3
        & \ca{60.9} & 73.8 & 73.4	 & 47.6 & 48.8
        \\

        $\checkmark$ & $\checkmark$ & \hphantom{\textsuperscript{*}}$\checkmark$\textsuperscript{*}  & $\checkmark$  
        & \ca{78.1} & 83.7 & 79.0 & 87.7  & 54.1 & 73.0
        & \ca{60.9} & 75.4 &	75.1 &		46.7 &	46.5
        \\

        $\checkmark$ & $\checkmark$ & $\checkmark$     & $\checkmark$ 
        & \ca{78.2} & 83.7 & 79.0 & 87.5  & 54.6 & 73.2 
        & \ca{61.6} & 75.8 &	75.5 &		47.4 &	48.1
        \\

        \shline
    \end{tabular}
    }
    \caption{{\bf Video Data Engine Ablation.}
    We ablate our video data engine in Fig.~\ref{fig:video_caption_pipeline} by finetuning on an in-development image-only version of PE by averaging the frame embeddings to create a single video CLIP embedding.
    Video captions are generated by PLM trained with or without\textsuperscript{*} human-refined data (see \S\ref{sec:pvd}).
    Frame captions are generated by the Llama 3.2 vision model.
    Each component helps on different metrics, overall culminating in a huge boost to \textit{both} image and video zero-shot performance. 
    }
    \label{tab:video-ft-ablation}
    \vspace{-10pt}
\end{wraptable}

\paragraph{Ablations.}
In Tab.~\ref{tab:video-ft-ablation}, we conduct an ablation study on the components of the video data engine by finetuning an intermediate image-only checkpoint on 17M of the 22M videos recaptioned by our video data engine.
The results show that the video data engine significantly enhances zero-shot classification and retrieval performance for both image and video benchmarks, compared to the image-only baseline encoder (first row). Notably, using the video data engine's video-level and frame-level captions provides significant improvements over relying solely on metadata such as video title and description (second row), highlighting the importance of building a robust video data engine to compensate for noise in web videos.
Our analysis reveals that the most critical components are the video metadata and PLM's video caption; however, all components are necessary to achieve peak performance in our video data engine.

\begin{figure}[b!]
    \centering
    \includegraphics[width=0.98\linewidth, trim=2.04in 0in 0in 14.38in, clip]{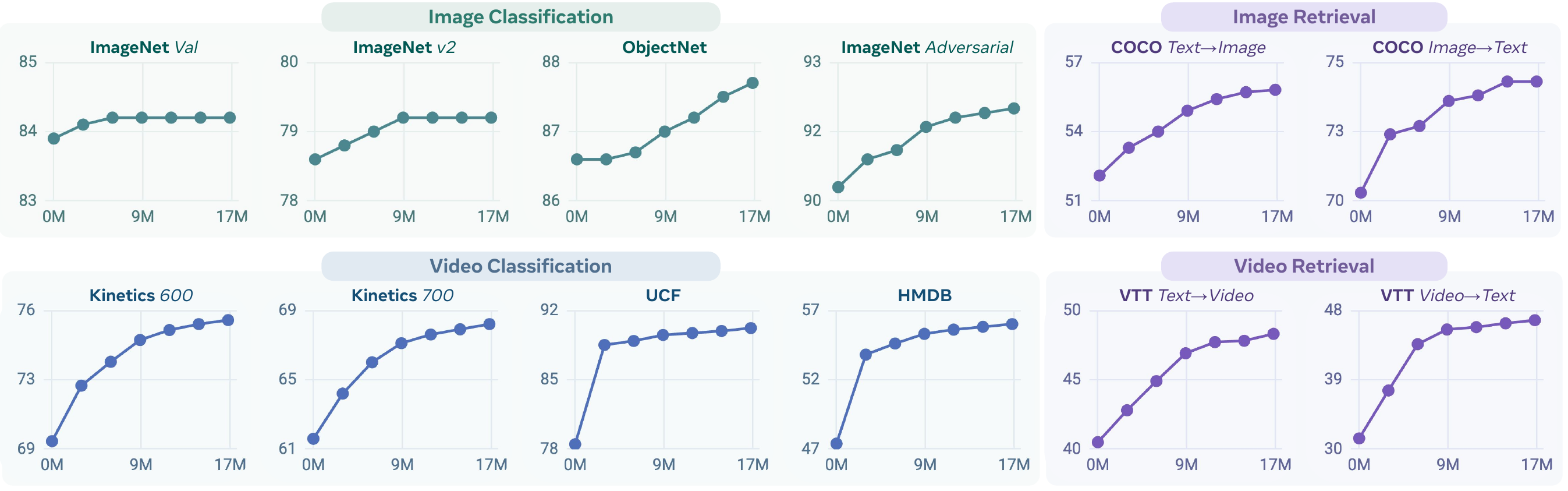}
    \caption{{\bf Video Data Scaling.} 
    Finetuning on videos recaptioned by the PE video data engine from 0M (baseline image-only model) to 17M samples consistently improves both image and video performance, both classification and retrieval.}
    \label{fig:core_video_scaling}
\end{figure}

In Fig.~\ref{fig:core_video_scaling}, we investigate the impact of scaling recaptioned video data 
on a later checkpoint of the same image-only model as in Fig.~\ref{tab:video-ft-ablation}.
Notably, scaling synthetic video data demonstrates consistent improvement in both image and video benchmarks. Full results of this scaling experiment can be found in the Appendix~\ref{tbl_app:video_ft}.

In the top row, scaling synthetic video data consistently improves performance on image benchmarks, with monotonic improvements of +1.1\% in ObjectNet and +1.6\% in ImageNet Adversarial. ImageNet val and ImageNet v2 have smaller gains, with accuracy increases of 0.3\% to 0.5\%, plateauing at $\sim$7M samples. We also observe a significant boost to zero-shot retrieval (here, COCO~\cite{coco}) of +3.8\% to +4.1\% top-1 recall.

The video tasks listed in the bottom row demonstrate a consistent story.
We observe a significant jump in performance between none and 3M videos across all video classification tasks, indicating that there is a domain gap for image-only models that hinders their ability to perform well on video out of the box.
Further scaling synthetic video data leads to substantial performance gains in both video classification and retrieval. Video classification accuracy improves consistently by +5.6\% to +11.7\% without plateauing, while video retrieval shows significant improvements of +7.7 to +15.3 top-1 recall.

These experiments highlight the quality of our video data engine and its ability to significantly improve encoder performance, even with only a relatively modest 17M videos compared to the billions of images seen during pretraining. Our video data engine is a vital component in build a strong, unified image-video encoder.

\subsection{PE Video Dataset (PVD)}
\label{sec:pvd}
For the benefit of the community, we release a new video dataset: PE Video Dataset (PVD).\footnote{PVD available at \url{https://ai.meta.com/datasets/pe-video/}}
PVD comprises of 1M high-quality and diverse videos with accompanying tags and descriptions. The videos are motion-centered, covering both first-person and third-person views with a wide coverage of scenes.

We additionally select 120K of these videos with the highest degree of motion to annotate with detailed captions by generating synthetic captions using our video captioner (\S\ref{sec:video_data_engine}) and employing 200 annotators to verify and refine them. We ask the human annotators to improve the synthetic captions by removing any hallucinations, correcting words that describe the video inaccurately, eliminating repetitive or redundant words to make the caption more concise, and adding any missing actions being performed in the video.

\begin{wraptable}{r}{0.25\textwidth}
\vspace{-10pt}
\centering
{
\tablestyle{4pt}{1.05} 
\begin{tabular}{z{60}y{40}}
    \shline
    \ct[c1]{Videos} & 998,862 \\
    \ct[c2]{Human Captions} & 118,862 \\

    \ct[c3]{Total Duration} & 4625 hrs \\
    \ct[c4]{Duration (s)} & 16.7$\pm$9.8  \\
    \ct[c5]{Human Caption Length} & 57.1$\pm$25.4\\
    \ct[c6]{Model Caption Length} & 111.7$\pm$43.2\\
    \shline
\end{tabular}
}

\captionsetup{justification=centering}
\caption{{\bf PVD Statistics.\label{tab:PEvideo_stat}} 
}
\vspace{-10pt}
\end{wraptable}

We release two versions of annotations for the 120K PVD subset:
(1) Human verified captions: extended summaries with an average length of 57.1 words that provide a high-level description of each video. These captions are suitable for CLIP-style training.
(2) Long automated captions: detailed and fine-grained descriptions with an average length of 111.7 words that capture spatial and temporal events. These captions are ideal for fine-grained video understanding.

\begin{figure*}
    \centering
    \includegraphics[width=\linewidth, trim=7.85in 0in 0in 6.38in, clip]{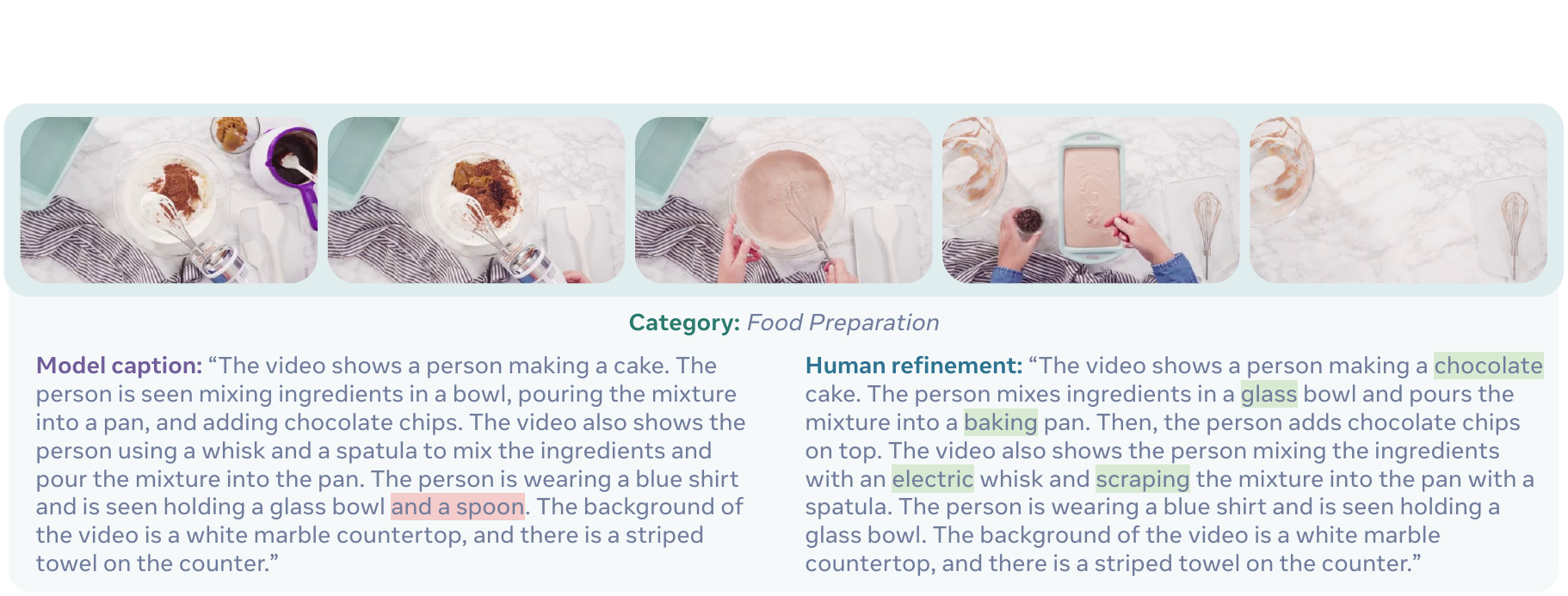}
    \caption{{\bf PE Video Dataset Example.} A sample from PVD, our released video-text dataset. Initial captions are generated by our video captioning model and then refined by human annotators. Annotators are instructed to add details and remove model hallucination. In this example, the model hallucination ``a spoon'' is removed; and more details such as ``glass bowl'' and the action ``scraping'' are added. See Appendix Fig.~\ref{fig:video_data_example_more} for more.}
    \label{fig:video_data_example}
\end{figure*}

In Fig.~\ref{fig:video_data_example}, we visualize a video example together with their model and human captions from PE Video Dataset (See Fig.~\ref{fig:video_data_example_more} for more). The dataset statistics are summarized in Tab.~\ref{tab:PEvideo_stat}. Finally, We use 105K of these refined samples to improve the data engine (\S\ref{sec:video_data_engine} phase 2) and 15K as a high-quality video retrieval benchmark.

\paragraph{PVD Benchmark.}  We use 15K of the human-refined video-caption pairs as a held-out test set, which we introduce as a new video retrieval benchmark, {PVD Benchmark}, to evaluate finegrained video-caption alignment.
We follow the format of MSR-VTT~\cite{vtt} to construct the benchmark. We select videos from 10 different categories, including hand actions, object interactions, food preparation, work activities, outdoor scenes, animals, water scenes, object handling, close-up shots, and nature scenes, with an overall average caption length of 51.7 words (see Appendix~\ref{appx:pvd_bench_distribution} for statistics).
We use PVD Benchmark to evaluate SigLIP~\cite{siglip}, SigLIP2~\cite{siglip2}, InternVL~\cite{internvl}, and PE models, and the results can be found in Tab.~\ref{tab:core_pe_bench}.

\subsection{A Unified Encoder for Image and Video}
\label{sec:unified-encoder}

Using a robust, scalable image pretraining recipe and video-pretraining data recaptioned by the proposed video data engine, in this section we present \textbf{\PEcore{}}, a unified image-and-video encoder.


\begin{wraptable}{r}{0.4\textwidth}
\vspace{-14pt}
\centering
\tablestyle{0pt}{1.05} 
\begin{tabular}{x{20}x{30}x{25}x{20}x{20}x{20}x{20}x{30}}
    \shline
        \ct{Scale} & \ct{Tower} & \ct[c1]{Params} & \ct[c2]{Width} & \ct[c3]{Depth} & \ct[c4]{MLP} & \ct[c5]{Heads} & \ct[c6]{CLIP Dim} \\
        \hline
       \addpadding
        \multirow{2}{*}{B} & Vision & 0.09B & 768 & 12 & 3072 & 12 & \multirow{2}{*}{1024}\addpadding{} \\
                           & Text   & 0.31B & 1024 & 24 & 4096 & 16 &  \\
       \hline
       \addpadding
        \multirow{2}{*}{L} & Vision & 0.32B& 1024 & 24 & 4096 & 16 & \multirow{2}{*}{1024}\addpadding{} \\
                           & Text   & 0.31B & 1024 & 24 & 4096 & 16 &  \\
                           \hline
       \addpadding
        \multirow{2}{*}{G} & Vision & 1.88B & 1536 & 50 & 8960 & 16 & \multirow{2}{*}{1280}\addpadding{} \\
                           & Text   & 0.47B & 1280 & 24 & 5120 & 20 &  \\
    \shline
\end{tabular}
\captionsetup{justification=centering}
\caption{{\bf PE Model Configurations.} } 
\label{tab:pe2b}
\vspace{-20pt}
\end{wraptable}

\paragraph{Model Architecture}. 
To capitalize on the promising scaling behavior observed in \S\ref{sec:core_image_pt}, we scale the largest \PEcore{} model to 2B parameters\footnote{We employ the setup described in \S\ref{sec:core_image_pt} except for the additional class token (only used for L and B). Interestingly, we find \textit{using the same high learning rate} ($2$\,$\times$\,10$^{-3}$) to perform well for G. We also did not find scaling the text encoder to be beneficial. 
} (G scale). Tab.~\ref{tab:pe2b} shows the detailed model configuration of the vision and text transformers and the dimension of the output clip embedding space.

\paragraph{Smaller Model Distillation.}
To maximize the performance of smaller models (B and L scales in Tab.~\ref{tab:pe2b}), we employ a distillation finetuning approach~\cite{distillation} using \PEcore{G} as the teacher. This process involves a short finetuning schedule where both the student and teacher models encode image and text inputs separately to compute image-to-text and text-to-image similarity distributions, similar to CLIP training~\cite{clip}. The student's distributions are then optimized to match those of the teacher by minimizing KL-divergence, distilling multimodal relational knowledge from the teacher into the student.

Notably, we find that using a smaller softmax temperature for the teacher's distributions, specifically 0.5$\times$ the temperature used for the student's distribution, significantly enhances the effectiveness of knowledge distillation. By leveraging the strong embeddings provided by \PEcore{G}, our short distillation finetuning schedule significantly boosts the performance of both B and L scale models of \PEcore{} (see Appendix~\ref{appx:core_smaller_models}).

\paragraph{Model Training}.
The training process of \PEcore{} involves three stages:
\begin{enumerate}
    \item  {\it Image pretraining.} We scale up image pretraining to 5.4B publicly available image alt-text pairs curated with MetaCLIP~\cite{metaclip} and a total of 86B samples seen to ensure convergence (58B for B and L). We use a global batch size of 131K, with progressive resolution from 98 to up to 448 depending on the model.
    \item {\it Image and video finetuning.} Following the initial pretraining, we subsequently finetune the model at max resolution with a short schedule for 50M samples on the image pretraining data (as cooldown) followed by 22M samples on the recaptioned videos with a smaller learning rate and batch size. The video captions are produced using the proposed video data engine (\S\ref{sec:video_data_engine}). For each video clip, we uniformly sample 8 frames, encode them, take their average to produce a single video embedding, and align them with the corresponding video captions using the same contrastive objective in image training.
    \item {\it Smaller model distillation.} We distill the 2B model (G scale) into smaller contrastive pretrained models at B and L scales under their final resolutions, using a short schedule that covers approximately 4B samples seen ($\sim$8\% of the pretraining schedule) with a lower learning rate and no weight decay.
\end{enumerate}
The detailed training configuration and setups are listed in Appendix~\ref{sec:appx_joint_train}.

\begin{table*}[t!]
    \centering
    \makebox[\linewidth][c]{
    \tablestyle{0pt}{1.15} 
    \begin{tabular}{y{55}www awwwwww awwwwwwww awwww}
        \shline
        \multirow{2}{*}{\vspace{-2.2cm} Model}  &&&& \multicolumn{7}{c}{\ct[c1]{\it Zero-Shot Classification}} %
        & \multicolumn{9}{c}{\ct[c2]{\it Zero-Shot Fine-Grained Classification}} %
        & \multicolumn{5}{c}{\ct[c3]{\it Zero-Shot Retrieval}}\\
            & \cb{Encoder Params}{}
            & \cb{Resolution}{}
            & \cb{Data}{}
            & \cb[c1]{\textit{\textbf{Avg Class.}}}{}
            & \cb[c1]{ImageNet}{val~\cite{imagenet}}
            & \cb[c1]{ImageNet}{v2~\cite{imagenetv2}}
            & \cb[c1]{ObjectNet}{IN Classes~\cite{objectnet}}
            & \cb[c1]{ImageNet}{Adversarial~\cite{imagenet-a}}
            & \cb[c1]{ImageNet}{Renditions~\cite{imagenet-r}}
            & \cb[c1]{ImageNet}{Sketch~\cite{imagenet-sketch}}
            & \cb[c2]{\textit{\textbf{Avg Fine.}}}{}
            & \cb[c2]{Food}{101~\cite{food101}}
            & \cb[c2]{Flowers}{Oxford~\cite{flower102}}
            & \cb[c2]{Pets}{Oxford~\cite{pets}}
            & \cb[c2]{Cars}{Stanford~\cite{cars}}
            & \cb[c2]{Aircrafts}{FGVC~\cite{aircraft}}
            & \cb[c2]{Countries}{211~\cite{thomee2016yfcc100m}}
            & \cb[c2]{Scenes}{SUN397~\cite{sun397}}
            & \cb[c2]{Satellite}{RESISC~\cite{cheng_2017_resisc}}
            & \cb[c3]{\textit{\textbf{Avg Retrieval}}}{}
            & \cb[c3]{MS-COCO}{txt$\rightarrow$img~\cite{coco}}
            & \cb[c3]{MS-COCO}{img$\rightarrow$txt~\cite{coco}}
            & \cb[c3]{Flickr-30k}{txt$\rightarrow$img~\cite{flickr}}
            & \cb[c3]{Flickr-30k}{img$\rightarrow$txt~\cite{flickr}}
            \\
        \hline
        \multicolumn{4}{l}{{\textit{Proprietary}}} & \cat{} &&&&&&& \cat{} &&&&&&&&& \cat{} \\
        BASIC~\cite{basic}          & 2.4B    & 224 & 6.6B    & \ca{\textcolor{black}{84.3}} & \textcolor{black}{85.7} & \textcolor{black}{80.6} & \textcolor{black}{82.3} & \textcolor{black}{85.6} & \textcolor{black}{95.7} & {\textcolor{black}{76.1}} %
                & \ca{\textcolor{black}-} & \textcolor{black}{95.1} & \textcolor{black}{91.2} & \textcolor{black}{97.9} &\textcolor{black}- &\textcolor{black}- &\textcolor{black}- & \textcolor{black}{76.2} & \textcolor{black}{72.7}
            & \ca{\textcolor{black}-} & \textcolor{black}- & \textcolor{black}- & \textcolor{black}- & \textcolor{black}-  \\
        CoCa~\cite{coca}            & 1.0B    & 576 & 4.8B    & \ca{\textcolor{black}{85.7}} & {\textcolor{black}{86.3}} & {\textcolor{black}{80.7}} & \textcolor{black}{82.7} & {\textcolor{black}{90.2}} & {\textcolor{black}{96.5}} & {\textcolor{black}{77.6}}  %
                & \ca{\textcolor{black}-} &\textcolor{black}- &\textcolor{black}- &\textcolor{black}- &\textcolor{black}- &\textcolor{black}- &\textcolor{black}- &\textcolor{black}- &\textcolor{black}-
        & \ca{\textcolor{black}{72.6}} & \textcolor{black}{51.2} & \textcolor{black}{66.3} & \textcolor{black}{80.4} & \textcolor{black}{92.5} \\
        LiT-22B~\cite{vit22b}       & 21.7B\,\,\,   & 224 & 15B     & \ca{\textcolor{black}-}    & {\textcolor{black}{85.9}} & {\textcolor{black}{80.9}} & {\textcolor{black}{87.6}} & \textcolor{black}{90.1} & \textcolor{black}{96.0} &  \textcolor{black}-  %
                & \ca{\textcolor{black}-} &\textcolor{black}- &\textcolor{black}- &\textcolor{black}- &\textcolor{black}- &\textcolor{black}- &\textcolor{black}- &\textcolor{black}- &\textcolor{black}- 
            & \ca{\textcolor{black}-} & \textcolor{black}- & \textcolor{black}- & \textcolor{black}- & \textcolor{black}-  \\
        \hline
        
        \multicolumn{4}{l}{{\textit{B Scale}}} & \cat{} &&&&&&& \cat{} &&&&&&&&& \cat{} \\
        SigLIP-B/16$^\dagger$~\cite{siglip}                & 0.1B  & 224 & 10B   
        & \ca{69.9} & 76.2 & 69.5 & 70.7 & 45.1 & 90.2 & 67.9 %
        & \ca{69.5} & 91.6 & 85.2 & 94.2 & 90.8 & 44.0 & 15.9 & 70.0 & 64.6 %
        & \ca{69.8} & 47.2 & 64.5 & 77.9 & 89.6 \\
        SigLIP2-B/16$^\dagger$~\cite{siglip2}      & 0.1B & 224 & 10B   
        & \ca{{73.1}} & {78.2} & {71.4} & \textbf{73.6} & {55.0} & \textbf{91.7} & \textbf{68.9} %
        & \ca{{73.1}} & \textbf{92.8} & {85.7} & \textbf{95.4} & \textbf{93.4} & {54.8} & {19.2} & {72.7} & {71.1} %
        & \ca{{73.7}} & \textbf{52.1} & {68.9} & {80.7} & {93.0} \\
        {\bf \PEcore{B}}                & 0.1B  & 224 & 5.4B    
        & \ca{\textbf{73.2}} & \textbf{78.4} & \textbf{71.7} & {71.9} & \textbf{62.4} & {88.7} & {66.1} %
        & \ca{\textbf{75.0}} & {92.5} & \textbf{86.5} & {94.6} & {92.1} & \textbf{57.0} & \textbf{30.5} & \textbf{74.0} & \textbf{72.7} %
        & \ca{\textbf{74.3}} & {50.9} & \textbf{71.0} & \textbf{80.8} & \textbf{94.4} \\
        \hline
        \multicolumn{4}{l}{{\textit{L Scale}}} & \cat{} &&&&&&& \cat{} &&&&&&&&& \cat{} \\
        SigLIP-L/16$^\dagger$~\cite{siglip} & 0.3B  & 384 & 10B   
        & \ca{80.7} & 82.1 & 75.9 & 80.9 & 76.5 & 95.0 & {73.6} %
        & \ca{74.4} & 95.6 & {89.4} & \textbf{96.8} & {94.8} & 53.2 & 24.7 & 72.5 & 67.9 %
        & \ca{74.7} & 52.8 & 70.5 & 82.6 & 92.9 \\
        SigLIP2-L/16$^\dagger$~\cite{siglip2}   & 0.3B & 384 & 10B
        & \ca{{83.3}} & {83.1} & {77.4} & {84.4} & {84.3} & \textbf{95.7} & \textbf{75.5} %
        & \ca{{78.4}}	& {96.1}	& \textbf{90.0}	& {96.4}	& \textbf{95.8} & {67.0} & {31.6} & {74.8} & {75.5}
        & \ca{{76.7}} & {55.3} & {71.4} & {85.0} & {95.2} \\
        {\bf \PEcore{L}}                & 0.3B  & 336 & 5.4B
        & \ca{\textbf{83.9}} & \textbf{83.5} & \textbf{77.9} & \textbf{84.7} & \textbf{89.0} & {95.2} & 73.4 %
        & \ca{\textbf{80.0}} & \textbf{96.2} & 87.2 & {96.4} & 93.7 & \textbf{67.8} & \textbf{45.6} & \textbf{77.4} & \textbf{75.7} %
        & \ca{\textbf{78.8}} & \textbf{57.1} & \textbf{75.9} & \textbf{85.5} & \textbf{96.6} \\
        \hline
        \multicolumn{4}{l}{{\textit{Unbounded Scale}}} & \cat{} &&&&&&& \cat{} &&&&&&&&& \cat{} \\
        
        DFN-H+$^\dagger$~\cite{dfn} & 0.6B  & 378 & 5B   & \ca{81.6} & 84.3 & 78.3 & 79.6 & 79.6 & 93.6 & 73.3 %
        & \ca{80.5} & 96.2 & \textbf{91.6} & 96.8 & \textbf{96.0} & 72.5 & 37.9 & 77.4 & {75.9} 
        & \ca{75.8} & 55.6 & 71.8 & 82.1 & 93.6  \\
        InternVL-C~\cite{internvl}  & 5.5B    & 224 & 5B      & \ca{82.5} & 83.2 & 77.3 & 80.6 & 83.8 & 95.7 & 74.3 %
            & \ca{76.4} & 95.3 & 85.8 & 96.3 & 94.4 & 53.3 & 35.1 & 76.3 & 74.4
            & \ca{{78.6}} & {\bf 58.6} & {74.9} & {85.0} & 95.7 \\
        EVA 18B~\cite{eva18b}       & 17.5B\,\,\, & 224 & 2B       & \ca{83.6} & 83.8 & 77.9 & 82.2 & 87.3 & 95.7 & 74.7 %
            & \ca{78.8} & 95.8 & 86.0 & 96.1 & {94.9} & 59.7 & 43.1 & {77.7} & \textbf{76.9}
            & \ca{77.5} & 56.2 & 73.6 & 83.3 & {\bf 96.7} \\
        EVA 18B+~\cite{eva18b} & 17.5B\,\,\, & 336 & 2B 
            & \ca{84.1} & 83.9 & 78.2 & 83.6 & 88.9 & 95.6 & 74.3 %
            & \ca{-} & -&- &- &- &- &- &- &-
            & \ca{-} & - & - & - & - \\            
        SigLIP2-g-opt$^\dagger$~\cite{siglip2}      & 1.1B & 384 & 10B      
        & \ca{{86.2}} & 85.0 & {79.8} &  {88.0} & 90.5 & \textbf{96.6} & \textbf{77.4} %
                & \ca{81.0} & \textbf{97.0} & {91.5} & \textbf{97.8} & {95.9} & {73.6} & 40.1 & 76.3 & {75.9}
            & \ca{78.0} & 56.1 & 72.8 & \textbf{86.0} & 95.4 \\
        {\bf \PEcore{G}} {\tiny\it (image only)}                & 1.9B  & 448 & 5.4B    & \ca{86.0} & {85.2} & \textbf{80.2} & 87.1 & {91.2} & 96.1 & 76.1 %
             & \ca{{82.7}} & 96.6 & 91.0 & 96.4 & 94.6 & {76.7} & {57.3} & 77.5 & 71.8
             & \ca{74.9} & 53.1 & 70.9 & 81.6 & 93.9 \\
        {\bf \PEcore{G}}                & 1.9B  & 448 & 5.4B    & \ca{{\bf 86.6}} & \textbf{85.4} & \textbf{80.2} & \textbf{88.2} & \textbf{92.6} & {96.5} & {76.5} %
                & \ca{\bf 83.7} & {96.9} & 91.4 & {96.9} & 94.7 & \textbf{78.2} & \textbf{57.6} & \textbf{78.5} & 75.8
            & \ca{\bf 78.9} & {58.1} & {\bf 75.4} & {85.7} & {96.2} \\
        \shline
    \end{tabular}
    }
    \caption{{\bf Zero-Shot Image Results.} Image zero-shot performance of \PEcore{} compared to the state-of-the-art for \textit{both} proprietary and open models.  \PEcore{G} is the first vision encoder to outperform the best models trained on the proprietary JFT-3B~\cite{vit} and WebLI~\cite{pali} on general classification. 
    Moreover at all model sizes, \PEcore{} obtains state-of-the-art results across general classification, retrieval, and finegrained classification.
    $^\dagger$Re-evaluated: DFN by~\cite{eva18b}; SigLIP and SigLIP2 by us with the same benchmark settings if not reported in~\cite{siglip2} (see Appendix~\ref{appx:zeroshot_settings}).
    }
    \label{tab:core_general_image}
\end{table*}

\subsection{Core Results}
\label{sec:core_results}

\paragraph{Zero-Shot Image Results.}
In Tab.~\ref{tab:core_general_image}, we present \PEcore{}'s performance on zero-shot image benchmarks for classification and retrieval \vs the strongest existing models, including SigLIP2~\cite{siglip2} and proprietary models using JFT-3B~\cite{vit}, which is likely tuned for ImageNet.
\PEcore{} outperforms all other contrastive models across the board on all zero-shot tasks, including the highly competitive average of zero-shot ImageNet robustness metrics~\cite{imagenet,imagenetv2,objectnet,imagenet-a,imagenet-r,imagenet-sketch}.
This marks a significant achievement, as we are the first to accomplish this in over 3 years without access to Google's internal JFT-3B~\cite{vit} or WebLI~\cite{pali} datasets. And \textit{at the same time}, \PEcore{} also exceeds the existing state-of-the-art on image-text retrieval and significantly improves on fine-grained classification---the first to simultaneously hold state-of-the-art on all common zero-shot categories.

By harnessing the power of our video data engine, training with a relatively small dataset of 22M videos and their corresponding synthetic captions leads to substantial \textit{gains in image benchmarks}, with average general image classification improving by +0.6\% with emphasis on more difficult benchmarks (notably +1.2\% ObjectNet, +1.4\% ImageNet Adversarial) and fine-grained classification by +1.0\% on average. Furthermore, due to the high level of detail and alignment of our synthetic captions, zero-shot retrieval is significantly boosted by +3.6\% on average.
These results emphasize that training with well-aligned video text data does not just improve video performance---it creates a strictly better model for both videos \textit{and} images.

\begin{wraptable}{r}{0.7\textwidth}
    \vspace{-10pt}
    \centering
    \makebox[\linewidth][c]{
    \tablestyle{0pt}{1.2} 
    \begin{tabular}{y{51}wx{11}x{9}x{16} awwwww awwwwww}
        \shline
        \multirow{2}{*}{\vspace{-2.2cm} Model}  &&&&& \multicolumn{6}{c}{\ct[c4]{\it Zero-Shot Classification}} %
        & \multicolumn{7}{c}{\ct[c5]{\it Zero-Shot Retrieval}}\\
            & \cb{Encoder Params}{}
            & \cb{Resolution}{}
            & \cb{\# Frames}{}
            & \cb{Video Data}{}
            & \cb[c4]{\textit{\textbf{Avg Class.}}}{}
            & \cb[c4]{Kinetics}{400~\cite{kay2017kinetics}}
            & \cb[c4]{Kinetics}{600~\cite{kay2017kinetics}}
            & \cb[c4]{Kinetics}{700~\cite{kay2017kinetics}}
            & \cb[c4]{UCF}{101~\cite{soomro2012ucf101}}
            & \cb[c4]{HMDB}{51~\cite{kuehne2011hmdb}}
            & \cb[c5]{\textit{\textbf{Avg Retrieval}}}{}
            & \cb[c5]{MSR-VTT}{txt$\rightarrow$video~\cite{coco}}
            & \cb[c5]{MSR-VTT}{video$\rightarrow$txt~\cite{coco}}
            & \cb[c5]{MSVD}{txt$\rightarrow$video~\cite{flickr}}
            & \cb[c5]{MSVD}{video$\rightarrow$txt~\cite{flickr}}
            & \cb[c5]{ActivityNet}{txt$\rightarrow$video~\cite{flickr}}
            & \cb[c5]{ActivityNet}{video$\rightarrow$txt~\cite{flickr}}
            \\
        \hline
        \multicolumn{5}{l}{{\textit{B Scale}}} & \cat{} &&&&&& \cat{} \\      
        CLIP~\cite{clip} & 0.1B  & 224 & 8 & n/a   & \ca{54.3} & 58.4 & {55.1} & 46.1 & 68.9 & 43.2 %
            & \ca{29.2} & 30.4 & 24.2 & 40.5 & 57.2  & 9.1 & 13.2 \\
        CLIP4CLIP~\cite{clip4clip} & 0.1B  & 224 & 12 & n/a   & \ca{-} & - & - & - & - & -  %
            & \ca{-} & 32.0 & - & 38.5 & -  & - & -  \\
        SigLIP2-B/16$^\dagger$~\cite{siglip2} & 0.1B  & 224 & 8 & n/a   
        & \ca{{{57.3}}} & {58.7} & 55.0 & {48.4}  & {82.0} & {42.3} %
            & \ca{{{39.9}}} & {38.5} & {30.1} & {49.0} & {67.2} & {28.6} & {25.8}  \\
        {\bf \PEcore{B}}                & 0.1B  & 224 & 8 & 22M    & \ca{{\textbf{63.9}}} & \textbf{65.6} & \textbf{65.1} & \textbf{55.8}  & \textbf{84.6} & \textbf{48.2} %
            & \ca{{\textbf{49.9}}} & \textbf{47.6} & \textbf{47.3} & \textbf{50.4} & \textbf{76.7} & \textbf{39.0} & \textbf{38.4}  \\
        \hline
        \multicolumn{5}{l}{{\textit{L Scale}}} & \cat{} &&&&&& \cat{} \\

        UMT-L~\cite{umt} & 0.3B  & 224 & 8 & 25M   & \ca{-} & - & - & - & - & -  %
            & \ca{{47.1}} & 40.7 & {37.1} & 49.0 & {74.5}  & {41.9} & {39.4}  \\
        
        SigLIP2-L/16$^\dagger$~\cite{siglip2} & 0.3B  & 384 & 8 & n/a   
        & \ca{{64.1}} & {65.3} & {62.5} & {56.8} & {86.7} & {49.3} %
        & \ca{44.7} & {41.5} & 31.4 & {53.7} & 74.2 & 35.9 & 31.5    \\
        
        {\bf \PEcore{L}}                & 0.3B  & 336 & 8 & 22M    & \ca{\textbf{71.4}} & \textbf{73.4} & \textbf{72.7}  & \textbf{65.3}  & \textbf{87.1} & \textbf{58.5} %
            & \ca{\textbf{54.8}} & \textbf{50.3} & \textbf{50.1} & \textbf{57.2} & \textbf{82.4} & \textbf{46.4} & \textbf{42.1}  \\
        \hline
        \multicolumn{5}{l}{{\textit{Unbounded Scale}}} & \cat{} &&&&&& \cat{} \\

        InternVL~\cite{internvl} & 5.5B  & 224 & 8 & n/a   & \ca{-} & 69.1 & 68.9 & 60.6 & - & -  %
            & \ca{-} & 44.7 & 40.2 & - & -  & - & -  \\

        InternVideo2 ~\cite{internvideo2} & 1.0B  & 224 & 8 & 102M   & \ca{70.7} & 73.1 & {72.8} & {64.9} & 88.8 & 53.9  %
            & \ca{\textbf{59.9}} & \textbf{51.9} & {50.9} & {58.1} & {83.3}  & \textbf{60.4} & \textbf{54.8}  \\
        VideoPrism-g\textsuperscript{*}~\cite{videoprism} & 1.1B  & 288 & 16 & 619M   & \ca{-} & {76.4} & - & - & -& -  %
            & \ca{-} & 39.7 & \textbf{71.0} & - & -  & 52.7 & 50.3  \\

        SigLIP2-g-opt$^\dagger$~\cite{siglip2} & 1.1B  & 384 & 8 & n/a  
        & \ca{68.2} & 69.8 & 67.0 & 61.8 & 90.7 & 51.8 %
        & \ca{46.6} & 43.1 & 34.2 & 55.8 & 74.6 & 38.3 & 33.4    \\

        {\bf \PEcore{G}} {\tiny\it (image only)}                  & 1.9B  & 448 & 8 & n/a    & \ca{{70.9}} & 73.1 & 72.2 & 64.3  & {89.5} & {55.5} %
            & \ca{47.6} & 44.3 & 35.2 & 54.3 & 73.9 & 41.4 & 36.3  \\
        
        {\bf \PEcore{G}}                & 1.9B  & 448 & 8 & 22M    & \ca{\textbf{74.8}} & \textbf{76.9} & \textbf{76.1} & \textbf{69.1}  & \textbf{90.7} & \textbf{61.1} %
            & \ca{{58.7}} & {51.2} & 49.9 & \textbf{59.7} & \textbf{85.4} & {54.7} & {51.2}  \\
        \shline
    \end{tabular}
    }
    \caption{{\bf Zero-Shot Video Results.} Video performance of \PEcore{} compared to recent video and image encoders. \PEcore{} obtains state-of-the-art in video classification and comparable performance on retrieval benchmarks while using only 22M videos. \textsuperscript{*}Proprietary models.
    $^\dagger$SigLIP2 are evaluated by us with the same zero-shot prompts frame embedding averaging strategy (as in~\cite{clip, internvl, clip4clip}). See Appendix~\ref{appx:zeroshot_settings}.
    }
    \label{tab:core_general_video}
    \vspace{-20pt}
\end{wraptable}

\paragraph{Zero-Shot Video Results.}
We assess the performance of \PEcore{} on zero-shot video benchmarks by employing the same model as a frame-based video encoder, utilizing 8 uniformly sampled frames, as described in \S\ref{sec:video_data_engine}.

We present the corresponding video results in Tab.~\ref{tab:core_general_video}. Our base image encoder already outperforms all other image-only encoders on both zero-shot classification and retrieval, including SigLIP2-g-opt. With video finetuning, \PEcore{}G significantly outperforms even native video models that use full temporal attention on video classification, and nearly matches the state-of-the-art on video retrieval using a simple frame-level encoder. This result underscores the importance of our video data engine, resulting in +3.9\% on average zero-shot video classification, and a massive +11.1\% on retrieval. Moreover, \PEcore{} does this with much less video data compared to other video-based approaches like InternVideo2~\cite{internvideo2} and VideoPrism~\cite{videoprism}, highlighting the benefits of a joint image-video encoder.

\paragraph{Additional Zero-Shot Benchmarks.}
We further evaluate \PEcore{} on an additional set of zero-shot classification and retrieval benchmarks we construct in Tab.~\ref{tab:core_pe_bench} to address key gaps in common benchmarks. For comparison, we also evaluate SigLIP2~\cite{siglip2} and InternVL-C~\cite{internvl} on these benchmarks.

First, we note that the version of ObjectNet~\cite{objectnet} that is standard to benchmark robustness (e.g., in Tab.~\ref{tab:core_general_image}) is \textit{not} the full set. ObjectNet consists of 313 classes of objects in challenging and uncommon orientations, locations, and viewpoints. However, the standard version used for benchmarking is a 113 class subset of classes that overlap with ImageNet-1k~\cite{imagenet}. Naturally, benchmarking in this way rewards performing well on ImageNet classes over generality. To remove this bias, we construct the full ObjectNet set with all classes and compare to the reduced ObjectNet set in Tab.~\ref{tab:core_pe_bench}. Surprisingly, we find that while \PEcore{G} performs +7.6\% over InternVL-C and only +0.2\% over SigLIP2-g-opt on the reduced ObjectNet set, it performs +11.8\% over InternVL-C and +0.9\% over SigLIP2-g-opt on the full set of classes, highlighting PE's generality.

Next, we include iNaturalist~\cite{inat2017} as a \textit{zero-shot} benchmark because of its level of specificity with 2,101 fine-grained long-tail classes.
\PEcore{G} outperforms the next best SigLIP2-g-opt model by \textit{+9.6\%}, emphasizing PE's long tail knowledge. We then evaluate PE's cultural diversity on Dollar Street~\cite{dollar_st}\footnote{We use the version provided by \cite{datacomp} and re-evaluate all models to ensure a fair comparison.}, which consists of images of under-represented populations. Here too we find \PEcore{G} to outperform existing methods, with +3.0\% over SigLIP2-g-opt. 
Further, we test OCR performance by setting up TextCaps~\cite{sidorov2020textcaps} as a retrieval dataset. 
Notably, \PEcore{} performs on par or better than SigLIP, which is known for good OCR performance. 
This is potentially surprising, as the horizontal flip augmentation we used during robust pretraining (\S\ref{sec:core_image_pt}) is typically thought to hurt OCR performance. However, instead it seems to have given \PEcore{} the ability to read backwards: we test the same TextCaps retrieval but with all images horizontally flipped. Other models suffer from this, but \PEcore{G}'s performance only drops by 0.1\%.
Finally, we evaluate \PEcore{G} on the PVD benchmark (\S\ref{sec:pvd}), a challenging video retrieval task on 15K 
diverse and human-refined videos. Here, \PEcore{G} significantly outperforms InternVL~\cite{internvl} by +13.6\% on text$\rightarrow$video and +9.5\% to SigLIP2~\cite{siglip2} on video$\rightarrow$text.

\paragraph{Frozen Encoder Probing Results.}
To compare against models that are not capable of zero-shot classification, we additionally evaluate \PEcore{} using k nearest neighbors (following~\cite{dinov2}), linear probing (following~\cite{internvl}), and attention probing (following~\cite{aimv2}) on top of the ImageNet-1k~\cite{imagenet} train set. We present these results in Tab.~\ref{tab:core_frozen_features} and compare to other encoders using their reported numbers. In every case, \PEcore{G} outperforms all existing open encoders, including those with significantly more parameters.

\paragraph{Summary.} \PEcore{}, a unified image-video encoder, achieves state-of-the-art performance across zero-shot classification and retrieval on both images and videos on a wide variety of benchmarks. 
This synergy is made possible by our robust image pretraining recipe (\S\ref{sec:core_image_pt}) and powerful video data engine (\S\ref{sec:core_video_ft}), which together enable the model to effectively leverage the strengths of both image and video data at scale.

\begin{table}[t!]
    \centering
    \begin{minipage}[t]{0.58\textwidth}
    \vspace{0pt} 
    \centering
    \tablestyle{0pt}{1.2} 
    \begin{tabular}{y{60}www wwww wwww}
        \shline
        \multirow{2}{*}{\vspace{-2.2cm} Model}  &&&& \multicolumn{4}{c}{\ct[c1]{\it Zero-Shot Classification}} %
        & \multicolumn{4}{c}{\ct[c4]{\it Zero-Shot Retrieval}}\\
            & \cb{Encoder Params}{}
            & \cb{Resolution}{}
            & \cb{Data}{}
            & \cb[c1]{ObjectNet~\cite{objectnet}}{IN Overlap (113)}
            & \cb[c1]{ObjectNet~\cite{objectnet}}{All Classes (313)}
            & \cb[c1]{iNaturalist}{2017~\cite{inat2017}}
            & \cb[c1]{Dollar St}{58~\cite{dollar_st, datacomp}}
            & \cb[c4]{TextCaps}{img$\rightarrow$txt~\cite{sidorov2020textcaps}}
            & \cb[c4]{TextCaps {\tiny \bf Flip}}{img$\rightarrow$txt~\cite{sidorov2020textcaps}}
            & \cb[c4]{PVD Bench}{text$\rightarrow$vid}
            & \cb[c4]{PVD Bench}{vid$\rightarrow$txt}  
            \\
        \hline
        \addpadding
        {SigLIP2-B/16~\cite{siglip2}} & 0.1B  & 224 & 10B & \textbf{73.6} & \textbf{59.1} & {16.9}	& \textbf{55.9} & {72.0} & {69.8} & {53.9} & {60.1} \\
        {\bf \PEcore{B}} & 0.1B  & 224 & 5.4B & {71.9} & {58.3} & \textbf{25.9} & {52.1} & \textbf{72.3} & \textbf{71.9} & \textbf{59.8} & \textbf{61.1} \\
        \hline
        \addpadding
        {SigLIP2-L/16~\cite{siglip2}} & 0.3B  & 384 & 10B & {84.4} & {73.2}
        & {26.7} & {57.6} & {78.0} & {76.2} & {61.9} & \textbf{67.1} \\
        {\bf \PEcore{L}} & 0.3B  & 336 & 5.4B & \textbf{84.7} & \textbf{74.3} & \textbf{35.3}	& \textbf{59.6} & \textbf{78.5} & \textbf{78.3} & \textbf{64.7} & {65.2} \\
        \hline
        \addpadding
        {InternVL-C~\cite{internvl}} & 5.5B & 224 & 5B & 80.6 & 67.2 & 19.4 & 58.2 &  72.3 & 67.8 & {63.4} & 65.1 \\
        {SigLIP2-g-opt~\cite{siglip2}} & 1.1B  & 384 & 10B & {88.0} & {78.1} & {31.5} &  {59.3} & \textbf{78.8} & {76.9} & {62.5} & {67.1} \\
        {\bf \PEcore{G}} & 1.9B  & 448 & 5.4B & \textbf{88.2} & \textbf{79.0} & \textbf{41.1}	& \textbf{62.3} & \textbf{78.8} 
        & \textbf{78.7} & \textbf{77.0} & \textbf{76.6} \\
        \shline
    \end{tabular}
    \caption{{\bf Additional Zero-Shot Results.} We present several additional zero-shot benchmarks from existing datasets and our own PVD (\S\ref{sec:pvd}) to address evaluation gaps left by standard benchmarks.
    }
    \label{tab:core_pe_bench}
    \end{minipage}
    \hfill
    \begin{minipage}[t]{0.36\textwidth}
    \vspace{0pt} 
    {
        \centering
        \tablestyle{0pt}{1.2} 
        \begin{tabular}{y{60}www www}
        \shline
        \multirow{2}{*}{\vspace{-2.2cm} Model} &&&& \multicolumn{3}{c}{\ct[c1]{\it Encoder Probing}} \\ %
            & \cb{Encoder Params}{}
            & \cb{Resolution}{}
            & \cb{Data}{}
            & \cb[c1]{ImageNet~\cite{imagenet}}{KNN}
            & \cb[c1]{ImageNet~\cite{imagenet}}{Linear}
            & \cb[c1]{ImageNet~\cite{imagenet}}{Attention}  \\    
            \hline
            \addpadding
            DINOv2-g~\cite{dinov2} & 1.1B & 224 & 145M & 83.5 & 86.5 & \hphantom{$^\dagger$}87.2$^\dagger$ \\
            RADIOv2.5-g~\cite{heinrich2024radio2.5} & 1.1B & 518 & - & {85.3} & - & - \\
            AIMv2 3B~\cite{aimv2} & 2.7B & 448 & 7.2B& -  & - & {89.5} \\
            InternVL-C~\cite{internvl} & 5.5B & 224 & 5B & - & 88.2  & - \\
            EVA 18B~\cite{eva18b} & 17.5B\,\,\, & 224 & 2B & - & {88.9} & - \\
            {\bf \PEcore{G}} & 1.9B & 448 & 5.4B & {\bf 86.8}  & {\textbf{89.5}} & {\bf 89.8} \\
            \shline
        \end{tabular}
    }
    \caption{\textbf{Encoder Probing Results.} We evaluate \PEcore{G}'s frozen features using the typical probing methods to compare to models without zero-shot support. $^\dagger$from \cite{aimv2}.}
    \label{tab:core_frozen_features}
    \end{minipage}
\end{table}

\clearpage

\section{General Features in a Contrastive Disguise} \label{sec:layerfinder}

\PEcore{} puts up strong results on the tasks contrastive encoders are known for, like zero-shot classification and retrieval. But while those tasks are useful, they are only a small part of the vision ecosystem. What \textit{really matters} is whether or not the features learned with our pretraining recipe are useful to downstream tasks.

Today's common wisdom in the vision community cites that different pretraining methods result in features useful for different tasks: e.g., contrastive for classification, captioning for language modeling, and self-supervised learning for spatial tasks. To see how \PEcore{} stacks up against against models with different pretraining techniques, we compare its \textit{frozen features} to the state-of-the-art large-scale models for captioning (AIMv2-3B~\cite{aimv2}) and self-supervised learning (DINOv2-g~\cite{dinov2}) on a variety of downstream tasks.

\begin{wrapfigure}{r}{0.6\textwidth}
\vspace{-10pt}
    \centering
    \includegraphics[width=\linewidth, trim = 12.7in 0in 0in 9.8in, clip]{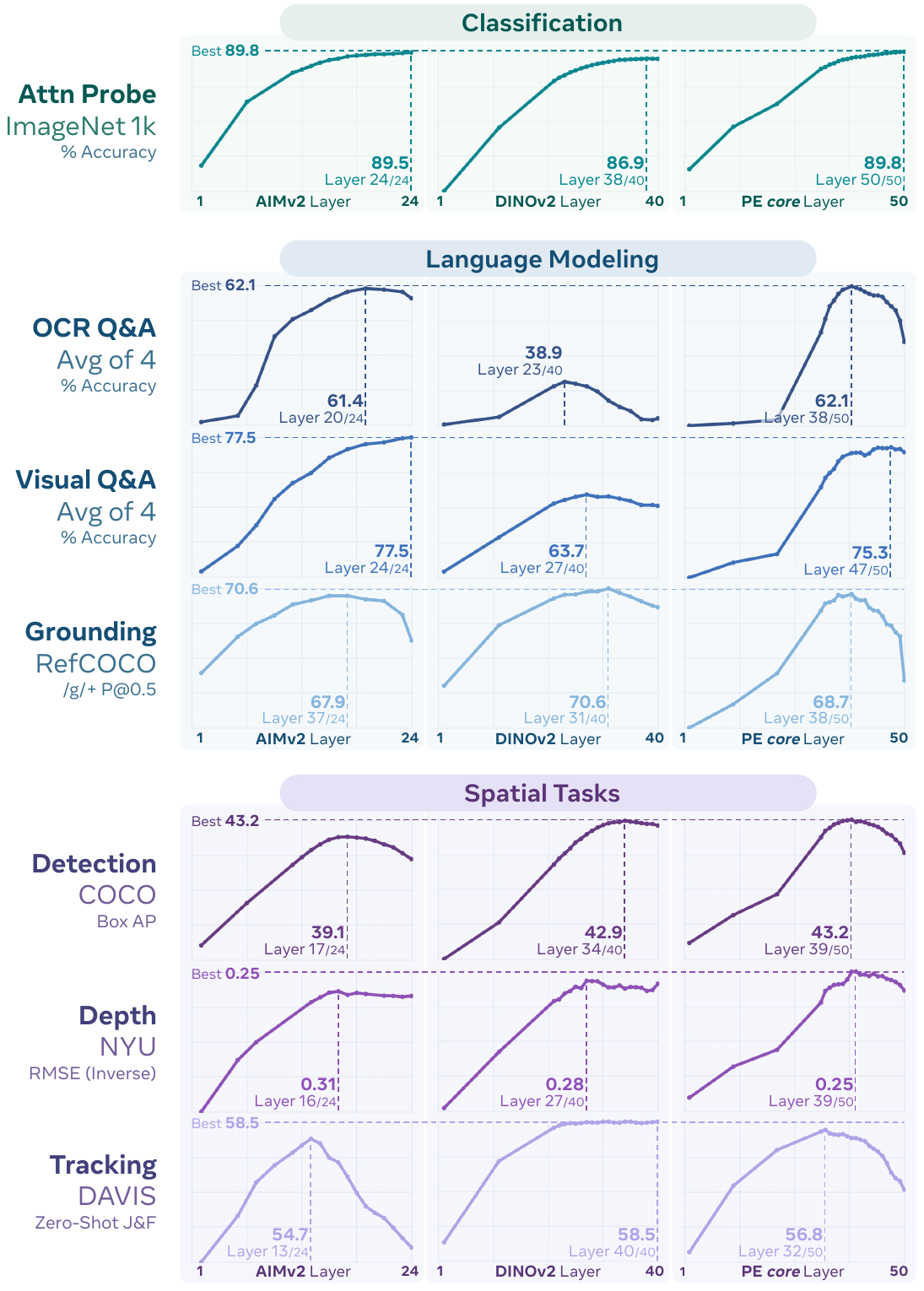}
    \caption{
        {\bf Layer Analysis.} Evaluating intermediate layers as frozen features across tasks for different pretraining methods: captioning (AIMv2-3B~\cite{aimv2}, left), spatially self-supervised (DINOv2-g~\cite{dinov2}, middle), and our contrastive recipe (\PEcore{G}, right). Vertical lines denote the best layer and horizontal lines the best performance across models. As expected, AIMv2 performs well on language but not spatial, and DINOv2 performs well on spatial but not language. 
        But surprisingly, \textit{intermediate layers} of \PEcore{G} perform well on \textit{both} language modeling and spatial tasks. 
    }
    \label{fig:layerfinder}
\vspace{-20pt}
\end{wrapfigure}

\paragraph{Layerwise Feature Analysis.} %
We summarize the results of our frozen feature analysis in Fig.~\ref{fig:layerfinder} for several downstream benchmarks in 3 categories: classification, language modeling, and spatial tasks. For classification, we probe each model using a randomly initialized cross attention transformer block. For language alignment, we use the Perception Language Model (PLM) \cite{PLM} frozen encoder evaluation setup, learning a projector and finetuning a decoder-only LLM (see \S\ref{sec:la}), and for spatial tasks we train with several different decoders (ViTDet~\cite{vitdet} Mask-RCNN~\cite{maskrcnn} with Absolute Win~\cite{abswin} for detection, DPT~\cite{dpt} for depth, and zero-shot feature correspondance for tracking~\cite{jabri2020space}). For each experiment, we sweep over the layers of the model as the optimal features are not necessarily the last~\cite{chen2024internvit2p5}. In each case, we use an equivalent image size (window size for detection) of $32\times32$ tokens. In each plot, we normalize performance by the maximum and minimum performance across models on that task.

\paragraph{An Alignment Problem.} %
This analysis reveals several insights.
First, as expected, AIMv2 performs well at classification and the best at visual Q\&A language tasks.
Similarly, DINOv2 performs the well on spatial tasks like detection, depth, and even performs the best at grounding through an LLM.
Then as already established by other works: 
DINOv2 lacks performance on OCR tasks~\cite{cambrian}.
This is no secret, but what is interesting is that its performance \textit{peaks in the middle of the network} and then drops significantly by the end. And so does the performance of other models for other downstream tasks (AIMv2: tracking, grounding, detection; DINOv2: VQ\&A, grounding).

\PEcore{} exhibits similar behavior, but with unexpected results.
Unlike the others, in earlier layers of the network \PEcore{} \textit{performs well on all tasks, often matching or exceeding the leading models}.
Remarkably, PE has intermediate layers that perform near to or on par with AIMv2 for language tasks and DINOv2 for spatial tasks, despite being trained with contrastive loss.
Depth estimation is particularly noteworthy, as contrastive encoders are not typically considered state-of-the-art in that area.

However, in almost all cases this strong performance \textit{diminishes rapidly} towards the end of the network. In fact, the performance of \PEcore{} in the final layer is \textit{abysmal} for certain tasks, such as LLM-based grounding (the reason for which will become apparent in \S\ref{sec:sa}). This behavior is less pronounced the closer the downstream task is to the pretraining method, suggesting an \textit{alignment problem}. Specifically, a well-tuned large-scale contrastive model can learn general embeddings in the process of fitting its objective, \textit{but it fails to output them}. Therefore, to reveal these embeddings, the model must be subsequentially aligned to downstream tasks.

\paragraph{Analysis.}
The finding that pure CLIP models possess features which match the performance of state-of-the-art pretraining methods in their specialized domains is new. In fact, recent work \cite{webdino} has shown the opposite---that CLIP models fail to scale on downstream tasks. We next investigate how our approach yields these results.

\begin{wrapfigure}{r}{0.545\textwidth}
\vspace{-10pt}
  \begin{center}
      \begin{tabular}{c}
        \includegraphics[width=1\linewidth, trim = 14.4in 0in 0in 7.6in, clip]{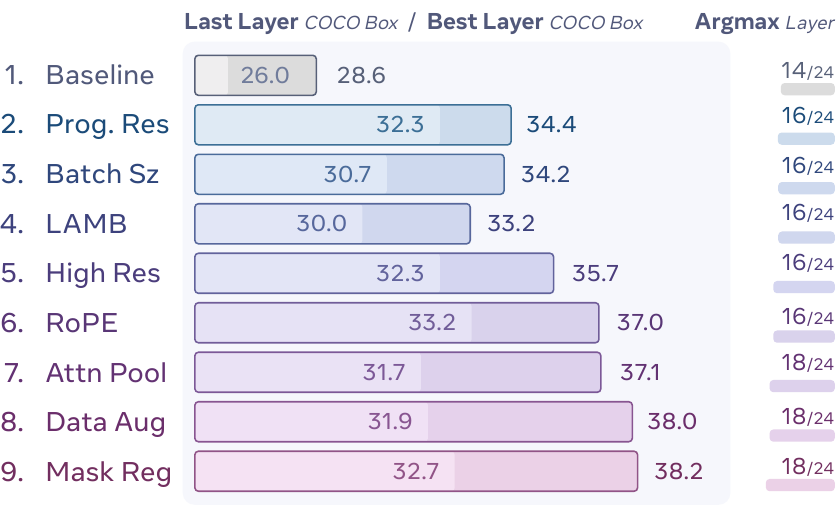}
    \end{tabular}
    \end{center}
    \caption{{\bf The Downstream Effects of Robust Pretraining.} The ViT-L/14 checkpoints from Fig.~\ref{fig:core_pt_ablations} evaluated as frozen features on COCO~\cite{coco} using Mask R-CNN~\cite{maskrcnn}. We report the last layer performance, best layer performance, and the best layer's index.
    }
    \label{fig:layerfinder_ablations}
\vspace{-10pt}
\end{wrapfigure}

\begin{wrapfigure}{l}{0.3\textwidth}
\vspace{-15pt}
  \begin{center}
      \begin{tabular}{c}
        \includegraphics[width=1\linewidth, trim = 16.5in 0in 0in 8.6in, clip]{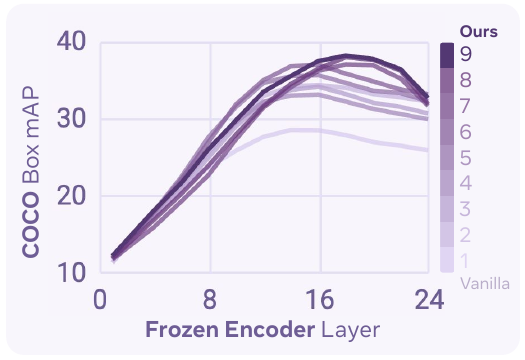}
    \end{tabular}
    \end{center}
    \caption{{\bf Layer Analysis} corresponding to the results presented in Fig.~\ref{fig:layerfinder_ablations}.}
    \label{fig:layerfinder_ablations_plot}
\vspace{-10pt}
\end{wrapfigure}

To start, we perform layerwise frozen feature analysis on COCO detection. \PEcore{} was particularly ``peaky'' on this task in Fig.~\ref{fig:layerfinder}, with its best layer on par with DINOv2, but last layer significantly worse. We already ablated each change we made from vanilla CLIP in Fig.~\ref{fig:core_pt_ablations} using a ViT-L/14 model. So to retrace our steps, we run frozen feature analysis on those checkpoints. For efficiency, we perform this experiment at a lower resolution and only sample even layers. In Fig.~\ref{fig:layerfinder_ablations}, we report COCO box mAP for the last and best layers for each cumulative ablation, along with the index of the best layer. Further, we plot the layerwise performance for each change in Fig.~\ref{fig:layerfinder_ablations_plot}.

Surprisingly, the simple changes we made in \S\ref{sec:core_image_pt} to construct our pretraining recipe overall improved the best layer's performance by \textit{almost 10 mAP} over vanilla CLIP! Some changes like high resolution (5) and RoPE (6) improving spatial features is to be expected, but unexpectedly data augmentation (8) and \textit{especially} progressive resolution (2) help considerably. It is possible that contrastive pretraining is prone to overfit to the ``global'' nature of the task through ``global tokens''~\cite{vitsneedregisters}. However, as the model cannot maintain global tokens in the same place due to the resolution progressively changing, it is forced to be more robust. Also of note is that both progressive resolution (2) and attention pooling (7) move the argmax layer deeper into the network (rightmost column of Fig.~\ref{fig:layerfinder_ablations}). Attention pooling in particular alters the whole shape of the layerwise performance curve (Fig.~\ref{fig:layerfinder_ablations_plot}), while the other changes typically only raise or lower it.

Potentially more interesting is what did not improve performance: specifically, increasing the batch size (3) and using LAMB with a high learning rate (4). Both of these changes explicitly help the model fit the CLIP loss better, which after a certain point may not improve the general features. Moreover, while the best layer overall improved significantly, the last layer performance stagnated after (2). This suggests that constructing the global CLIP token requires a substantial ``decoder'' (in this case, 6 layers for the final L/14 model). Although the features of this decoder are beneficial for some tasks (e.g., Visual Q\&A as shown in Fig.~\ref{fig:layerfinder}), they are not general. Nevertheless, this does not prevent the model from learning general features; it merely limits their expression in the output.

\begin{wrapfigure}{r}{0.6\textwidth}
\vspace{-14pt}
  \begin{center}
      \begin{tabular}{c}
        \includegraphics[width=1\linewidth, trim = 10.65in 0in 0in 17.1in, clip]{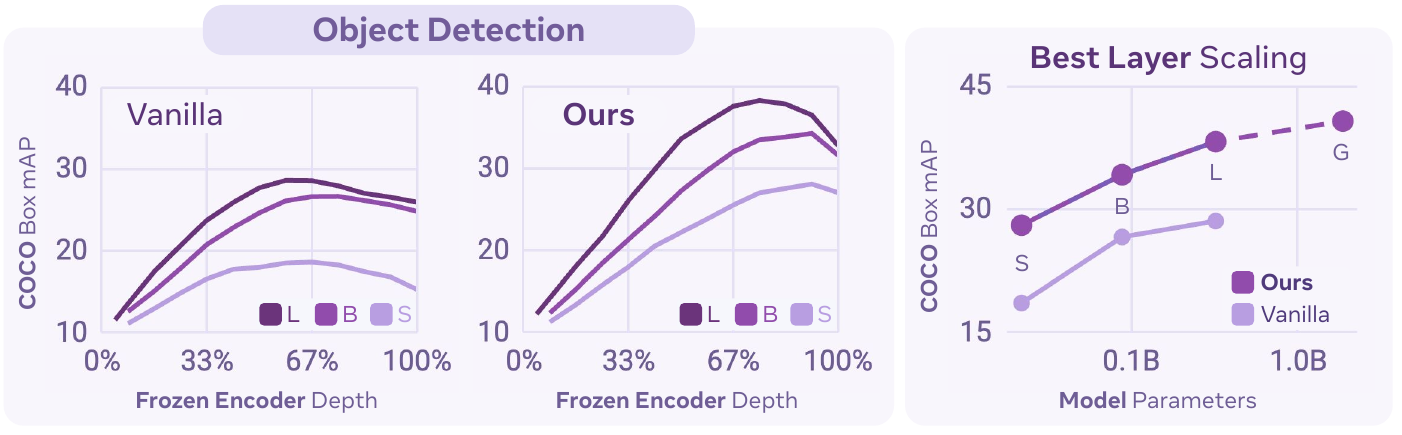}
    \end{tabular}
    \end{center}
    \caption{{\bf The Downstream Scalability of Robust Pretraining.} Left: frozen feature layer analysis of the S/14, B/14, and L/14 models from Fig.~\ref{fig:core_pt_scaling} using the same setup as Fig.~\ref{fig:layerfinder_ablations}. Right: scaling behavior of the \textit{best layer} for each model. Note: G is our final model and has a different schedule.
    }
    \label{fig:layerfinder_scaling_coco}
\vspace{-10pt}
\end{wrapfigure}

\paragraph{Scaling Behavior.}
Finding a simple, easily scalable vision pretraining method that produces generally useful features has been the white whale of the vision community for a while.
Evidently, our robust recipe can enable contrastive pretraining to produce general features. So that begs the question, ``does it scale?''

\begin{wrapfigure}{r}{0.6\textwidth}
\vspace{18pt}
  \begin{center}
      \begin{tabular}{c}
        \includegraphics[width=1\linewidth, trim = 10.7in 0in 0in 5.52in, clip]{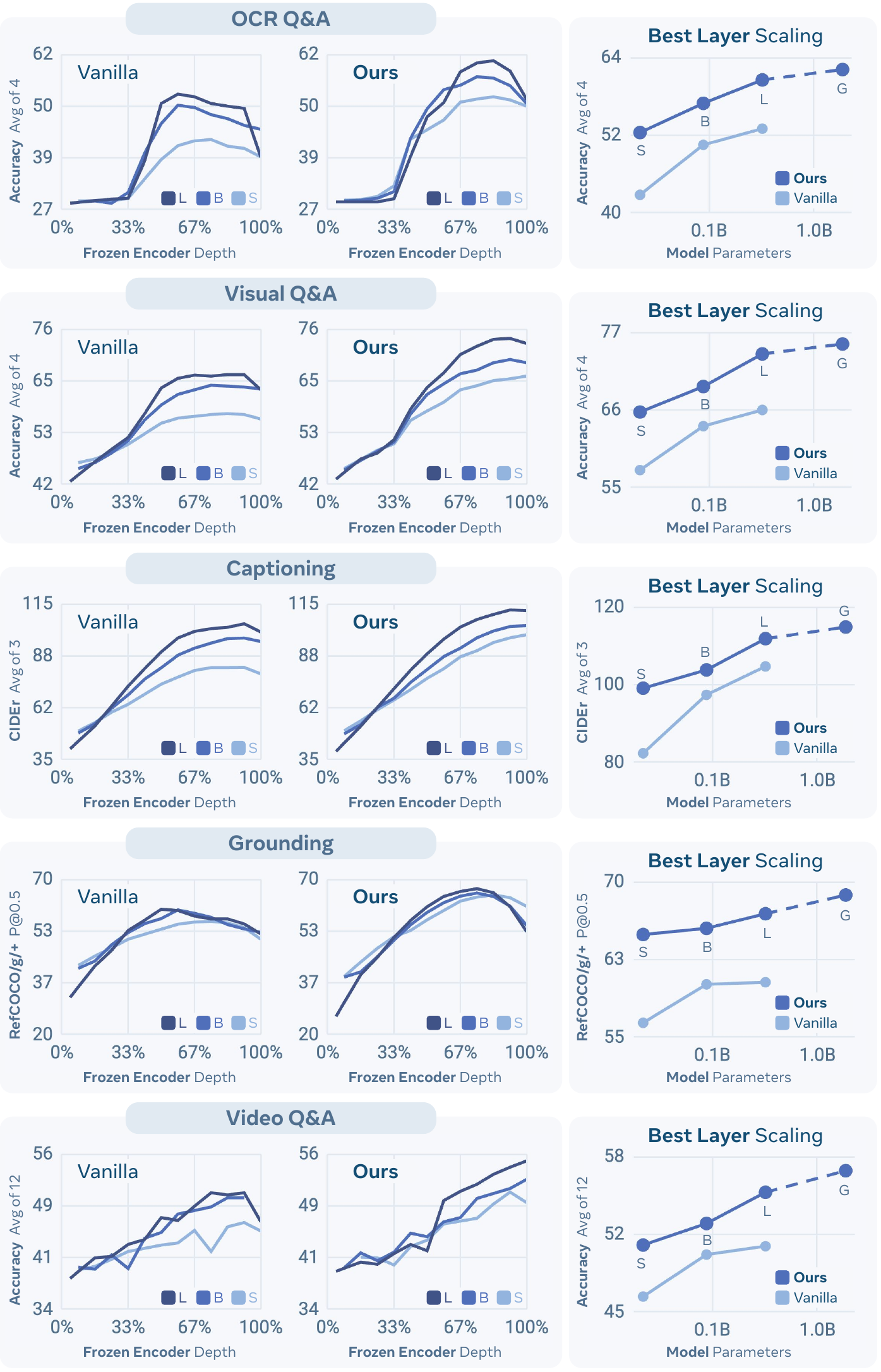}
    \end{tabular}
    \end{center}
    \caption{{\bf Further Scalability Analysis.} We repeat the analysis from Fig.~\ref{fig:layerfinder_scaling_coco} on a wide range of downstream tasks by adapting to a language model. Each category is an average of several downstream tasks (see \S\ref{sec:la}).
    }
    \label{fig:layerfinder_scaling_lang}
\vspace{-30pt}
\end{wrapfigure}

We can answer this question in the same way: by performing frozen feature layer analysis of our S/14, B/14, and L/14 scaling ablation checkpoints from Fig.~\ref{fig:core_pt_scaling}. We report the result of that analysis in Fig.~\ref{fig:layerfinder_scaling_coco}. We also include our final \PEcore{}G model using the same setup, but note this is an estimate as our ablation and final schedules are different.

Immediately, we see a stark contrast between the scaling behavior of the vanilla CLIP recipe and ours. While the vanilla recipe quickly plateaus at L scale (300M), the best layer of our robust pretraining recipe demonstrates scaling to G scale (2B) and potentially beyond---despite being trained with a decidedly non-spatially aligned global contrastive loss. However, this is the \textit{best} layer. The \textit{last} layer performance still stagnates for both the vanilla recipe and ours. This may be why prior work~\cite{webdino} finds contrastive pretraining to not scale for downstream tasks---CLIP loss obfuscates its general features even with our recipe, placing them several layers deep.

However, this is just for a single spatial task. To see whether the trend is consistent, we repeat this scaling analysis on a wide variety of downstream language modeling tasks using the same frozen evaluation setup as Fig.~\ref{fig:layerfinder} and report the results in Fig.~\ref{fig:layerfinder_scaling_lang}. Surprisingly, the simple change in pretraining recipe improves scaling for most language tasks as well---including output-side grounding (RefCOCO). Note that in this benchmarking setup, the LLM never sees videos during training so the Video Q\&A per-layer results are noisy. Yet, the best layer trend is still the same.

Clearly, contrastive pretraining with our robust recipe produces strong general features that scale. However, these features are not going to be much use stuck in the middle of the network. To remedy this, in the remaining sections we will discuss methods for \textit{aligning} these general features to the output of the network for both language modeling and spatial tasks.

\clearpage


\section{Perception Encoder: \textit{Language Alignment}}
\label{sec:la}
\vspace{-5pt}
In \S\ref{sec:layerfinder} we have seen that \PEcore{} already possesses useful features for vision-language modeling. 
In this section, we \emph{lift} these features through \textit{alignment tuning} to construct a new encoder, \PElang{}, specialized for multimodal large language models (MLLMs). 
Our principle is to design not only the most performant, but also the most \textit{general} vision encoder for use in MLLM development. 
To this end, we want a single language-aligned encoder that performs well across language models, across input resolutions, and for a wide variety of MLLM tasks.

\paragraph{MLLM Evaluation Tasks.} 
In this section, our main testbed is to adapt vision encoders to MLLMs and test on various MLLM tasks.  
We evaluate the downstream performance of each MLLM across five task categories: (1) \textit{OCR}, \textit{Chart}, \textit{Document Q\&A} on ChartQA~\cite{zheng2024chartqa}, DocVQA~\cite{mathew2021docvqa}, InfoVQA~\cite{mathew2022infographicvqa} and AI2D~\cite{kembhavi2016ai2d}; (2) \textit{Visual Q\&A} on TextVQA~\cite{singh2019textvqa}, OK-VQA~\cite{schwenk2022okvqa}, POPE~\cite{li2023popebenchmark}, and VQAv2~\cite{goyal2017vqav2}; (3) \textit{Captioning} on Flicker~\cite{flickr}, COCO~\cite{coco}, and No Cap~\cite{agrawal2019nocaps}; (4) \textit{Video Understanding} on VideoMME~\cite{fu2024videomme}, STAR~\cite{wu2021star}, TGIF-QA~\cite{jang2017tgif}, EgoSchema~\cite{mangalam2024egoschema}, MVBench~\cite{li2024mvbench}, and PerceptionTest~\cite{patraucean2024perceptiontest}; and finally (5) \textit{Grounding} on RefCOCO~\cite{kazemzadeh2014referitgame}.

\vspace{-5pt}
\subsection{Language Alignment Method}
\label{sec:la_method}
We begin by searching for the optimal language alignment method.
We design our alignment tuning based on the \textit{midtraining} stage of Perception Language Model (PLM)~\cite{PLM}, which is to adapt \PEcore{} to a pretrained decoder-only LLM (Llama 3~\cite{llama3}) connected by a vision projector. 
We start with ``warmup'' training stage with autoregressive next-token prediction loss on 1M image-text samples from pretraining, where everything but the projector is frozen. Then, we proceed to finetune all parameters on 70M data samples~\cite{PLM} covering natural images, documents/charts/diagrams, and videos, using the same next-token prediction loss.
After completing this language alignment, we extract the vision encoder from the model and refer to it as~\PElang{}.

\begin{wraptable}{r}{0.375\textwidth}
\vspace{-8pt}
    \centering
    \makebox[\linewidth][c]{
    \tablestyle{0pt}{1.05} 
    \setlength{\ccustomlen}{1.5cm}
    \begin{tabular}{wwwwwww awwww}
        \shline
            \ccustom{LLM scale}{}
            & \ccustom{LLM unfrozen}{}
            & \ccustom{Regularization?}{}
            & \ccustom{Projector}{}
            & \ccustom{Layer}{}
            & \ccustom[c4]{\textbf{Avg.}}{}
            & \ccustom[c4]{{OCR Q\&A}}{Average of 4}
            & \ccustom[c4]{Captioning}{Average of 3}
            & \ccustom[c4]{Visual Q\&A}{Average of 4}
            & \ccustom[c4]{Video Q\&A}{Average of 6}
            \\
        \hline
        \multicolumn{5}{l}{\it \addpadding LLM Setup}&\ca{}\\
        1B &    &  & MLP     & 47  & \ca{76.5} & 60.7 & 115.1 & 76.0 & 54.0 \\
        3B &   &  & MLP     & 47  & \ca{78.1} & 65.9 & 115.7 & 76.6 & 54.1 \\
        3B & $\checkmark$  &  & MLP  & 47  & \ca{78.4} & 65.8 & 117.6 & 76.3 & 53.7 \\
        \hline
        \multicolumn{5}{l}{\it \addpadding Vision Projector}&\ca{}\\
        3B &    &  & Linear  & 47  & \ca{77.2} & 64.5 & 114.1 & 76.5 & 53.7 \\
        3B &    &  & MLP    & 47  & \ca{78.1} & 65.9 & 115.7 & 76.6 & 54.1 \\
        \hline
        \multicolumn{5}{l}{\it \addpadding PE Output Layer}&\ca{}\\
        3B &    &  & MLP     & 50  & \ca{75.9} & 56.6 & 116.7 & 76.5 & 53.7 \\
        3B &    &  & MLP     & 47  & \ca{78.1} & 65.9 & 115.7 & 76.6 & 54.1 \\
        3B &    &  & MLP     & 41 & \ca{76.9} & 65.5 & 112.8 & 75.4 & 53.9 \\
        \hline
        \multicolumn{5}{l}{\it \addpadding PE Regularization}&\ca{}\\
        3B &    & $\checkmark$ & MLP  & 47 & \ca{79.9} & 69.0 & 117.5 & 77.4 & 55.6 \\
        3B & $\checkmark$      & $\checkmark$ & MLP & 47 & \ca{\textbf{80.1}} & 68.7 & 118.3 & 77.0 & 56.3 \\
        \shline
    \end{tabular}
    }
    \caption{{\bf Language Alignment.} 
    We find the best configuration to language align \PEcore{G} using autoregressive language training. 
    }
    \vspace{-30pt}
    \label{tab:align_ablation}
\end{wraptable}

To arrive at the optimal training configuration presented in PLM~\cite{PLM}, we first conduct ablation studies using a 20M subset of the data.
In Tab.~\ref{tab:align_ablation}, we ablate the LLM sizes, training parameters, vision projector types, output layers to project, and encoder regularization. 
We evaluate across OCR Q\&A, Captioning, Visual Q\&A, and Video Q\&A and find the best configuration. 

\textbf{LLM Setup.} We explore different \textit{scales} (1B or 3B parameters) and \textit{freezing} weights of the LLM. We observe that going from 1B to 3B parameters increases average score by 1.6 points (76.5$\rightarrow$78.1). Unfreezing the LLM boosts this number to 78.4. 

\textbf{Vision Projector.}  Using a \textit{2-layer MLP} vision projector instead of a \textit{linear layer} improves the average score from 77.2 to 78.1, while only adding few parameters (13.5M $\rightarrow$ 27M).

\textbf{PE Output Layer.} 
As shown in \S\ref{sec:layerfinder}, \PEcore{G} has intermediate layers that perform significantly better than the last layer when used as features for certain tasks. However, it is not clear if that same behavior applies when finetuning. We test applying the projector to layers 41, 47, and 50 (the last layer), and find that layer 47 works best. Incidentally, this is also the optimal layer for frozen VQ\&A in Fig.~\ref{fig:layerfinder}.

\textbf{PE Regularization.} We apply LayerScale~\cite{layerscale} and DropPath~\cite{droppath} to the vision encoder during the alignment, for stabilizing training. 
This improves the 78.1 average score to 79.9 ($+1.8$ points). 
Unfreezing the LLM boosts this number further to 80.1. 
We choose this configuration (last row) as our final alignment setup. 

To construct \PElang{}, we scale this recipe up the 70M samples mentioned above (more details in~\cite{PLM}). 
In summary, we use a pretrained Llama3.2 3B, unfrozen, with a 2-layer MLP as a vision projector on top of layer \PEcore{G} layer 47 (with the last 3 discarded) and regularize the encoder with LayerScale and DropPath. 
Compared to the 20M sample ablation setting in Tab.~\ref{tab:align_ablation}, the final \PElang{} trained on 70M total samples gives another +2.1 points to 82.2 on the average across OCR Q\&A, Captioning, Visual Q\&A, and Video Q\&A.

\label{sec:lang_layerfinder}

\paragraph{Effects.}
The goal of alignment tuning is to \textit{lift} the strong features found in intermediate layers of \PEcore{} described in \S\ref{sec:layerfinder} to the end of the network. To see if we actually accomplished that, we perform the same layerwise
\begin{wrapfigure}{r}{0.45\textwidth}
\vspace{-0pt}
  \begin{center}
      \begin{tabular}{c}
        \includegraphics[width=1\linewidth, trim = 13.8in 0in 0in 14.59in, clip]{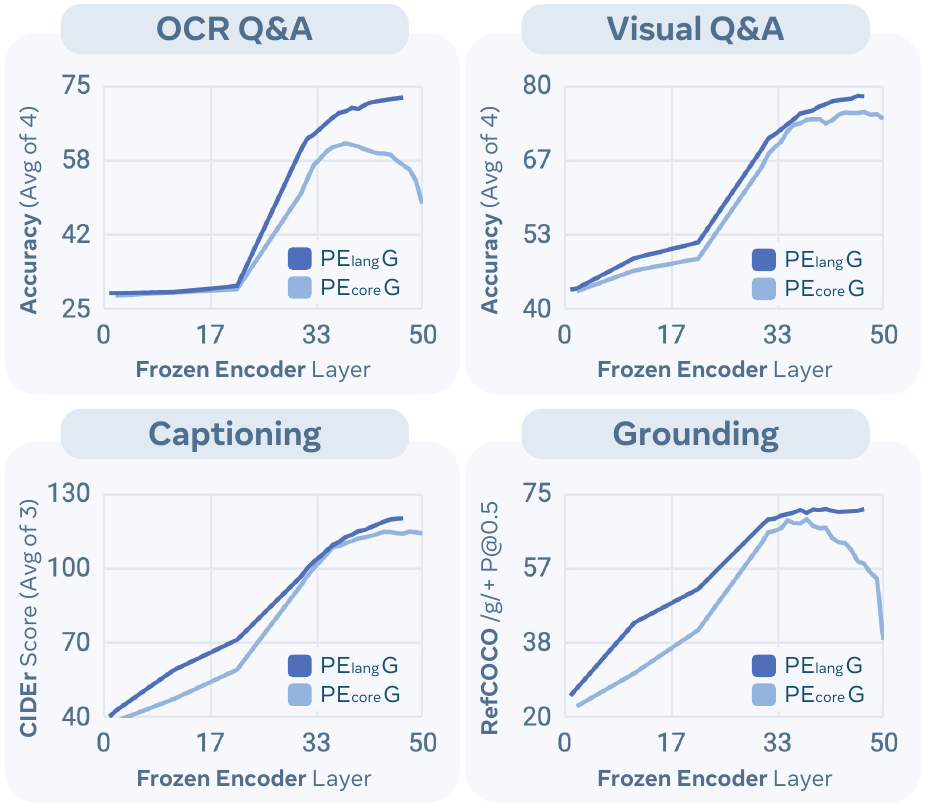}
    \end{tabular}
    \end{center}
    \caption{{\bf Language Alignment.} We analyze how language alignment changes the internal features of PE. 
    Similar to our \PEcore{} analysis in Fig.~\ref{fig:layerfinder_scaling_lang}, we extract \PElang{} and adapt each layer to a new LLM. 
    }
    \label{fig:lang_language_alignment_analysis}
\vspace{-10pt}
\end{wrapfigure}
analysis as in Fig.~\ref{fig:layerfinder} on our final \PElang{G} model and compare it to the original \PEcore{G} checkpoint it was initialized from.  We present the results of this analysis in Fig.~\ref{fig:lang_language_alignment_analysis}, and immediately we see that language alignment was a success: across all categories, the performing layer for the aligned model was the last, no matter the performance of the original checkpoint.
Notably, our \PElang{} training mix did \textit{not} contain grounding data, which means that this significantly lifted grounding performance is entirely due to the strong intermediate grounding features in \PEcore{} now being aligned to the end of the network.
Moreover, specific domains such as OCR Q\&A that \textit{were} represented in the training mix see a significant boost to performance compared to even the best layer of \PEcore{}, which was already strong. 
Thus, with an order of magnitude fewer samples compared to pretraining, we were able to \textit{language align} \PEcore{G} to create a single, strong encoder for all visual language modeling tasks.
Following this success, we align \PEcore{L} in a similar manner to construct \PElang{L} (see~\cite{PLM}).
\label{sec:la_layer_find}

\begin{table*}[b!]
    \centering
    \vspace{-10pt}
    \makebox[\linewidth][c]{
    \tablestyle{0pt}{1.05} 
        \begin{tabular}{y{50}wx{20} awwww awwww awww a awwwwww}
        \shline
        \multirow{2}{*}{\vspace{-2.2cm} Model}  &&& \multicolumn{5}{c}{\ct[c3]{\it OCR / Chart / Doc. Q\&A}} %
        & \multicolumn{5}{c}{\ct[c4]{\it Visual Q\&A}} & \multicolumn{4}{c}{\ct[c5]{\it Captioning}} & \multicolumn{1}{c}{\ct[c6]{}} & \multicolumn{7}{c}{\ct[c7]{\it Video}} \\
            & \cb{Encoder Params}{}
            & \cb{Resolution}{Patch Size}
            & \cb[c3]{\textit{\textbf{Avg. OCR QA}}}{}
            & \cb[c3]{ChartQA}{Acc.~\cite{zheng2024chartqa}}
            & \cb[c3]{DocVQA}{Acc.~\cite{mathew2021docvqa}}
            & \cb[c3]{Info. QA}{Acc.~\cite{mathew2022infographicvqa}}
            & \cb[c3]{AI2D}{Acc.~\cite{kembhavi2016ai2d}}
            & \cb[c4]{\textit{\textbf{Avg. VQA}}}{}
            & \cb[c4]{TextVQA}{Acc.~\cite{singh2019textvqa}}
            & \cb[c4]{OK-VQA}{Acc.~\cite{schwenk2022okvqa}}
            & \cb[c4]{POPE}{Acc. ~\cite{li2023popebenchmark}}
            & \cb[c4]{VQAv2}{Acc.~\cite{goyal2017vqav2}}
            & \cb[c5]{\textit{\textbf{Avg. Cap.}}}{}
            & \cb[c5]{Flicker}{CIDEr~\cite{flickr}}
            & \cb[c5]{COCO}{CIDEr ~\cite{coco}}
            & \cb[c5]{No Cap}{CIDEr~\cite{agrawal2019nocaps}}
            & \cb[c6]{\textit{\textbf{Avg. Ground.}}}{RefCOCO/g/+~\cite{kazemzadeh2014referitgame}} 
            & \cb[c7]{\textit{\textbf{Avg. Video}}}{}
            & \cb[c7]{VideoMME}{Acc.~\cite{fu2024videomme}}
            & \cb[c7]{STAR}{Acc.~\cite{wu2021star}}
            & \cb[c7]{TGIF-QA}{Acc.~\cite{jang2017tgif}}
            & \cb[c7]{EgoSchema}{Acc.~\cite{mangalam2024egoschema}}
            & \cb[c7]{MVBench}{Acc.~\cite{li2024mvbench}}
            & \cb[c7]{PerceptionTest}{Acc.~\cite{patraucean2024perceptiontest}} \\
        \hline
\multicolumn{1}{l}{{\textit{256 Tokens per Image}}}                & & & \cat{} &&&&& \ca{} &&&&& \ca{} &&&& \ca{} & \ca{} &&&&&&  \\
MetaCLIP-L~\cite{metaclip}                  & 0.3B           & \rp{224}{14} & \ca{44.9} & 47.9 & 33.0 & 28.7 & 70.2 & \ca{68.4} & 47.6 & 62.5 & 86.9 & 76.5 & \ca{110.5} & 87.5 & 130.0 & 114.1& \ca{60.6}  & \ca{53.9} & 46.1 & 51.0 & 66.4 & 58.6 & 49.4 & 51.9  \\
MetaCLIP-G~\cite{metaclip}                  & 1.8B             & \rp{224}{14} & \ca{44.8} & 47.6 & 33.1 & 27.9 & 70.6 & \ca{68.8} & 48.2 & 63.5 & 86.5 & 76.9 & \ca{111.1} & 86.5 & 132.1 & 114.8& \ca{60.5}  & \ca{53.1} & 45.0 & 50.7 & 66.4 & 56.0 & 48.7 & 51.9  \\
\textbf{\PElang{} G}$^\dagger$                 & \,\,\,1.7B$^*$          & \rp{224}{14} & \ca{53.7} & 61.3 & 47.1 & 32.2 & 74.1 & \ca{71.8} & 55.1 & 65.3 & 86.8 & 79.8 & \ca{116.4} & 91.0 & 136.9 & 121.2 & \ca{65.7} & \ca{55.5} & 47.3 & 55.7 & 68.9 & 59.6 & 48.6 & 52.9 \\

\hline
\multicolumn{1}{l}{{\textit{576 Tokens per Image}}}                 & & & \cat{} &&&&& \ca{} &&&&& \ca{} &&&& \ca{} & \ca{} &&&&&&  \\
CLIP~\cite{clip}                          & 0.3B           & \rp{336}{14} & \ca{53.5} & 61.7 & 49.5 & 32.8 & 70.1 & \ca{72.7} & 60.7 & 63.9 & 87.3 & 78.9 & \ca{113.3} & 92.0 & 132.9 & 115.0& \ca{65.0}  & \ca{54.2} & 46.3 & 52.1 & 68.6 & 57.4 & 48.5 & 52.3  \\
AIMv2-L~\cite{aimv2}                        & 0.3B           & \rp{336}{14} & \ca{53.3} & 61.6 & 48.0 & 32.1 & 71.4 & \ca{73.7} & 62.7 & 64.3 & 87.7 & 80.1 & \ca{115.2} & 90.9 & 135.6 & 119.2 & \ca{63.3} & \ca{52.5} & 44.3 & 50.9 & 67.5 & 54.4 & 44.9 & 53.2  \\
AIMv2 L Dist.~\cite{aimv2}                      & 0.3B           & \rp{336}{14} & \ca{53.7} & 61.1 & 49.4 & 31.5 & 72.7 & \ca{74.1} & 62.8 & 64.8 & 88.3 & 80.3 & \ca{117.8} & 94.7 & 137.5 & 121.2& \ca{62.6}  & \ca{53.8} & 44.3 & 52.4 & 65.0 & 57.4 & 50.0 & 53.6  \\
SigLIP2-so~\cite{siglip2}                 & 0.4B           & \rp{384}{16} & \ca{58.9} & 69.0 & 58.3 & 35.2 & 73.1 & \ca{76.8} & 69.8 & \textbf{67.2} & 88.7 & 81.6 & \ca{116.5} & 92.1 & 137.7 & 119.8& \ca{67.4}  & \ca{54.5} & 45.5 & 53.1 & 67.2 & 57.6 & 49.3 & 54.5  \\
SigLIP2-g-opt~\cite{siglip2}                 & 1.1B                       & \rp{384}{16} & \ca{56.2} & 63.1 & 55.3 & 34.0 & 72.4 & \ca{77.0} & 70.3 & 66.7 & 89.6 & 81.6 & \ca{117.7} & 94.9 & 137.8 & 120.3& \ca{66.5}  & \ca{53.9} & 46.2 & 53.9 & 66.6 & 53.8 & 48.5 & 54.7  \\
\textbf{\PElang{} G}$^\dagger$                 & \,\,\,1.7B$^*$           & \rp{336}{14} & \ca{66.9} & 76.8 & 73.6 & 41.1 & 76.1 & \ca{76.2} & 68.5 & 66.0 & 89.1 & 81.3 & \ca{119.7} & 96.1 & 139.6 & 123.4 & \ca{68.9} & \ca{58.1} & 48.7 & 58.9 & 70.5 & 61.8 & 52.7 & 55.9 \\
\hline
\multicolumn{1}{l}{{\textit{1024 Tokens per Image}}}                 & & & \cat{} &&&&& \ca{} &&&&& \ca{} &&&& \ca{} & \ca{} &&&&&&  \\
InternViT 2.5 L~\cite{chen2024internvit2p5}  & 0.3B           & \rp{448}{14} & \ca{60.6} & 74.1 & 59.2 & 35.9 & 73.1 & \ca{74.2} & 65.4 & 64.4 & 87.6 & 79.6 & \ca{112.3} & 88.4 & 133.7 & 114.9& \ca{66.9}  & \ca{50.6} & 45.2 & 44.8 & 62.7 & 54.2 & 46.0 & 50.5  \\
SigLIP2-so~\cite{siglip2}                 & 0.4B           & \rp{512}{16} & \ca{63.3} & 72.1 & 69.3 & 39.0 & 72.7 & \ca{77.9} & 74.8 & 66.0 & 89.0 & \textbf{81.8} & \ca{117.4} & 93.5 & 138.3 & 120.2& \ca{69.6}  & \ca{55.8} & 46.2 & 55.4 & 67.0 & \textbf{62.0} & 50.0 & 54.5  \\
\textbf{\PEcore{L}}                      & 0.3B & \rp{448}{14}           & \ca{59.4} & 68.7 & 62.5 & 36.6 & 69.7 & \ca{74.7} & 67.7 & 64.3 & 88.3 & 78.7 & \ca{112.7} & 89.6 & 133.4 & 114.9& \ca{59.7}  & \ca{50.9} & 41.7 & 51.2 & 61.6 & 52.6 & 47.4 & 50.6  \\
\textbf{\PElang{L}}                      & 0.3B & \rp{448}{14}           & \ca{71.1} & 81.0 & 81.9 & 46.4 & 75.0 & \ca{77.1} & 73.0 & 65.5 & 89.3 & 80.8 & \ca{117.3} & 94.3 & 137.3 & 120.1& \ca{70.5}  & \ca{56.5} & 47.0 & 57.2 & 68.0 & 59.8 & 52.3 & 54.7  \\
\hline
DINOv2-g~\cite{dinov2}                      & 1.1B             & \rp{448}{14} & \cat{30.0} & 19.6 & 14.7 & 24.2 & 61.5 & \ca{61.0} & 19.3 & 60.4 & 88.6 & 75.8 & \ca{109.4} & 86.5 & 131.6 & 110.1& \ca{64.9}  & \ca{49.5} & 39.7 & 52.1 & 60.1 & 46.8 & 47.4 & 50.8 \\
AIMv2 3B~\cite{aimv2}                     & 2.7B             & \rp{448}{14} & \ca{48.9} & 40.5 & 53.9 & 33.9 & 67.2 & \ca{73.0} & 64.1 & 64.0 & 85.2 & 78.9 & \ca{115.7} & 93.8 & 135.2 & 118.1& \ca{36.1}  & \ca{54.6} & 45.1 & 54.5 & 66.7 & 55.4 & 51.7 & 54.3 \\
InternViT2.5-6B~\cite{chen2024internvit2p5}  & 5.5B             & \rp{448}{14} & \ca{59.9} & 72.3 & 59.4 & 35.2 & 72.5 & \ca{75.5} & 68.9 & 64.9 & 88.2 & 80.2 & \ca{115.0} & 92.2 & 136.3 & 116.3& \ca{68.0}  & \ca{49.6} & 44.5 & 47.0 & 62.6 & 45.8 & 48.9 & 48.5 \\
\textbf{\PEcore{G}}                       & 1.9B           & \rp{448}{14} & \ca{60.8} & 69.9 & 65.4 & 36.7 & 71.1 & \ca{73.3} & 65.9 & 60.7 & 88.4 & 78.0 & \ca{112.5} & 91.6 & 133.6 & 112.4 & \ca{66.6}  & \ca{52.0} & 42.3 & 53.1 & 62.9 & 51.4 & 48.8 & 53.6 \\ 
\textbf{\PElang{G}}                 & \,\,\,1.7B$^*$ & \rp{448}{14} & \ca{\textbf{72.4}} & \textbf{80.5} & \textbf{84.4} & \textbf{48.3} & \textbf{76.4} & \ca{\textbf{78.1}} & \textbf{75.2} & 65.4 & \textbf{90.1} & \textbf{81.8} & \ca{\textbf{120.1}} & \textbf{96.6} & \textbf{140.0} & \textbf{123.6} & \ca{\textbf{71.3}} & \ca{\textbf{58.0}} & \textbf{48.0} & \textbf{60.1} & \textbf{69.4} & \textbf{62.0} & \textbf{52.4} & \textbf{56.0} \\

        \shline
    \end{tabular}
    }
    \caption{{\bf MLLM Results with Llama 3.1 8B.} We compare various vision encoders at their native resolution using Llama 3.1-instruct 8B~\cite{llama3} as the language model. The table compares models of similar class in number of vision tokens and parameters. \PElang{} shows strong performance across all benchmarks, including against models 3$\times$ its size. $^*$\PElang{} has 1.7B parameters since we discard the last 3 layers during language alignment. $^\dagger$Interpolated without extra training. }
    \label{tab:lang_mllm_bench}
\end{table*}

\subsection{Comparisons with Existing Vision Encoders}
\label{sec:la_main_results}

We compare \PEcore{} and \PElang{} with other vision encoders that are popular choices in MLLM literature: MetaCLIP~\cite{metaclip}, SigLIP2~\cite{siglip2}, CLIP~\cite{clip}, AIMv2~\cite{aimv2}, DINOv2~\cite{dinov2}, and InternViT2.5~\cite{chen2024internvit2p5}.
Overall, these encoders span several different pretraining losses (e.g., contrastive, captioning, self-supervised, and mixed supervision), encoder sizes (from 300M to 6B parameters), and resolutions (from 224 to 512).
\textit{For all vision encoders, we find the best intermediate layers to train MLLM for fair comparison} (more in Appendix~\ref{appx:mmlm_benchmark_set}).

\paragraph{MLLM Benchmarking Setup.}
\label{sec:mllm_bench_setting} 
We connect each vision encoder, including \PElang{}, to a language decoder with a fresh 2-layer MLP projector. Similar to the alignment stage, we first train only the projector on a subset of 1M image-text pairs from pretraining. Then, we train both the projector and LLM on 2.6M visual Q\&A pairs, image captions, and image grounding samples (see Appendix~\ref{appx:mmlm_benchmark_set} for details). We benchmark at the native resolution of each encoder (with higher resolution tiling results in Appendix~\ref{appx:mmlm_benchmark_results}). Finally, we ablate over two language decoders, Llama 3.1 8B~\cite{llama3} and QwenLM 2.5 7B~\cite{qwen2.5}, to measure generalization across LLMs.

\paragraph{Results.}
Tab.~\ref{tab:lang_mllm_bench} shows benchmarks results for native resolution input across existing encoders, \PEcore{} and \PElang{}. Notably, AIMv2~\cite{aimv2}, InternViT2.5~\cite{chen2024internvit2p5}, SigLIP2~\cite{siglip2} and \PElang{} are trained jointly with a language decoder using next token prediction objective, and thus they perform better overall compared to the base contrastive and self-supervised models across all the metrics. However, \PElang{} uses a fraction of the training FLOPs for language alignment tuning, while significantly outperforming all vision encoders by large margin (an average of +3.5 points for G and +2.0 points for L). 
Similarly, when tiling with 4 tiles and 1 thumbnail (see Appendix Tab.~\ref{tab:lang_mllm_bench_tiling}), both \PElang{L} and \PElang{G} outperform all existing vision encoders, including InternViT2.5~\cite{chen2024internvit2p5}, which was specifically pretrained in a tiling setting and with grounding data.
Appendix \ref{appx:mmlm_benchmark_results}, shows a breakdown of the RefCOCO results, as well as results for tiling with higher resolution.

\paragraph{Transferability.} As \PElang{} is aligned with Llama 3.2-instruct 3B, we conduct a separate set of experiments to check if our model performs well with a different base LLM. In Tab.~\ref{tab:lang_mllm_bench_qwen} we repeat the native resolution comparison with QwenLM 2.5 7B~\cite{qwen2.5}. Interestingly, \PElang{} not only outperforms all vision encoders in this setting, but it also outperforms InternViT2.5~\cite{chen2024internvit2p5}, which is specifically aligned to QwenLM 2~\cite{qwen2} throughout midtraining. In fact, \PElang{G} with QwenLM even improves its performance with Llama in some cases like with OCR Q\&A and video benchmarks, emphasizing the generality of our language alignment.

\begin{table*}[ht]
    \centering
    \vspace{-3pt}
    \makebox[\linewidth][c]{
    \tablestyle{0pt}{1.05} 
        \begin{tabular}{y{50}wx{20} awwww awwww awww a awwwwww}
        \shline
        \multirow{2}{*}{\vspace{-2.2cm} Model}  &&& \multicolumn{5}{c}{\ct[c3]{\it OCR / Chart / Doc. Q\&A}} %
        & \multicolumn{5}{c}{\ct[c4]{\it Visual Q\&A}} & \multicolumn{4}{c}{\ct[c5]{\it Captioning}} & \multicolumn{1}{c}{\ct[c6]{}} & \multicolumn{7}{c}{\ct[c7]{\it Video}}\\
            & \cb{Encoder Params}{}
            & \cb{Resolution}{Patch Size}
            & \cb[c3]{\textit{\textbf{Avg. OCR QA}}}{}
            & \cb[c3]{ChartQA}{Acc.~\cite{zheng2024chartqa}}
            & \cb[c3]{DocVQA}{Acc.~\cite{mathew2021docvqa}}
            & \cb[c3]{Info. QA}{Acc.~\cite{mathew2022infographicvqa}}
            & \cb[c3]{AI2D}{Acc.~\cite{kembhavi2016ai2d}}
            & \cb[c4]{\textit{\textbf{Avg. VQA}}}{}
            & \cb[c4]{TextVQA}{Acc.~\cite{singh2019textvqa}}
            & \cb[c4]{OK-VQA}{Acc.~\cite{schwenk2022okvqa}}
            & \cb[c4]{POPE}{Acc. ~\cite{li2023popebenchmark}}
            & \cb[c4]{VQAv2}{Acc.~\cite{goyal2017vqav2}}
            & \cb[c5]{\textit{\textbf{Avg. Cap.}}}{}
            & \cb[c5]{Flicker}{CIDEr~\cite{flickr}}
            & \cb[c5]{COCO}{CIDEr ~\cite{coco}}
            & \cb[c5]{No Cap}{CIDEr~\cite{agrawal2019nocaps}}
            & \cb[c6]{\textit{\textbf{Avg. Ground.}}}{RefCOCO/g/+~\cite{kazemzadeh2014referitgame}} 
            & \cb[c7]{\textit{\textbf{Avg. Video}}}{}
            & \cb[c7]{VideoMME}{Acc.~\cite{fu2024videomme}}
            & \cb[c7]{STAR}{Acc.~\cite{wu2021star}}
            & \cb[c7]{TGIF-QA}{Acc.~\cite{jang2017tgif}}
            & \cb[c7]{EgoSchema}{Acc.~\cite{mangalam2024egoschema}}
            & \cb[c7]{MVBench}{Acc.~\cite{li2024mvbench}}
            & \cb[c7]{PerceptionTest}{Acc.~\cite{patraucean2024perceptiontest}} \\
        \hline
\multicolumn{1}{l}{{\textit{576 Tokens per Image}}}                & & & \cat{} &&&&& \ca{} &&&&& \ca{} &&&& \ca{} & \ca{} &&&&&& \\
SigLIP2-so~\cite{siglip2}                 & 0.4B           & \rp{384}{16} & \ca{60.5} & 72.0 & 59.1 & 36.7 & 74.3 & \ca{76.2} & 69.0 & 65.4 & 89.2 & 81.1 & \ca{116.3} & 91.6 & 137.3 & 120.0 & \ca{70.0} & \ca{57.0} & 51.3 & 55.8 & 66.0 & 61.0 & 51.9 & 55.7 \\
SigLIP2-g-opt~\cite{siglip2}                 & 1.1B             & \rp{384}{16} & \ca{60.8} & 71.0 & 60.4 & 36.7 & 75.2 & \ca{76.8} & 70.3 & 65.6 & 89.5 & 81.8 & \ca{118.8} & 96.4 & 139.0 & 121.1 & \ca{69.9} & \ca{58.3} & 52.0 & 57.6 & 68.1 & 62.0 & 52.8 & 57.4 \\
\textbf{\PElang{} G}$^\dagger$                 & \,\,\,1.7B$^*$ & \rp{336}{14} & \ca{66.8} & 77.5 & 72.4 & 41.1 & 76.4 & \ca{76.0} & 67.9 & 65.4 & 89.1 & 81.5 & \ca{118.8} & 94.6 & 139.5 & 122.3 & \ca{70.1} & \ca{60.2} & 54.6 & 61.7 & 69.8 & 63.6 & 54.3 & 57.2 \\
\hline
\multicolumn{1}{l}{{\textit{1024 Tokens per Image}}}                & & & \cat{} &&&&& \ca{} &&&&& \ca{} &&&& \ca{} & \ca{} &&&&&& \\
InternViT2.5~\cite{chen2024internvit2p5}  & 0.3B           & \rp{448}{14} & \ca{60.3} & 75.4 & 61.1 & 36.2 & 68.4 & \ca{74.2}          & 65.6 & 63.7 & 87.8 & 79.5 & \ca{112.1} & 88.5 & 133.5 & 114.1 & \ca{68.1} & \ca{55.8} & 50.3 & 54.7 & 66.6 & 59.0 & 50.6 & 53.8 \\
SigLIP2-so~\cite{siglip2}                 & 0.4B         & \rp{512}{16} & \ca{66.3} & 77.2 & 71.9 & 42.4 & 73.9 & \ca{\textbf{77.9}} & 74.2 & 65.6 & 89.9 & 81.8 & \ca{117.1} & 93.0 & 138.0 & 120.3 & \ca{70.5} & \ca{55.9} & 50.3 & 57.3 & 67.2 & 62.6 & 50.3 & 47.4 \\
\textbf{\PEcore{L}} & 0.3B & \rp{448}{14} & \ca{63.5} & 73.9 & 67.4 & 40.5 & 72.2 & \ca{75.7} & 69.2 & 64.0 & 89.4 & 80.2 & \ca{113.3} & 88.7 & 135.2 & 115.9 & \ca{66.5} & \ca{57.3} & 49.6 & 57.8 & 67.7 & 60.8 & 52.3 & 55.5 \\
\textbf{\PElang{L}} & 0.3B & \rp{448}{14} & \ca{70.2} & 80.6 & 80.7 & 46.0 & 73.5 & \ca{76.8} & 72.8 & 64.1 & 89.4 & 81.0 & \ca{116.4} & 93.4 & 137.6 & 118.1 & \ca{70.4} & \ca{58.3} & 51.6 & 59.8 & 67.4 & 62.2 & 53.4 & 55.4 \\
\hline
\addpadding
DINOv2~\cite{dinov2}                      & 1.1B             & \rp{448}{14} & \cat{31.3} & 21.7 & 14.7 & 24.6 & 64.3 & \ca{61.0} & 18.9 & 59.5 & 88.9 & 76.9 & \ca{110.1} & 87.3 & 132.1 & 110.8 & \ca{69.3} & \ca{54.3} & 46.9 & 56.5 & 63.4 & 56.8 & 49.7 & 52.2 \\
AIMv2 3B~\cite{aimv2}                     & 2.7B             & \rp{448}{14} & \ca{66.0} & 76.7 & 70.5 & 41.4 & 75.2 & \ca{\textbf{77.9}} & 74.2 & \textbf{66.2} & 89.4 & \textbf{81.9} & \ca{\textbf{119.2}} & \textbf{96.4} & 139.2 & 122.0 & \ca{67.6} & \ca{56.3} & 45.9 & 58.0 & 67.8 & 60.8 & 51.4 & 53.9 \\
InternViT2.5~\cite{chen2024internvit2p5}  & 5.5B             & \rp{448}{14} & \ca{64.2} & 78.2 & 65.3 & 39.6 & 73.6 & \ca{76.4} & 70.1 & 64.5 & 89.3 & 81.7 & \ca{117.6} & 95.9 & 138.4 & 118.6 & \ca{\textbf{72.8}} & \ca{56.1} & 50.3 & 59.1 & 67.3 & 56.6 & 51.1 & 52.2 \\
\textbf{\PEcore{G}}                       & 1.9B           & \rp{448}{14} & \ca{64.8} & 75.9 & 68.8 & 41.6 & 72.9 & \ca{75.2} & 67.9 & 62.4 & 89.7 & 80.7 & \ca{113.1} & 91.7 & 135.2 & 112.3 & \ca{70.5} & \ca{57.0} & 48.7 & 58.3 & 66.9 & 60.8 & 52.9 & 54.5 \\
\textbf{\PElang{G}}                    & \,\,\,1.7B$^*$ & \rp{448}{14} & \ca{\textbf{72.9}} & \textbf{81.6} & \textbf{83.7} & \textbf{49.5} & \textbf{76.7} & \ca{\textbf{77.9}} & \textbf{74.9} & 64.5 & \textbf{90.3} & \textbf{81.9} & \ca{118.9} & 94.6 & \textbf{139.8} & \textbf{122.3} & \ca{72.1} & \ca{\textbf{60.4}} & \textbf{54.1} & \textbf{62.5} & \textbf{68.3} & \textbf{66.6} & \textbf{54.2} & \textbf{56.8} \\
        \shline
    \end{tabular}
   }
    \caption{{\bf MLLM Results with QwenLM 2.5 7B.} Same setting as Tab.~\ref{tab:lang_mllm_bench}, but with QwenLM2.5 7B~\cite{qwen2.5} as the language model. Although \PElang{} is aligned to Llama3.2 3B, the language alignment transfers well to a different language model.   }
    \label{tab:lang_mllm_bench_qwen}
    \vspace{-14pt}
\end{table*}

\vspace{10pt}

\paragraph{System-Level MLLM Comparison.} In Tab.~\ref{tab:lang_mllm_system_level}, we conduct a system-level comparison to the state-of-the-art open-access MLLMs: LLaVA-OneVision 7B~\cite{llava-onevision}, Gemma3 12B~\cite{gemma3}, Molmo-D 7B~\cite{molmo}, Qwen2 VL 7B~\cite{qwen2vl}, InternVL 2.5 8B~\cite{chen2024internvit2p5} and the very recent InternVL 3 8B~\cite{internvl3}. 
Each baseline uses a contrastively pretrained ViT (SigLIP-so400M~\cite{siglip}, CLIP-L~\cite{clip}, DFN-H~\cite{dfn}, and InternViT 2.5 300M~\cite{chen2024internvit2p5}). 
For our PLM-8B we use \PElang{G} as the vision encoder with 36 tiles for images and 32 frames for video and Llama 3.1-instruct 8B as the language decoder (more details in~\cite{PLM}).
We show numbers from their respective works or evaluate them ourselves if they are not reported (except for Gemma and InternVL 3). PLM-8B outperforms all other models tested, emphasizing that \PElang{G} can be used to drive strong results across a wide range of tasks.


\begin{table*}[ht]
    \centering
    \vspace{-2pt}
    \makebox[\linewidth][c]{
    \tablestyle{0pt}{1.05} 
        \begin{tabular}{y{52} y{52} awwww awwww awww awwwwww}
        \shline
        \multirow{2}{*}{\vspace{-2.3cm} Model}  & & \multicolumn{5}{c}{\ct[c3]{\it OCR / Chart / Doc. Q\&A}} %
        & \multicolumn{5}{c}{\ct[c4]{\it Visual Q\&A}} & \multicolumn{4}{c}{\ct[c5]{\it Captioning}} & \multicolumn{7}{c}{\ct[c6]{\it Video}}\\
            & Encoder 
            & \cb[c3]{\textit{\textbf{Avg. OCR QA}}}{}
            & \cb[c3]{ChartQA}{Acc.~\cite{zheng2024chartqa}}
            & \cb[c3]{DocVQA}{Acc. (test)~\cite{mathew2021docvqa}}
            & \cb[c3]{Info. QA}{ Acc. (test)~\cite{mathew2022infographicvqa}}
            & \cb[c3]{AI2D}{w/o mask~\cite{kembhavi2016ai2d}}
            & \cb[c4]{\textit{\textbf{Avg. VQA}}}{}
            & \cb[c4]{TextVQA}{Acc.~\cite{singh2019textvqa}}
            & \cb[c4]{OK-VQA}{Acc.~\cite{schwenk2022okvqa}}
            & \cb[c4]{POPE}{Acc. ~\cite{li2023popebenchmark}}
            & \cb[c4]{VQAv2}{Acc. (val)~\cite{goyal2017vqav2}}
            & \cb[c5]{\textit{\textbf{Avg. Cap.}}}{}
            & \cb[c5]{Flicker}{CIDEr~\cite{flickr}}
            & \cb[c5]{COCO}{CIDEr ~\cite{coco}}
            & \cb[c5]{No Cap}{CIDEr~\cite{agrawal2019nocaps}}
            & \cb[c6]{\textit{\textbf{Avg. Video}}}{}
            & \cb[c6]{VideoMME}{Acc.~\cite{fu2024videomme}}
            & \cb[c6]{STAR}{Acc.~\cite{wu2021star}}
            & \cb[c6]{TGIF-QA}{Acc.~\cite{jang2017tgif}}
            & \cb[c6]{EgoSchema}{(test) Acc.~\cite{mangalam2024egoschema}}
            & \cb[c6]{MVBench}{Acc.~\cite{li2024mvbench}}
            & \cb[c6]{PerceptionTest}{Acc. (test)~\cite{patraucean2024perceptiontest}}            \\
        \hline
        \addpadding
LLaVA-OV 7B~\cite{llava-onevision} & SigLIP-so400M & \ca{81.4} & 80.0 & 86.7 & 68.8 & 90.1 & \ca{79.9} & 77.3  & \textbf{69.6} & 89.2 & 83.5 & \ca{79.5} & 55.7 & 70.7 & 112.1 & \ca{63.8} & 57.7 & 66.0 & 77.2 & 65.2 & 57.1 & 58.1 \\
Gemma3 12B~\cite{gemma3} & SigLIP-so400M & \ca{-} & 75.7 & 87.1 & 64.9 & - & \ca{-} & 67.7 & - &  - & 71.6 & \ca{ -} & - & - & - & \ca{-} & - & - & - & - & - & 54.9   \\
Qwen2 VL 7B~\cite{qwen2vl} & DFN-H & \ca{86.6} & 83.6 & {94.5} & 76.5 & {91.7} & \ca{80.9} & 83.6 & 67.9 & 88.3 & 83.8 & \ca{93.7} & 79.9 & 102.5 & 98.7 & \ca{67.7} & {62.9} & 67.3 & 81.8 & 65.4 & 61.6 & 66.9 \\
InternVL 2.5 8B~\cite{chen2024internvit2p5} & InternViT 2.5-300M & \ca{87.0} & 84.6 & 93.0 & 77.6 & \textbf{92.8} & \ca{79.9} & 79.3 & {69.2} & {90.6} & 80.6 & \ca{113.0} & 96.5 & 125.8 & 116.7 & \ca{72.9} & 60.6 & 77.6 & 91.3 & 66.2 & 72.6 & 68.9   \\
InternVL 3 8B~\cite{internvl3} & InternViT 2.5-300M & \ca{87.2} & \textbf{86.6} & 92.7 & 76.8 & 92.6 & \ca{-} & 80.2 & - & \textbf{91.1} & - & \ca{-} & - & - & - & \ca{-} & \textbf{66.3} & - & - & - & 75.4 & -   \\
\textbf{PLM-8B} & \textbf{\PElang{G}} & \ca{\textbf{88.4}} & {85.5} & \bb{94.6} & \textbf{{80.9}} & {92.7} & \ca{\textbf{82.9}} & \textbf{86.5} & \textbf{69.6} & 89.9 & \textbf{85.6} & \ca{\textbf{127.4}} & \textbf{105.6} & \textbf{146.7} & \textbf{129.9} & \ca{\textbf{77.9}} & {58.3} & \textbf{84.9} & \textbf{95.5} & \textbf{{68.8}} & \textbf{77.1} & \textbf{82.7} \\
\shline
    \end{tabular}
   }
    \caption{{\bf MLLM System-Level Comparison.} 
    We show a system-level comparison between PLM-8B based on \PElang{G} and popular open-access models of similar LLM scale using existing encoders. 
    We report test set results where specified.
    }
    \vspace{-10pt}
    \label{tab:lang_mllm_system_level}
\end{table*}

\clearpage


\section{Perception Encoder: \textit{Spatial Alignment}}
\label{sec:sa}
While language alignment with a pretrained LLM decoder is well-established, the best way to spatially align a model is not obvious.
As shown in \S\ref{sec:layerfinder}, \PEcore{} already has features that perform well for spatial tasks. However, the layer that performs the best for higher level spatial tasks like detection or depth estimation (layer $\sim$40) is vastly different than the layer that performs the best for a pure spatial task like tracking (layer $\sim$30). While we were able to ignore this disparity during language alignment by aligning to an LLM decoder that could do all tasks, classical spatial tasks have decoders that come in all shapes and sizes. It would be impractical to simply align the model using all downstream decoders mirroring language alignment. Thus, we must first answer the question, what is happening in the features at those layers to make them useful for spatial tasks?

\subsection{Core Feature Analysis} \label{sec:core_feature_analysis}
\begin{wrapfigure}{r}{0.67\textwidth}
    \vspace{-35pt}
    \centering
  \begin{center}
      \begin{tabular}{c}
    \includegraphics[width=1\linewidth, trim=10.45in 0in 0in 15.1in, clip]{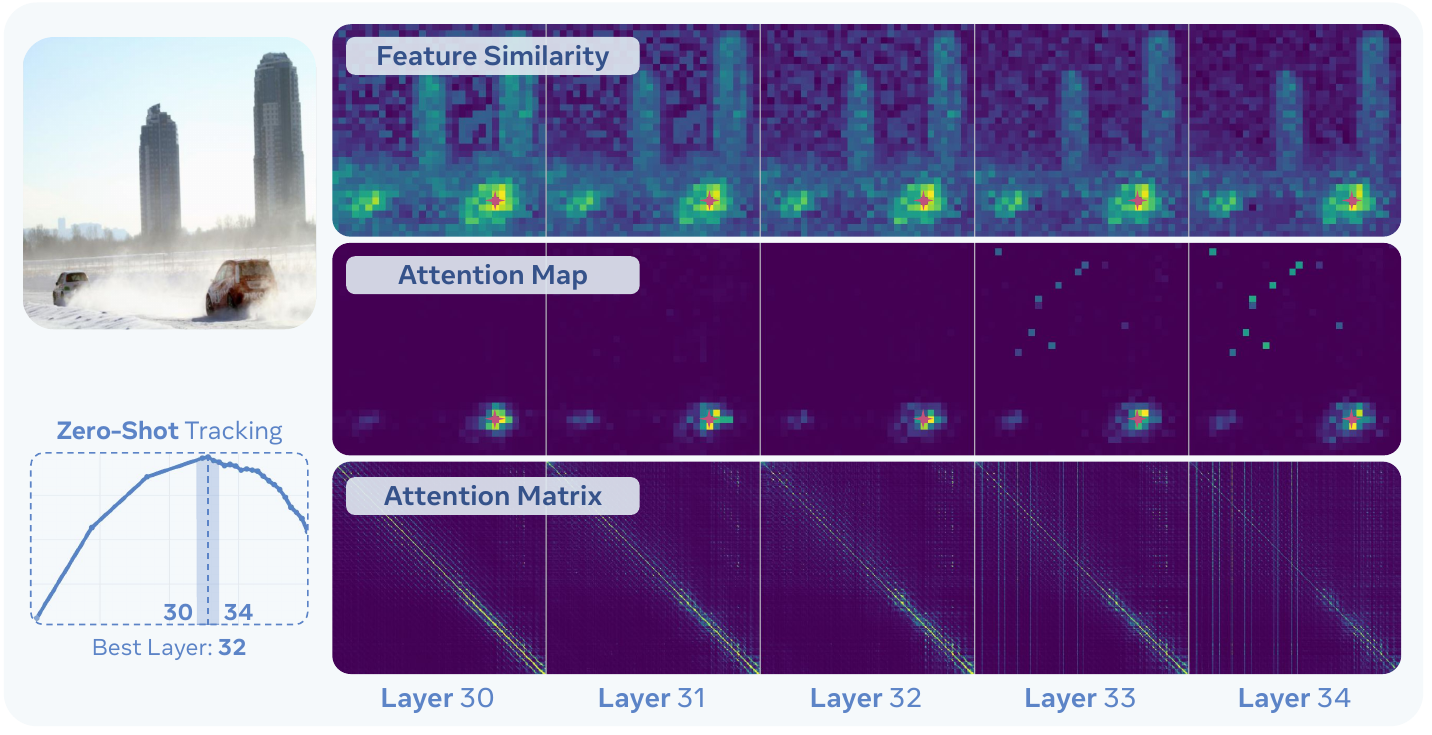}
    \end{tabular}
    \end{center}
    \caption{\textbf{\PEcore{}G Feature Analysis.} To understand the dichotomy between optimal \PEcore{} features for spatial tasks observed in Fig.~\ref{fig:layerfinder}, we analyze the spatial properties of the features between layers 30 and 34.}
    \label{fig:core_feature_example}
    \vspace{-15pt}
\end{wrapfigure}
We begin by analyzing the spatial properties of the features for \PEcore{}G in the range of layers where it performed optimally for zero-shot tracking in \S\ref{sec:layerfinder}. In Fig.~\ref{fig:core_feature_example}, we plot (1) the pairwise feature cosine similarity between the pink token and all others, (2) the head average attention map for that token, and (3) the full attention matrix ($HW\times HW$).

\paragraph{An 18 Layer Decoder.}
Remarkably, the cause for the tracking performance peak at layer 32 is abundantly clear from observing the visualizations. Up until layer 32, the attention maps remain local. However, that changes abruptly at layer 33, at which point several tokens in the background of the image become ``global'' tokens. As shown by the vertical lines in the full attention matrix, starting from layer 33 every token attends to them. Thus, every layer 33 and up become part of a \textit{decoder} for global information.

This is not a new phenomenon. Recent work~\cite{vitsneedregisters} shows this happening in all modern vision transformers above L scale.
But notably these ``global tokens'' are not necessarily harmful. Given the optimal layer for most tasks in Fig.~\ref{fig:layerfinder} lies within the global token region, the information they aggregate is useful downstream. However, tracking in \S\ref{sec:layerfinder} is zero-shot and relies purely on spatial correspondences, meaning it cannot make use of the global tokens. This explains why tracking peaks right before their introduction, while tasks that rely on semantic understanding or have larger decoders that can benefit from them do well with the later layers.

\subsection{Spatial Alignment Method}
\label{sec:sa_method}
Given the analysis in \S\ref{sec:core_feature_analysis}, we have two objectives in creating a spatial alignment method: (1) we must preserve the optimal \textit{semantic information} of the model (including the global tokens) that peaks around layer 40, and (2) we must do so while emphasizing \textit{local alignment} in service of spatial tasks with shallow decoders. The first can be easily achieved by aligning with the model's own features (e.g., with MaskFeat~\cite{maskfeat}), but the second is more challenging.
To accomplish this, we employ the Segment Anything Model (SAM) 2.1~\cite{sam2} in a novel way to enforce spatial correspondence information in PE.

\paragraph{Retaining Semantics.}
To retain the strong semantic features from \PEcore{}, we finetune the model with itself as a teacher. Specifically, we train the model to minimize the cosine similarity between its \textit{last layer} and the frozen layer 41 features of its initialization (a layer around the peak for many tasks in Fig.~\ref{fig:layerfinder}). On its own this would be a tautology, so we apply heavy regularization to the student: DropPath~\cite{droppath} and LayerScale~\cite{layerscale} similar to language alignment, as well as performing MaskFeat~\cite{maskfeat} with 75\% masking. We keep the teacher fixed in contrast to other state-of-the-art spatial models, which all employ an EMA teacher~\cite{dinov2,siglip2}. This could potentially help, but we opt for simplicity.

\begin{wrapfigure}{r}{0.48\textwidth}
    \vspace{-26pt}
    \centering
  \begin{center}
      \begin{tabular}{c}
    \includegraphics[width=1\linewidth, trim=13.44in 0in 0in 16.83in, clip]{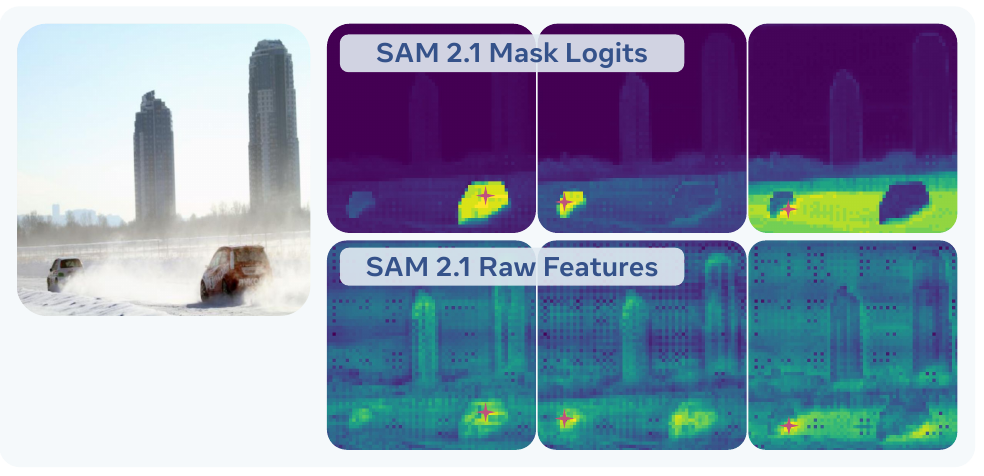}
    \end{tabular}
    \end{center}
    \caption{\textbf{SAM 2.1 Feature Similarity.} The cosine similarity between the pink marked token and all others for SAM 2.1-L~\cite{sam2} features \vs our proposed mask logit features.}
    \label{fig:sam_feature_example}
    \vspace{-10pt}
\end{wrapfigure}
\paragraph{Encouraging Locality.}
While we could ``retain'' locality by self-distilling from layer 32 features, that may be less effective as we are already distilling another layer of the model. Instead, we turn to a model that is explicitly tuned for locality: SAM~\cite{sam,sam2}. Notably, several works~\cite{ranzinger2023radio,shang2024theia,sariyildiz2024unic} have shown SAM to \textit{not} be an effective teacher when distilling from multiple sources (though recently \cite{heinrich2024radio2.5} has shown it can help with some tricks). However, upon observation of the raw features of SAM 2.1-L (Fig.~\ref{fig:sam_feature_example}), 
the main problem may be the same one we are currently trying to solve: \textit{SAM has global tokens as well}! In this case, they appear as dark spots in a grid-like arrangement across all examples in Fig.~\ref{fig:sam_feature_example} raw features.

Using the features of a model that itself has global tokens to mitigate the effect of global tokens is dubious at best. But, we don't have to use SAM's \textit{features} to learn locality.
At its core, SAM is a model that transforms points into spatially contiguous masks of select object.
If what we want is smooth, locally consistent features, we can use the \textit{mask predictions} themselves. Specifically, we query SAM 2.1-L with 1024 points arranged in a $32$\,$\times$\,$32$ grid.
For each point, SAM returns a $H$\,$\times$\,$W$ mask logit the size of the image, which it normally would threshold and NMS.
However, we instead concatenate those logits into a $H$\,$\times$\,$W$\,$\times$\,$1024$ tensor and use \textit{that} as the feature map for alignment. This explicitly produces locally well-aligned features compared to the underlying feature space and has no spatial artifacts caused by global tokens, as shown in Fig.~\ref{fig:sam_feature_example}.

Then to align, we distill the spatial correspondences between tokens by computing their pairwise cosine similarity for both the student and the teacher (creating a $HW$\,$\times$\,$HW$ matrix for each) and aligning them with MSE loss. Unlike SAM's underlying feature space (which~\cite{heinrich2024radio2.5} shows may be brittle to interpolation), the mask logit features are robust to interpolation, so we simply interpolate them down and train at the \PEcore{} model's original 448px resolution.
Finally, like for self-distillation we add the same masking and regularization. For both teachers, we apply loss to all tokens and add no extra parameters other than LayerScale.

\begin{wrapfigure}{r}{0.48\textwidth}
    \vspace{-18pt}
    \centering
    \includegraphics[width=1\linewidth, trim = 13.8in 0in 0in 14.05in, clip]{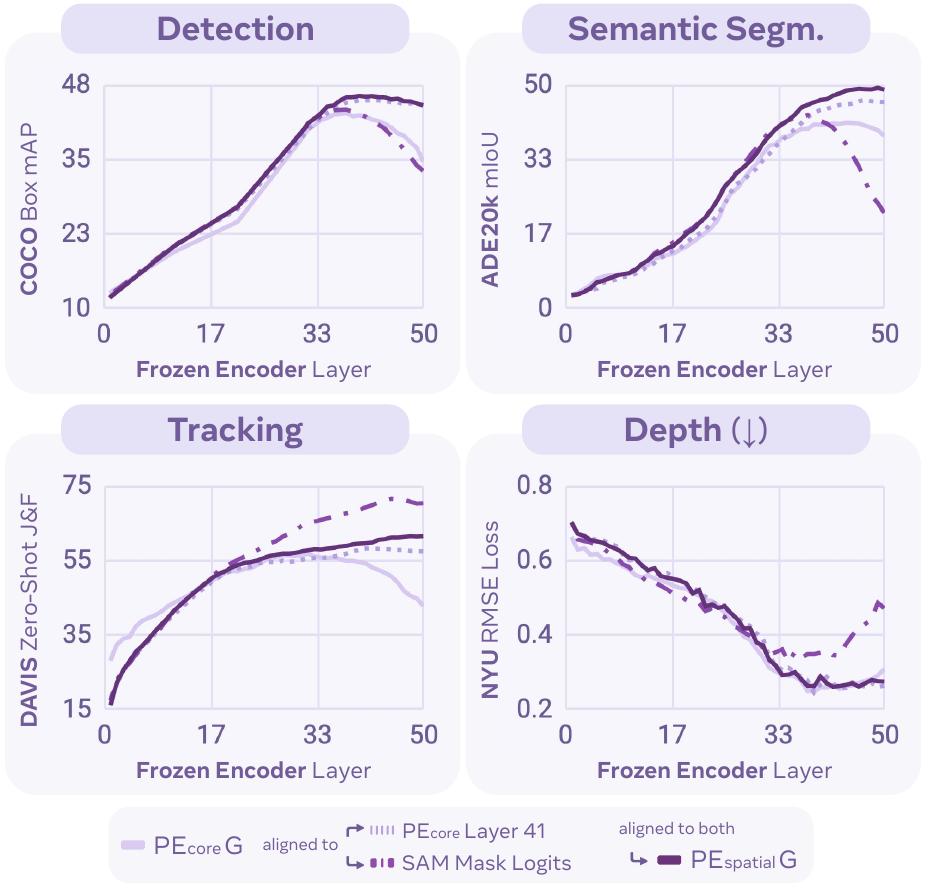}
    \caption{\textbf{Spatial Alignment.}
    We analyze how our two spatial alignment methods individually change the internal features of \PEcore{G}. Then we combine both alignment methods to create \PEspat{G} (see Appendix~\ref{appx:spatial_align_details}).
    }
    \label{fig:spatial_alignment}
    \vspace{-20pt}
\end{wrapfigure}

\paragraph{Effects.}
Again, the goal of alignment is to \textit{lift} the strong features already learned by the core model as shown in \S\ref{sec:layerfinder}. Thus, like we did for language alignment in \S\ref{sec:la_method}, we perform layerwise frozen feature analysis on spatial tasks in Fig.~\ref{fig:spatial_alignment}. This time, we evaluate the original \PEcore{G} checkpoint as well \PEcore{G} aligned to its own layer 41, to SAM 2.1 mask logits, and finally both. We denote aligning to both as \PEspat{G}.

\begin{wrapfigure}{r}{0.5\textwidth}
    \vspace{-10pt}
    \centering
    \includegraphics[width=1\linewidth, trim=12.4in 0in 0in 15.03in, clip]{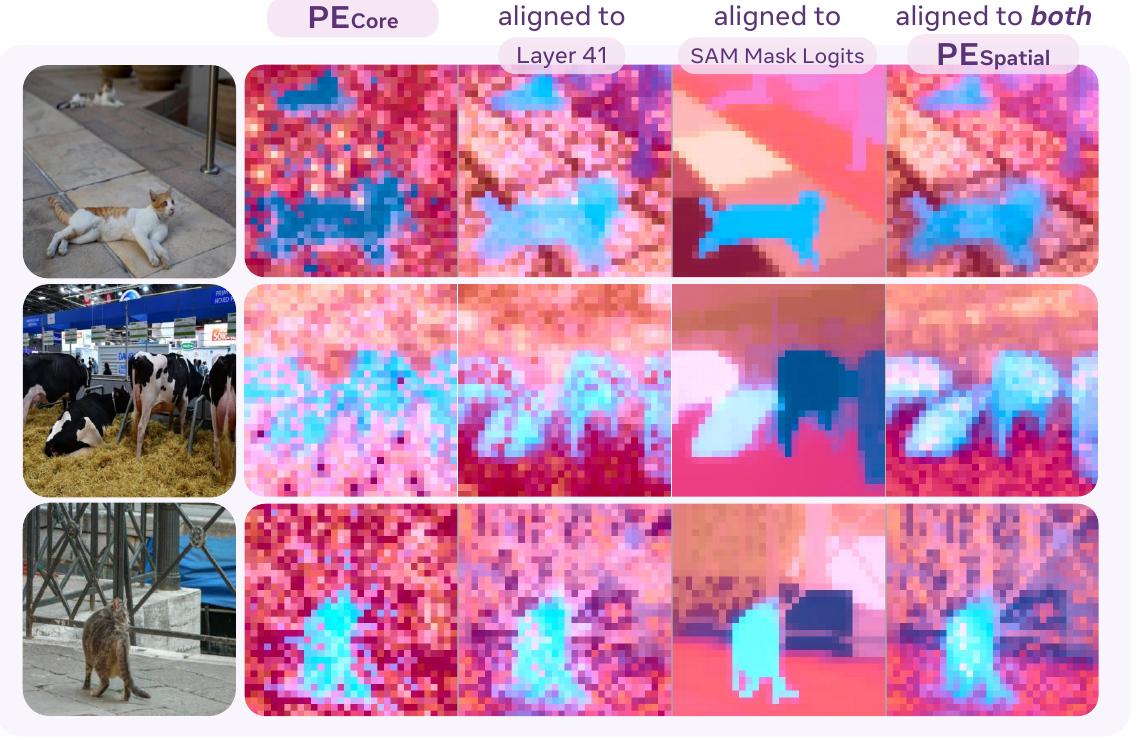}
    \caption{\textbf{Last Layer Visualization}
    for the models in Fig.~\ref{fig:spatial_alignment} using 3 dimensional PCA to map features to LCh color space (see Appendix~\ref{appx:feature_viz}). More examples in Appendix~\ref{appx:more_feature_viz}.
    }
    \label{fig:feature_viz}
    \vspace{-20pt}
\end{wrapfigure}

Aligning purely based on the original model's layer 41 features performs well on detection, depth, and semantic segmentation, but falls short for zero-shot tracking, where precise locality is necessary to define boundaries between objects. In contrast, aligning to SAM 2.1 mask logits lowers last layer performance on every task except for tracking, where it significantly improves performance. Understandably, this is because the mask logits have little semantics (see Fig.~\ref{fig:feature_viz}). Thus, the optimal approach is to combine both teachers. As a result, \PEspat{G} not only lifts the features for all tasks to the end of the network, but it also improves over self-alignment alone. Notably, \PEspat{G}'s tracking performance is lower than the SAM-aligned model, but it is still ahead of other methods while being a generally good model, see \S\ref{sec:sa_results}.

\paragraph{Last Layer Feature Visualization.}
In Fig.~\ref{fig:feature_viz}, we visualize the last layer features for the \PEcore{G} and the 3 aligned models, with similar colors denoting similar features. In the first column, we see why the last layer performance of \PEcore{} is so poor: while the last layer features contain information about the salient objects, they seem to have lost spatial coherence. Aligning to the model's own layer 41 features fixes this, but its spatial quality is lacking. In contrast, the model aligned to SAM 2.1 mask logits has locally clear features, but without semantics (similar objects have dissimilar features, see row 1 cats and row 2 cows). \PEspat{}, using both teachers at once, retains the semantics of \PEcore{} while producing high quality spatial features.

\begin{table}[h]
\vspace{10pt}
\begin{minipage}[t]{0.55\textwidth}
\centering
{
    \vspace{0pt} 
    \tablestyle{0pt}{1.1} 
    \newcommand{\loo}[2]{\ca{\tiny \textcolor{gray}{#1/#2}}}
    \begin{tabular}{y{47}x{25}x{36} x{16}x{16}x{16} x{16}x{16}x{16} x{16}x{16}x{16}}
        \shline
        &&& 
        \multicolumn{3}{c}{\ct[c3]{Tracking}} & 
        \multicolumn{3}{c}{\ct[c4]{Segmentation}} & 
        \multicolumn{3}{c}{\ct[c5]{Depth}} \\
        &&&
        \multicolumn{3}{c}{\ct[c3]{DAVIS ($\uparrow$)~\cite{davis2017}}} &
        \multicolumn{3}{c}{\ct[c4]{ADE20k ($\uparrow$)~\cite{ade20k}}} & 
        \multicolumn{3}{c}{\ct[c5]{NYU ($\downarrow$)~\cite{nyu_depth}}} \\
        Encoder & Params & Resolution &
        \cc[c3]{Best} & \cc[c3]{Last} & \cc[c3]{Idx} &
        \cc[c4]{Best} & \cc[c4]{Last} & \cc[c4]{Idx} &
        \cc[c5]{Best} & \cc[c5]{Last} & \cc[c5]{Idx} \\
        \hline
        \addpadding
        OAI CLIP-L~\cite{clip}       & 0.3B              & \rp{224}{14} &     39.4  &     37.1  & \loo{17}{24} &     39.4  &     38.3  & \loo{19}{24} &     .366  &     .397  & \loo{19}{24} \\
        AIMv2-3B~\cite{aimv2}        & 2.7B              & \rp{448}{14} &     54.7  &     29.3  & \loo{13}{24} &     41.6  &     31.9  & \loo{20}{24} &     .311  &     .326  & \loo{16}{24} \\
        SigLIP-so~\cite{siglip}      & 0.4B              & \rp{384}{14} &     48.7  &     36.3  & \loo{16}{27} &     40.1  &     38.3  & \loo{22}{27} &     .339  &     .369  & \loo{21}{27} \\
        SigLIP2-so~\cite{siglip2}    & 0.4B              & \rp{512}{16} &     51.4  &     45.3  & \loo{15}{27} &     44.0  &     42.9  & \loo{24}{27} &     .306  &     .329  & \loo{25}{27} \\
        SigLIP2-g-opt~\cite{siglip2} & 1.1B              & \rp{384}{16} &     43.5  &     38.8  & \loo{32}{40} &     42.1  &     41.3  & \loo{34}{40} &     .302  &     .324  & \loo{34}{40} \\
        DINOv2-L~\cite{dinov2}       & 0.3B              & \rp{448}{14} & \uu{58.7} &     58.2  & \loo{23}{24} &     47.3  &     47.3  & \loo{24}{24} &     .297  &     .308  & \loo{23}{24} \\
        DINOv2-g~\cite{dinov2}       & 1.1B              & \rp{448}{14} &     58.5  & \uu{58.5} & \loo{40}{40} & \uu{48.7} & \uu{48.4} & \loo{37}{40} &     .279  & \uu{.290} & \loo{27}{40} \\
        \textbf{\PEcore{G}}          & 1.9B              & \rp{448}{14} &     56.8  &     42.8  & \loo{32}{50} &     41.5  &     38.6  & \loo{44}{50} & \bb{.249} &     .309  & \loo{39}{50} \\
        \textbf{\PEspat{G}}          & 1.9B              & \rp{448}{14} & \bb{61.5} & \bb{61.5} & \loo{50}{50} & \bb{49.3} & \bb{48.9} & \loo{49}{50} & \uu{.262} & \bb{.275} & \loo{46}{50} \\
        \shline
    \end{tabular}
}
\caption{\textbf{Frozen Feature Dense Prediction} including zero-shot tracking, semantic segmentation and depth estimation. We report best and last layer performance, along with which layer was best for each model. See Appendix~\ref{appx:dense_pred} for experimental settings.}
\label{tbl:dense_prediction}
\end{minipage}%
\hfill%
\begin{minipage}[t]{0.42\textwidth}
\centering
{
    \vspace{0pt} 
    \tablestyle{0pt}{1.1} 
    \begin{tabular}{y{47}x{25}x{30}x{24}x{24}x{24}x{24}}
        \shline
        &&\multirow{2}{*}{\parbox{1.0cm}{\centering Pretrain Resolution}}& \multicolumn{2}{c}{\ct[c6]{ LVIS~\cite{lvis}}} & \multicolumn{2}{c}{\ct[c7]{COCO~\cite{coco}}} \\
        Encoder & Params &  
        & \ct[c6]{$\text{AP}_\text{box}$}{}
        & \ct[c6]{$\text{AP}_\text{mask}$}{}
        & \ct[c7]{$\text{AP}_\text{box}$}{}
        & \ct[c7]{$\text{AP}_\text{mask}$}{}\\
        \hline
        \addpadding
        OAI CLIP-L~\cite{clip}  & 0.3B & \rp{224}{14} &45.0 & 41.9 & 54.0 & 47.5 \\
        MetaCLIP-G~\cite{metaclip} & 1.8B & \rp{224}{14} & 45.1 & 41.9 & 53.2 & 46.7   \\
        SigLIP-so~\cite{siglip} & 0.4B & \rp{224}{14}&  45.0 & 41.9 & 54.4 & 47.6 \\
        MAE-L~\cite{mae} & 0.3B	& \rp{224}{14} & 46.1 & 43.9 & 55.6 & 49.3 \\
        EVA02-L~\cite{eva2}  & 0.3B &  \rp{224}{14} & 49.3 & 45.2 & 54.9 & 48.2 \\
        SigLIP2-so~\cite{siglip2} & 0.4B & \rp{512}{16} & 49.3 & 45.6 & 56.0 & 49.4 \\
        SigLIP2-g-opt~\cite{siglip2} & 1.1B & \rp{384}{16} & {52.9} & {48.5} & 57.1 & {50.2} \\
        DINOv2-L~\cite{dinov2} & 0.3B & \rp{518}{14} & 46.7 & 43.5 & 55.7 & 49.0 \\
        DINOv2-g~\cite{dinov2}  & 1.1B & \rp{518}{14} & 51.5 & 47.3 & {57.2} & 50.0 \\
        \textbf{\PEcore{G}} & 1.9B & \rp{448}{14} &  51.9 & 47.9 & 57.0 & 49.8 \\
        \textbf{\PEspat{G}} & 1.9B & \rp{448}{14} & \textbf{54.2} & \textbf{49.3} & \textbf{57.8} & \textbf{50.3} \\
        \shline
    \end{tabular}
}
\caption{\textbf{End-to-End Finetuning Detection and Segmentation} using Mask R-CNN~\cite{maskrcnn} and VitDet~\cite{vitdet} in a controlled setting. Details in Appendix~\ref{appx:det}.}
\label{tbl:det_SFT}
\end{minipage}
\vspace{-5pt}
\end{table}

\subsection{Comparisons with Existing Vision Encoders}
\label{sec:sa_results}

\paragraph{Frozen Feature Dense Prediction.} In Tab.~\ref{tbl:dense_prediction}, we compare different vision encoder's frozen features on three dense prediction tasks: DAVIS tracking~\cite{davis2017} (J\&F) following the training-free setting from~\cite{jabri2020space,vgpt}, ADE20k semantic segmentation~\cite{ade20k} (mIoU) linear probing, and NYU depth estimation~\cite{nyu_depth} (RMSE) with a DPT head~\cite{dpt}.
For each model, we report both its best layer and last layer performance. 
Across the board, \PEspat{} performs outperforms other state-of-the-art spatial models, with its best features being much better aligned to the last layer than the \PEcore{} it started from. Notably, SigLIP2, which during pretraining combines spatial, captioning, and contrastive losses~\cite{siglip2} is \textit{not} aligned well to the last layer in comparison.

\paragraph{End-to-End Finetuning Detection and Segmentation.}
In Tab.~\ref{tbl:det_SFT}, we compare \PEcore{} and \PEspat{} with other popular vision encoders in the standard \textit{full-finetuning} ViTDet~\cite{vitdet} Mask-RCNN~\cite{maskrcnn} setting using COCO~\cite{coco} and LVIS~\cite{lvis} as benchmarks. In this controlled experiment, \PEspat{} is state-of-the-art among various vision backbones.
This is significant, as contrastive encoders (especially large ones like MetaCLIP-G~\cite{metaclip}) usually perform very poorly on detection, with smaller models often performing better. Typically, encoders only scale for detection if using spatial pretraining or a significant amount of detection data~\cite{dinov2} is used to align them directly to downstream tasks. In contrast, \PEspat{} \textit{uses no detection data for alignment}, making it general.

\begin{wraptable}{r}{0.4\textwidth}
\vspace{-15pt}
\centering
{
    \tablestyle{0pt}{1.1} 
    \begin{tabular}{y{60}x{35}x{35}x{45}}
        \shline
        Encoder  & Params & Detector
        & \ct[c7]{COCO AP$_\text{box}$}{}\\
        \hline
        \addpadding{}%
        SwinV2-G~\cite{swin2} & 3.0B & HTC++~\cite{htc}	 & 62.5 \\
        Swin-L~\cite{swin}& 0.3B & DINO~\cite{dino_det}	 & 63.2 \\
        EVA02-L~\cite{eva2} & 0.3B & Cascade~\cite{cascadercnn} & 64.1 \\
        InternImage-G~\cite{internimage} & 3.0B & DINO~\cite{dino_det}	 & {65.3} \\
        EVA02-L~\cite{eva2} & 0.3B & CoDETR~\cite{codetr} & 65.9 \\
        \textbf{\PEspat{G}} & 1.9B & DETA~\cite{deta} & \textbf{66.0} \\
        \shline
    \end{tabular}
}
\caption{{\bf System-Level Comparison on Detection.} Comparing to the leading results on COCO~\cite{coco} val2017. See Appendix~\ref{appx:det_sota_setting} for training recipe.
}
\label{tbl:det_coco}
\vspace{-20pt}
\end{wraptable}

\paragraph{System-Level Detection.}
In Tab.~\ref{tbl:det_coco}, we provide a system-level end-to-end finetuning comparison \vs the absolute state-of-the-art in COCO detection. With only Object365~\cite{o365} as extra detection data, \PEspat{} can match the performance of more complex models tuned for detection, while only using a simple DETR-style decoder~\cite{carion2020detr,deta}. \PEspat{} marks the first general, contrastively pretrained model to accomplish this.

\clearpage


\newpage
\section{Related Work}

Learning vision-semantic representations has long been the leading approach for developing foundational models in perception. By aligning visual and textual representations, these models excel not only in vision tasks such as zero-shot image classification and image-text retrieval~\cite{clip,openclip,laion}, open-vocabulary detection~\cite{owlv1, fvlm, owlv2} and segmentation~\cite{ding2023zss, cho2024catseg}, but also serve as the basis for multi-modal large language models (MLLMs)~\cite{qwen-vl, kosmos-2, llava, paligemma, mm1, cambrian}.

\paragraph{Contrastive Language-Image Pretraining.} The early works of Virtex~\cite{desai2021virtex}, ICMLM~\cite{sariyildiz2020icmlm}, and ConVIRT~\cite{pmlr-v182-zhang22a} developed the techniques for learning through contrastive objectives between vision and language modalities. Subsequently, vision encoders such as CLIP~\cite{clip,openclip} and ALIGN~\cite{align} scaled these techniques to much larger datasets and model sizes, popularizing vision-language contrastive learning. A series of open-weight contrastive models have been developed to enhance the performance and robustness of CLIP~\cite{EVA-CLIP, siglip, li2023clipav2, dfn, metaclip, laion}. For instance, SigLIP~\cite{siglip} replaces the traditional softmax with a sigmoid function in contrastive learning, while FLIP~\cite{flip} employs masking techniques to expedite the training process. We are among this effort and build a state-of-the-art open Perception Encoder (PE) (\S\ref{sec:core_image_pt}). Other objectives that have proven useful for building visual encoders include captioning loss, which learns to predict image descriptions using a language model decoder and transfers well to downstream multi-modal language modeling tasks~\cite{aimv2,cappa}. Many works are now attempting to combine two or more objectives to address different downstream tasks through pretraining with multiple objectives~\cite{aimv2, coca} or training sequentially~\cite{internvl, llava-onevision}.

\paragraph{Efficient Training.} Various axes of efficient training of clip models have been explored. BASIC~\cite{basic} and LAION~\cite{laion} explored scaling the batch size up to 160K, and shows the benefits of large batch sizes during training. EVA-CLIP~\cite{eva18b} uses LAMB optimizer~\cite{lamb} for large batch training of clip models. Rotary positional embedding (RoPE) ~\cite{rope} has been successfully adopted in large language models. In vision transformers ~\cite{heo2024rotary, agrawal2024pixtral} adopted 2D rotatory positional embeddings. For data engine, a series of works focus on large-scale sourcing and filtering through efficient data curation~\cite{datacomp, laion, dfn, metaclip} and explore recaptioning training images using MLLMs or VLMs~\cite{rewrite, veclip, Nguyen2023recap, altogether}. We extend these concepts to build a video data engine and scale our model to function as one strong model for both image and video (\S\ref{sec:core_video_ft}).

\paragraph{Best Embedding Layer Inside the Network.} Typically, most vision encoders rely on the last layer to extract features for the task it is trained on. However, when trained on proxy or self-supervised tasks, the last layer is often not the ideal candidate for other tasks~\cite{zheng2016good,bordes2022guillotine,shekhar2023objectives,igpt,aimv1,vgpt,repa,ma2024sit,sun2024cliper,walmer2023teaching,chen2020simclr}. For example, when using image colorization as pretraining objective, \cite{zhang2016colorful,zheng2016good} showed that the middle layers were better at image classification compared to last layers. Subsequently, in iGPT~\cite{igpt}, when trained for next token prediction, intermediate layers performed better at image classification. AIMv1~\cite{aimv1} also showed similar behavior for image based next token prediction with patch normalized MSE loss. Toto~\cite{vgpt} showed this can be extended for next token prediction in videos, and intermediate layers are best for image classification, video classification, tracking and robotics. REPA~\cite{repa} showed this behavior for image generation models, where the intermediate layers of SiT~\cite{ma2024sit} has better linear probing accuracy compared to earlier or later layers. In CLIP models, CLIPer~\cite{sun2024cliper} identified that early layers in CLIP possess good spatial understanding. In contrast to these lines of work, in this paper, we first show this behavior is not limited to one class of encoders. Specifically, we show this behavior exists in a spatially self-supervised model~\cite{dinov2}, generative captioning model~\cite{aimv2}, and also in our own PE. Then we study this behavior for PE encoder in depth, and show it is possible for CLIP training to produce rich spatial and semantic features in intermediate layers (\S\ref{sec:layerfinder}).

\paragraph{Alignment Tuning.} We explore alignment tuning for language (\S\ref{sec:la}) and for spatial understanding (\S\ref{sec:sa}). 
For language alignment, we focus on adapting to multimodal large language models (MLLMs); for spatial alignment, we employ self-distillation of the models own features combined with a teacher for locality. 
In MLLM literature, \emph{midtraining}---i.e., a middle stage of training used to exploit large-scale multimodal data---has been actively studied. 
LLaVA-OneVision~\cite{llava-onevision}, InternVL series~\cite{internvl,chen2024internvit2p5}, QwenVL series~\cite{qwen-vl,qwen2vl}, and several other leading MLLMs~\cite{llama3,gemma3} adopt this paradigm. 
Our \PElang{} can be seen as a variant of midtraining, but with one critical difference in principle:
our goal is \emph{not} to build the best MLLM, but to make the vision encoder the most \textit{general}. 
Throughout \S\ref{sec:la}, we benchmark our \PElang{} across different language models, input resolution, on various tasks for image and video to show this generality.
For spatial tasks, we utilize the hidden embeddings in the intermediate layers. 
Recently, several works showed the effectiveness of distilling teacher model via representation alignment with cosine similarity.
REPA~\cite{repa} distilled an early layer features of DINO for image diffusion models,
RADIO~\cite{ranzinger2023radio} used multi-teacher distillation (DINO, CLIP and SAM). 
The key idea is to borrow semantic understanding  (\textit{e.g.}, CLIP) and spatial understanding (\textit{e.g.}, SAM, DINO) of a pretrained vision encoders. 
In our \PEspat{}, we exploit the intermediate features of \PEcore{} for semantics, and a novel way to use SAM for spatial understanding.

   
\section{Conclusion}
We have presented Perception Encoders (PE), a family of best-in-class foundation models comprising \PEcore{}, \PElang{}, and \PEspat{}.
We have shown that \PEcore{} can outperform models trained with WebLI and JFT-3B, which were previously the undisputed leaders in zero-shot image recognition, while also excelling in zero-shot video recognition.
We have demonstrated that \PElang{} can be used to build a multimodal language model \cite{PLM} that is at the forefront of the field in terms of performance.
We have established that \PEspat{} can match the long-standing state-of-the-art in object detection with a significantly simpler decoder.
Throughout all of this, one conclusion is abundantly clear:
Perception Encoder unlocks the potential to scale simple contrastive vision-language pretraining to address a wide range of downstream vision tasks.

\paragraph{Additional Contributors and Acknowledgments.}
We would like to thank Abhimanyu Dubey, Adel Ahmadyan, Andrew Westbury, Arkabandhu Chowdhury, Azita Shokrpour, Babak Damavandi, Chay Ryali, Cyprien de Lichy, Didac Suris Coll-Vinent, Dong Wang, Filip Radenovic, George Orlin, Han Zou, Harry Tran, Jitendra Malik, Joelle Pineau, Joseph Greer, Kavya Srinet, Kirmani Ahmed, Laura Gustafson, Lu Zhang, Muhammad Maaz, Natalia Neverova, Nicolas Carion, Oleksandr Maksymets, Ramya Raghavendra, Romy Luo, Ronghang Hu, Sam Doud, Sasha Mitts, Sean Bell, Shane Moon, Shuming Hu, Soerian Lieve, Stephane Kasriel, Valentin Gabeur, Vanessa Stark, Vignesh Ramanathan, Vivian Lee, Xuan Hu, Yang Li, and  Ziyang Wang for their contributions and support for the project.
And we thank you, the reader, for reading this far.

\clearpage

\appendix

\section{Video Data Engine}

\subsection{Video Caption}
\label{sec:appx_video_caption}
\paragraph{LLM Summarization prompt}

\promptbox{LLM Summarization prompt 72 tokens}{Create a concise caption of a video using the provided metadata, video caption, and frame captions.

TASK: Extract key information from the captions and combine it into an alt text format using single phrase or set of phrases that includes all relevant details.

Steps to Follow:

1. Review the metadata (title and description) for general context, you can rely it for entity names but do not rely on it as the primary source of information for your caption. 

2. Blend title / description with video caption and frame captions for the main storyline

3. Extract the most relevant and concise information.

4. Combine extracted information into a alt text format using short phrase or set of phrases with approximately 120 tokens, considering special characters like comma as part of the token count.

5. Prioritize including all key information over sentence structure or grammar.

6. Minimize the use of special characters and focus of key information.

What to Avoid:

- Avoid adding or inferring information not present in the original metadata and captions.

- Avoid using complex sentence structures or prioritizing sentence flow.

Create a concise caption of the video based on the metadata, video caption, and frame captions. }

\subsection{PE Video Dataset Details}
\label{sec:appx_video_datasets}

PE Video is a dataset that we collected and curated from a licensed data source. The videos are high-resolution and high-quality with a focus on motion. The total number of videos is 1M. Among these, 120K videos have human-refined video captions, and we selected 15K from the 120K videos as a benchmark.

\subsubsection{Video Data Filtering Pipeline}
The goal of video data filtering is to identify videos that contain motions such as object motion, camera motion, interaction between objects, human actions, sequences of actions, and manipulation of objects, while rejecting videos with static scenes, like landscapes, or those that are artificial or highly edited.

To achieve this, we created a video filtering pipeline consisting of the following steps:

\paragraph{Step 1}: Compute motion features. For each video, we compute a list of features from video frames, including frames per second (fps), number of frames, number of I-frames, motion vector magnitude, and motion vector variance, using off-the-shelf tools like OpenCV~\cite{opencv}.

\paragraph{Step 2}: Extract video frame features. For each video, we uniformly sample three frames and encode them using a DINOv2 model~\cite{dinov2} and a SigLIP model~\cite{siglip}.

\paragraph{Step 3}: LLM Features. For each video, we also run a multimodal large language model (LLM) like Llava-Onevision QwenLM 2 0.5B~\cite{llava-onevision} to extract MLLM features. We composed a list of 26 questions and performed MLLM inference on the videos. The questions can be found here in \S\ref{text:llm_feature_extraction_qeustions}.

\paragraph{Step 4}: Video Quality Scoring. We combine all the features collected so far and use a random forest model to predict a score between 0 and 5. To train the model, we manually annotated approximately 1,000 videos with scores between 0 and 5. A low score indicates that the video is almost static and can be nearly summarized by a single frame, while a high score indicates that there are multiple temporal events in the video, requiring several frames to accurately caption it. We use these annotated videos as training data to fit a random forest model for video quality score prediction.

\paragraph{Step 5}: We apply k-means clustering to the videos and rank them within each cluster. By selecting the top-ranked videos from each cluster, we effectively reduce the number of duplicated videos in the final dataset.

\newpage
\subsubsection{LLM Feature Extraction}
\label{text:llm_feature_extraction_qeustions}

\promptbox{LLM Feature extraction question list}{
Is the camera capturing the scene static? Reply yes or no.

Is the camera capturing the scene moving? Reply yes or no.

Is the video capturing a landscape? Reply yes or no.

Is the video capturing a static scene? Reply yes or no.

Is the scene captured from a distance? Reply yes or no.

Is the video captured with a drone? Reply yes or no.

Is the video computer-generated? Reply yes or no.

Is the video content abstract? Reply yes or no.

Is there something moving through the scene? Reply yes or no.

Is there someone doing something in the video? Reply yes or no.

Are there several things moving in the video? Reply yes or no.

Is there an object that is being manipulated? Reply yes or no.

Are there animals in the video? Reply yes or no.

Is the scene mostly static? Reply yes or no.

Are things occluding each other in this video? Reply yes or no.

Is there something obstructing the view apart from the watermark? Reply yes or no.

Is there a large number of things in the video? Reply yes or no.

Are there more than 5 different objects in the video? Reply yes or no.

Is it hard to keep track of some entities because they are moving so much? Reply yes or no.

Is someone looking at a phone, a tablet or a computer screen? Reply yes or no.

Are they looking at a phone, a tablet or a computer screen during the whole video? Reply yes or no.

Are there several moving persons in this video? Reply yes or no.

Are there several moving animals in this video? Reply yes or no.

Are there several objects in this video? Reply yes or no.

Are there several similar-looking objects in the video? Reply yes or no.

Do they look similar? Reply yes or no.
}

We use LLaVA-OneVision~\cite{llava} model to extract LLM features from the videos. For each video, we prompt with 26 different questions to extract features ranging from, ``is the video a landscape video?'' to, ``are there any moving objects in the video?'' The features are then used by a random forest model to determine the video quality score.

\subsubsection{PVD Benchmark Distribution}
\label{appx:pvd_bench_distribution}

\begin{table}[!h]
\centering
\tablestyle{4pt}{1.05} 
\begin{tabular}{x{45}x{40}x{40}}
    \shline
    \addpadding
    \multirow{2}{*}{Category} & Number of videos & Avg. Caption Length\\
    
    \hline
    \addpadding
    \cc[c1]{Hand Actions} & 2143 & 54.2 \\
    \cc[c2]{Object Interactions} & 1864 & 42.6 \\
    \cc[c3]{Food Preparation} & 1691 & 56.8 \\
    \cc[c4]{Work Activities} & 1689 & 47.8 \\
    \cc[c5]{Outdoor Scenes} & 1558 & 50.7 \\
    \cc[c6]{Animals} & 1423 & 50.9 \\
    \cc[c7]{Water Scenes} & 1337 & 44.6 \\
    \cc[c8]{Object Handling} & 1307 & 51.6 \\
    \cc[c9]{Close-up Shots} & 1122 & 45.1 \\
    \cc[c10]{Nature Scenes} & 866 & 38.4 \\
    \shline
\end{tabular}

\caption{{\bf PVD} Benchmark Statistics. We created a dataset of 15K videos together with human-verified captions. The videos are motion-centered, covering both first-person and third-person views with a wide coverage of scenes. \label{tab:app_PEvideo_distribution}} 

\end{table}

\begin{figure*}[t!]
    \centering
    \vspace{1cm}
    \includegraphics[width=\linewidth, trim=10.5in 0in 0in 10in, clip]{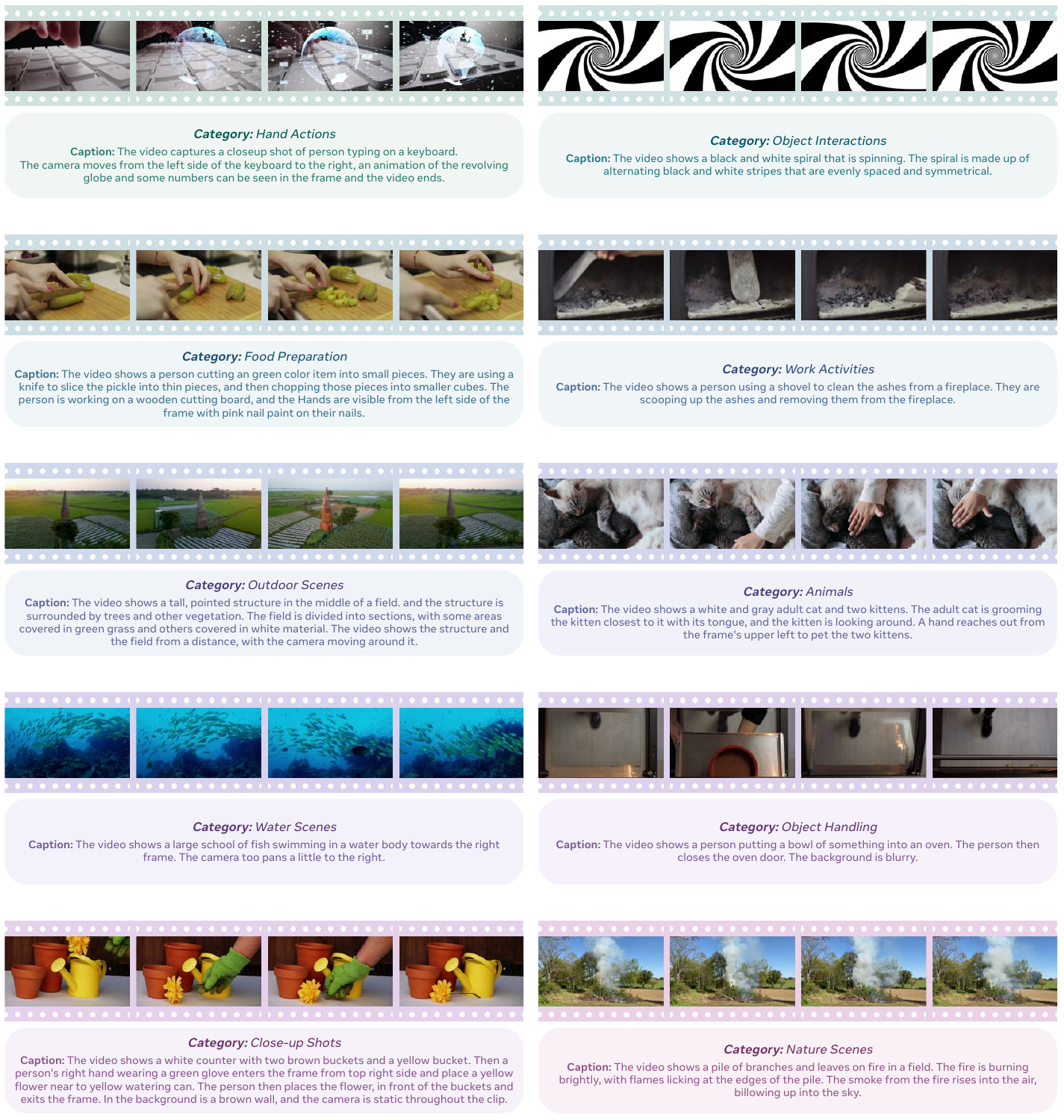}
    \caption{{\bf More PE Video Dataset Examples.} For each of the ten categories, we randomly pick one video and show its video caption. The captions were generated by our video data pipeline and then refined by human annotators.}
    \label{fig:video_data_example_more}
    \vspace{2cm}
\end{figure*}

\clearpage

\section{Implementation Details}

\subsection{PE Core}
We provide additional implementation details for building \PEcore{}. Our implementation is based on OpenCLIP\footnote{\url{https://github.com/mlfoundations/open\_clip}}.

\subsubsection{Architecture and Training Setups}
\paragraph{Model Architecture.} Following CLIP, \PEcore{} comprises a Transformer-based~\cite{transformer} vision and a text encoder. We employ customized Transformer configurations as detailed in Tab.~\ref{tab:app_model_arch}. For pooling, we an attention pooling block in the style of SigLIP~\cite{siglip} \textit{with 8 heads} from the last-layer feature to construct image and video embeddings. Regarding positional embedding, we use 2D RoPE~\cite{rope} for relative positional embeddings and 2D learnable absolute positional embeddings (abs) the same size as the model's input resolution. We interpolate positional embeddings to enable support for various resolutions beyond the default. The text context length is 72 for G-scale and 32 for B and L-scale models. Originally a bug, we find it optimal to \textit{not disable the class token} when using attention pooling for smaller models. Thus, the B and L models use a class token, then the attention pooling layer probes all features at once (class token included). Finally, we use an input mean and standard deviation of $(0.5, 0.5, 0.5)$ for simplicity.

\begin{table}[h!]
\centering
\tablestyle{0pt}{1.05} 
\begin{tabular}{x{20}x{30}x{25}x{20}x{20}x{20}x{20}x{20}x{40}x{40}x{40}x{40}x{40}x{40}}
    \shline
        \ct{Scale} & \ct{Tower} & \ct[c1]{Params} & \ct[c2]{Width} & \ct[c3]{Depth} & \ct[c4]{MLP} & \ct[c5]{Heads} & \ct[c6]{CLIP Dim} & \ct[c7]{Pooling} & \ct[c8]{Positional Embedding} & \ct[c9]{Resolution \& Context Len}  & \ct[c10]{Patch Size} & \ct[c10]{Class Token Register}\\
        \hline
       \addpadding
        \multirow{2}{*}{B} & Vision & 0.09B & 768 & 12 & 3072 & 12 & \multirow{2}{*}{1024} & Attn Pool & RoPE+Abs & 224 & 16 & \cmark \\
                           & Text   & 0.31B & 1024 & 24 & 4096 & 16 & &  EOS Token & Abs & 32 & - & - \\
       \hline
       \addpadding
        \multirow{2}{*}{L} & Vision & 0.32B& 1024 & 24 & 4096 & 16 & \multirow{2}{*}{1024}& Attn Pool & RoPE+Abs & 336 & 14 & \cmark \\
                           & Text   & 0.31B & 1024 & 24 & 4096 & 16 & &  EOS Token & Abs & 32 & - & - \\
                           \hline
       \addpadding
        \multirow{2}{*}{G} & Vision & 1.88B & 1536 & 50 & 8960 & 16 & \multirow{2}{*}{1280} & Attn Pool & RoPE+Abs & 448 & 14 & \xmark \\
                           & Text   & 0.47B & 1280 & 24 & 5120 & 20 & &  EOS Token & Abs & 72 & - & - \\
    \shline
\end{tabular}
\captionsetup{justification=centering}
\caption{{\bf PE} Model Configurations with full details.} 
\label{tab:app_model_arch}
\end{table}

\paragraph{PE Core Training.}
\label{sec:appx_joint_train}
As discussed in \S\ref{sec:unified-encoder}, the training of \PEcore{} involves three stages: 1) image pretraining; 2) image and video finetuning; and 3) an additional model distillation for smaller models. These three stages work together to develop a robust and effective \PEcore{} model.

We first provide training recipes for 1) image pretraining in Tab.~\ref{tbl_app:pretrain} and 2) video finetuning in Tab.~\ref{tbl_app:video_ft}.

\label{sec:appx_image_pretrain}

\begin{table}[!h]
\begin{minipage}[t]{0.33\textwidth}
\vspace{0pt}
\centering
{
\tablestyle{5pt}{1.1}
\begin{tabular}{l | c }
config  & values \\
\hline
\addpadding
optimizer & LAMB \\ 
$\beta_1, \beta_2$ & (0.9, 0.95)\\
weight decay & 0.05 \\
learning rate & 2e-3\\
batch size & 131,072 \\
warm-up steps & 2K \\
training steps & 443K (B, L) / 656K (G) \\
data quantity & 5.4B \\
samples seen & 58B (B, L) / 86B (G) \\
max logit scale & 100 \\\\
mask reg ratio & 0.4\\
mask reg batch & 8192\\
\\
progressive res & \makecell{112-160-224 (B) \\ 98-154-224-336 (L) \\ 98-154-224-336-448 (G)} \\
\\
data aug & %
    \makecell{
        \text{aspect jitter} \text{\texttt{\tiny ar(0.75,1.33)}} \\
        \text{rand crop} \text{\texttt{\tiny s(0.08,1)}} \\
        \text{color jitter} \text{\texttt{\tiny j(0.32,0,0.32,0)}}\\
        \text{hflip} \text{\texttt{\tiny p(0.5)}}
    }\\
\end{tabular} \\
}
\captionsetup{justification=centering}
\caption{\bf Image Pretraining.}
\label{tbl_app:pretrain}
\end{minipage}
\hfill
\begin{minipage}[t]{0.33\textwidth}
\vspace{0pt}
\centering
{
\tablestyle{5pt}{1.1}
\begin{tabular}{l | c }
config  & values \\
\hline
\addpadding
optimizer & LAMB \\ 
$\beta_1, \beta_2$ & (0.9, 0.95)\\
weight decay & 0.05 \\
learning rate &  1e-6 \\
batch size & 4096\\
warm-up steps & 2K \\
training steps & 5.4K \\
data quantity & 22M \\
samples seen & 22M \\
max logit scale & 100 \\\\
number of frames & 8 \\\\
data aug & %
    \makecell{
        \text{aspect jitter} \text{\texttt{\tiny ar(0.75,1.33)}} \\
        \text{rand crop} \text{\texttt{\tiny s(0.08,1)}} \\
        \text{color jitter} \text{\texttt{\tiny j(0.32,0,0.32,0)}}\\
        \text{hflip} \text{\texttt{\tiny p(0.5)}}
    }\\
\end{tabular} \\
}
\captionsetup{justification=centering}
\caption{\bf Video Finetuning.}
\label{tbl_app:video_ft}
\end{minipage}
\hfill
\begin{minipage}[t]{0.3\textwidth}
\vspace{0pt}
\centering
{
\tablestyle{5pt}{1.1}
\begin{tabular}{l | c }
config  & values \\
\hline
\addpadding
optimizer & LAMB \\ 
$\beta_1, \beta_2$ & (0.9, 0.95)\\
weight decay & 0.05 \\
learning rate &  1e-6 \\
batch size & 16384 \\
warm-up steps & 2K \\
training steps & 269K \\
data quantity & 5.4B \\
samples seen & 4.4B \\
max logit scale & 100 \\\\
teacher logit scale & 200 (\S\ref{appx:core_smaller_models}) \\\\
data aug & None \\
\end{tabular} \\
}
\captionsetup{justification=centering}
\caption{\bf Distillation.}
\label{tbl_app:distillation}
\end{minipage}
\end{table}

After training the largest G-scale model, we train the smaller models with image pretraining, then distill with image distillation in Tab.~\ref{tbl_app:distillation}, then finally apply video finetuning at the end.

\subsubsection{Zero-Shot Classification and Retrieval}
\label{appx:zeroshot_settings}

\paragraph{Zero-Shot Evaluation on Images and Videos.} We use CLIPBench\footnote{\url{https://github.com/LAION-AI/CLIP\_benchmark}} for zero-shot classification and retrieval benchmarking. The benchmark datasets and splits are obtained from the original dataset websites or HuggingFace. We extend the CLIPBench zero-shot evaluation to include video datasets such as MSR-VTT and Kinetics, and will release our model checkpoints, evaluation code, and scripts for reproducibility.

\paragraph{Prompt Design.} For zero-shot image-text and video-text retrieval, we rely solely on the original captions without any additional prompts. In contrast, for zero-shot classification, we utilize task-specific prompts graciously provided by the InternVL~\cite{internvl} authors. All additional prompts will be released.

For example, we employ specific prompts for zero-shot image classification on various ImageNet benchmarks (e.g., ImageNet val, ImageNet v2) and video classification on Kinetics datasets (e.g., K400, K600, K700).
\label{text:Zero-shot Image Classification Prompts}
\promptbox{Zero-Shot Image Classification Prompts - ImageNet}{
a bad photo of a \{c\}. a photo of many \{c\}. a sculpture of a \{c\}. a photo of the hard to see \{c\}. a low resolution photo of the \{c\}. a rendering of a \{c\}. graffiti of a \{c\}. a bad photo of the \{c\}. a cropped photo of the \{c\}. a tattoo of a \{c\}. the embroidered \{c\}. a photo of a hard to see \{c\}. a bright photo of a \{c\}. a photo of a clean \{c\}. a photo of a dirty \{c\}. a dark photo of the \{c\}. a drawing of a \{c\}. a photo of my \{c\}. the plastic \{c\}. a photo of the cool \{c\}. a close-up photo of a \{c\}. a black and white photo of the \{c\}. a painting of the \{c\}. a painting of a \{c\}. a pixelated photo of the \{c\}. a sculpture of the \{c\}. a bright photo of the \{c\}. a cropped photo of a \{c\}. a plastic \{c\}. a photo of the dirty \{c\}. a jpeg corrupted photo of a \{c\}. a blurry photo of the \{c\}. a photo of the \{c\}. a good photo of the \{c\}. a rendering of the \{c\}. a \{c\} in a video game. a photo of one \{c\}. a doodle of a \{c\}. a close-up photo of the \{c\}. a photo of a \{c\}. the origami \{c\}. the \{c\} in a video game. a sketch of a \{c\}. a doodle of the \{c\}. a origami \{c\}. a low resolution photo of a \{c\}. the toy \{c\}. a rendition of the \{c\}. a photo of the clean \{c\}. a photo of a large \{c\}. a rendition of a \{c\}. a photo of a nice \{c\}. a photo of a weird \{c\}. a blurry photo of a \{c\}. a cartoon \{c\}. art of a \{c\}. a sketch of the \{c\}. a embroidered \{c\}. a pixelated photo of a \{c\}. itap of the \{c\}. a jpeg corrupted photo of the \{c\}. a good photo of a \{c\}. a plushie \{c\}. a photo of the nice \{c\}. a photo of the small \{c\}. a photo of the weird \{c\}. the cartoon \{c\}. art of the \{c\}. a drawing of the \{c\}. a photo of the large \{c\}. a black and white photo of a \{c\}. the plushie \{c\}. a dark photo of a \{c\}. itap of a \{c\}. graffiti of the \{c\}. a toy \{c\}. itap of my \{c\}. a photo of a cool \{c\}. a photo of a small \{c\}. a tattoo of the \{c\}.
}

\label{text:Zero-shot Video Classification Prompts}
\promptbox{Zero-Shot Video Classification Prompts - Kinetics}{
a photo of \{c\}. a photo of a person \{c\}. a photo of a person using \{c\}. a photo of a person doing \{c\}. a photo of a person during \{c\}. a photo of a person performing \{c\}. a photo of a person practicing \{c\}. a video of \{c\}. a video of a person \{c\}. a video of a person using \{c\}. a video of a person doing \{c\}. a video of a person during \{c\}. a video of a person performing \{c\}. a video of a person practicing \{c\}. a example of \{c\}. a example of a person \{c\}. a example of a person using \{c\}. a example of a person doing \{c\}. a example of a person during \{c\}. a example of a person performing \{c\}. a example of a person practicing \{c\}. a demonstration of \{c\}. a demonstration of a person \{c\}. a demonstration of a person using \{c\}. a demonstration of a person doing \{c\}. a demonstration of a person during \{c\}. a demonstration of a person performing \{c\}. a demonstration of a person practicing \{c\}.
}

\paragraph{Evaluation Method.}
\label{appx:zeroshot_eval_method}
Several works use different input transformations for different datasets when evaluating zero-shot performance (e.g., \cite{eva18b,dfn,siglip,siglip2}). To be as fair as possible, we follow~\cite{eva18b} in evaluating with two transformations---center crop and non aspect ratio preserving resize (``squash'')---and report the max between the two for all models and all datasets we evaluate.
Additionally, ObjectNet has a red border around every image to facilitate deduplication, which we remove for evaluation.
Finally, we follow \cite{internvl} in using \textit{retrieval reweighting} (DSL), applying the softmax score distribution to the similarities used for retrieval:
\begin{equation}
    \texttt{scores = scores * softmax(scores, dim=0)}
\end{equation}
This slightly improves retrieval for most models, so we do it for all models we evaluate for fairness.
Notably, we were able to reproduce the reported numbers for most papers with these techniques, but for cases where we could not, we default to the reported number.

\subsection{PE: Language Alignment}
\label{appx:mmlm_benchmark_set}

We provide details of the MLLM experimental setup in \S~\ref{sec:la}. 
We describe \textit{data}, \textit{model}, and \textit{training} separately. 

\paragraph{Data.} 
Our MLLM training contains \textit{warmup} data and \textit{supervised finetuning (SFT)} data.
Our warmup data is a 1M subset image-text pairs of our \PEcore{} pretraining dataset.  
For SFT data, we use a diverse data mix consisting of 2.6M unique samples. This dataset is composed of 1.7M~\footnote{We excluded multi-images samples.} visual QAs samples from the Cauldron~\cite{laurençon2024matters}, 0.5M grounded QA pairs from Visual Genome~\cite{krishna2017visual}, Flickr-Entities~\cite{plummer2015flickr30k} and Densely Captioned Images~\cite{Urbanek_2024_CVPR}, 0.1M image-captioning pairs from COCO~\cite{coco} and 0.3M text-only samples. This comprehensive data mix allows us to thoroughly assess our model's capabilities in various MLLM tasks.

\paragraph{Model.} As described in \S~\ref{sec:la_method}, we use a simple vision-language model architecture where a vision encoder and a pretrained decoder-only LLM are connected by a vision projector. 
For all tables, we use either Llama3.1-instruct 8B or QwenLM 2.5-instruct 7B as a language model, and 2-layer MLP as a vision projector. 
For fair comparison, we use the native resolution for image input. 
During inference, we evaluate the models on video tasks in \textit{zeroshot} manner: We concatenate all video frames into a sequence and feed to language model, without seeing video samples during SFT. 
For all video tasks, we use $8$ frames with the same native resolution of height and width. 
For \PEcore{} and \PElang{}, this makes $448\times 448\times8$ input and $32\times 32\times 8$ vision tokens.

\paragraph{Training.} 
MLLM training consists of \textit{warmup} and \textit{supervised finetuning (SFT)} stages. 
In both stages, we freeze vision encoder and train vision projector and LLM.
During warmup stage, we use a global batch size of $128$ with a learning rate of $1\times 10^{-4}$.
We gradually increase the learning rate from $1\times 10^{-6}$ to $1\times 10^{-4}$ over 120 steps, and follow a cosine learning rate decay schedule to train a total of 8,000 steps. 
During SFT stage, we use a global batch size $256$ with a learning rate of $1\times 10^{-5}$.
Similar to the warmup, we gradually increase the learning rate from $1\times 10^{-7}$ to $1\times 10^{-5}$ over 300 steps, and follow a cosine learning rate decay schedule to train a total of 12.5K steps. 
We truncate text-sequences longer than 2,048 tokens on top the visual tokens. 
This makes the maximum sequence length to be $\texttt{(num. vision tokens)} + 2,048$.
With $448\times 448$ input resolution and patch size of $14$, we set the maximum sequence length to $1,024 + 2,048 = 3,072$.
To represent bounding boxes on output side for image grounding tasks, we simply use text tokens to represent each bounding box: each coordinate is normalized between \texttt{000} and \texttt{999}, in ``\texttt{[x, y, x, y]}'' box format for top-left and bottom-right corners (\emph{e.g.}, \texttt{[012, 122, 633, 782]}). 

For all baselines, we search for the \textbf{best} intermediate layer features to adapt to LLM. 
We search over $\{-1, -2, -4, -6, -8, -10, -12,-14, -16, -18, -20, -40\}$ layers (counting from last) and report the best result in average over OCR/Chart/Document Q\&A, Visual Q\&A, Image Captioning and Video Understanding.

\subsection{PE: Spatial Alignment}
\subsubsection{Training Details}
\label{appx:spatial_align_details}
\paragraph{Loss Functions.}
For self-aligning to frozen \PEcore{G} layer 41 features ($L_\text{core}$), we minimize cosine similarity:
\begin{equation} \label{eq:loss_core}
    L_\text{core} = \frac{1}{n_\text{tok}}\sum\left(\frac{(S_{50})(T_{41})^T}{||S_{50}||\cdot||T_{41}||}\right)
\end{equation}
where $S_{50}$ denotes the last layer features of the student, $T_{41}$ denotes frozen layer 41 features from \PEcore{G}, and $n_\text{tok}$ represents the number of tokens. Note that we chose 41 fairly arbitrarily (it is layer 40 when written with indexing from 0). Judging by Fig.~\ref{fig:layerfinder}, any layer around 40 should work (and 39 may be slightly better).

For the encouraging locality loss ($L_\text{loc}$), we compute the pairwise cosine similarity between a model's own tokens and itself. This forms a ``spatial correspondence map'' for what tokens should be considered similar. We then compute the same for the student, and minimize the difference between the two with MSE loss:
\begin{equation} \label{eq:loss_sam}
    L_\text{loc} = \frac{1}{n_\text{tok}^2} \sum\left(\frac{{(S_{50})} ({S_{50}})^T}{||{S_{50}}||^2} - \frac{(T_\text{SAM}) (T_\text{SAM})^T}{||T_\text{SAM}||^2}\right)^2
\end{equation}
where $T_\text{SAM}$ denotes the ``SAM Mask Logits'' constructed in \S\ref{sec:sa_method}.
We also find using a temperature ($t$) on the SAM teacher's pairwise cosine similarity term ($x$) useful: $e^{t(x - 1)}$.
The full loss is $L_\text{spatial} = L_\text{core} + L_\text{loc}$.

\paragraph{Hyperparameters.}
In Tab.~\ref{tbl_app:spatial_train} we show the training hyperparameters for spatial alignment, finetuned on top of the initial \PEcore{G} checkpoint. Then in Tab.~\ref{tbl_app:spatial_sam} and Tab.~\ref{tbl_app:spatial_core}, we show the settings for the two teachers and losses. Note that when running the teachers, we run them on the exact same image as the student (same data aug and all). Additionally, because the SAM 2.1 teacher operates at a resolution of 1024, we upsample the image, generate the mask logits, and then downsample the result. Both teachers are frozen.

\begin{table}[!h]
\begin{minipage}[t]{0.3\textwidth}
\vspace{0pt}
\centering
{
\tablestyle{5pt}{1.1}
\begin{tabular}{l | c }
config  & values \\
\hline
\addpadding
optimizer & LAMB \\ 
$\beta_1, \beta_2$ & (0.9, 0.95)\\
weight decay & 0.05 \\
learning rate & 5e-4\\
batch size & 12,288 \\
warm-up steps & 0 \\
training steps & 24K \\
data quantity & 5.4B {(\tiny \PEcore{} PT Data)} \\
samples seen & 300M \\
\\
resolution & 448 \\
mask ratio & 0.75\\
mask size & 2$\times$2 tokens \\
\\
droppath & 0.4 \\
layerscale & 0.1 \\
\\
data aug & %
    \makecell{
        \text{aspect jitter} \text{\texttt{\tiny ar(0.75,1.33)}} \\
        \text{color jitter} \text{\texttt{\tiny j(0.32,0,0.32,0)}}\\
        \text{hflip} \text{\texttt{\tiny p(0.5)}}
    }\\
\end{tabular} \\
}
\captionsetup{justification=centering}
\caption{\bf Spatial Alignment.}
\label{tbl_app:spatial_train}
\end{minipage}
\hfill
\begin{minipage}[t]{0.3\textwidth}
\vspace{0pt}
\centering
{
\tablestyle{5pt}{1.1}
\begin{tabular}{l | c }
config  & values \\
\hline
\addpadding
model & SAM 2.1-L \\
layer & mask logits \\
resolution & 1024 (\texttt{interp}$\rightarrow$448) \\
\\
loss & Eq.~\ref{eq:loss_sam} \\
loss weight & 1 \\
temperature & 20 \\
\\
sample points & 32$\times$32 (1024) \\
pred iou threshold & 0 \\
stability score threshold & 0 \\
mask threshold & 0 \\

\end{tabular} \\
}
\captionsetup{justification=centering}
\caption{\bf SAM 2.1 Teacher.}
\label{tbl_app:spatial_sam}
\end{minipage}
\hfill
\begin{minipage}[t]{0.3\textwidth}
\vspace{0pt}
\centering
{
\tablestyle{5pt}{1.1}
\begin{tabular}{l | c }
config  & values \\
\hline
\addpadding
model & \PEcore{G} \\
layer & 41 \\
resolution & 448 \\
\\
loss & Eq.~\ref{eq:loss_core} \\
loss weight & 1 \\
\end{tabular} \\
}
\captionsetup{justification=centering}
\caption{\bf \PEcore{G} Teacher.}
\label{tbl_app:spatial_core}
\end{minipage}
\end{table}

\subsubsection{Visualization Method}
\label{appx:feature_viz}
\begin{figure}[h!]
    \centering
    \includegraphics[width=1\linewidth, trim=1.5in 0in 0in 0in, clip]{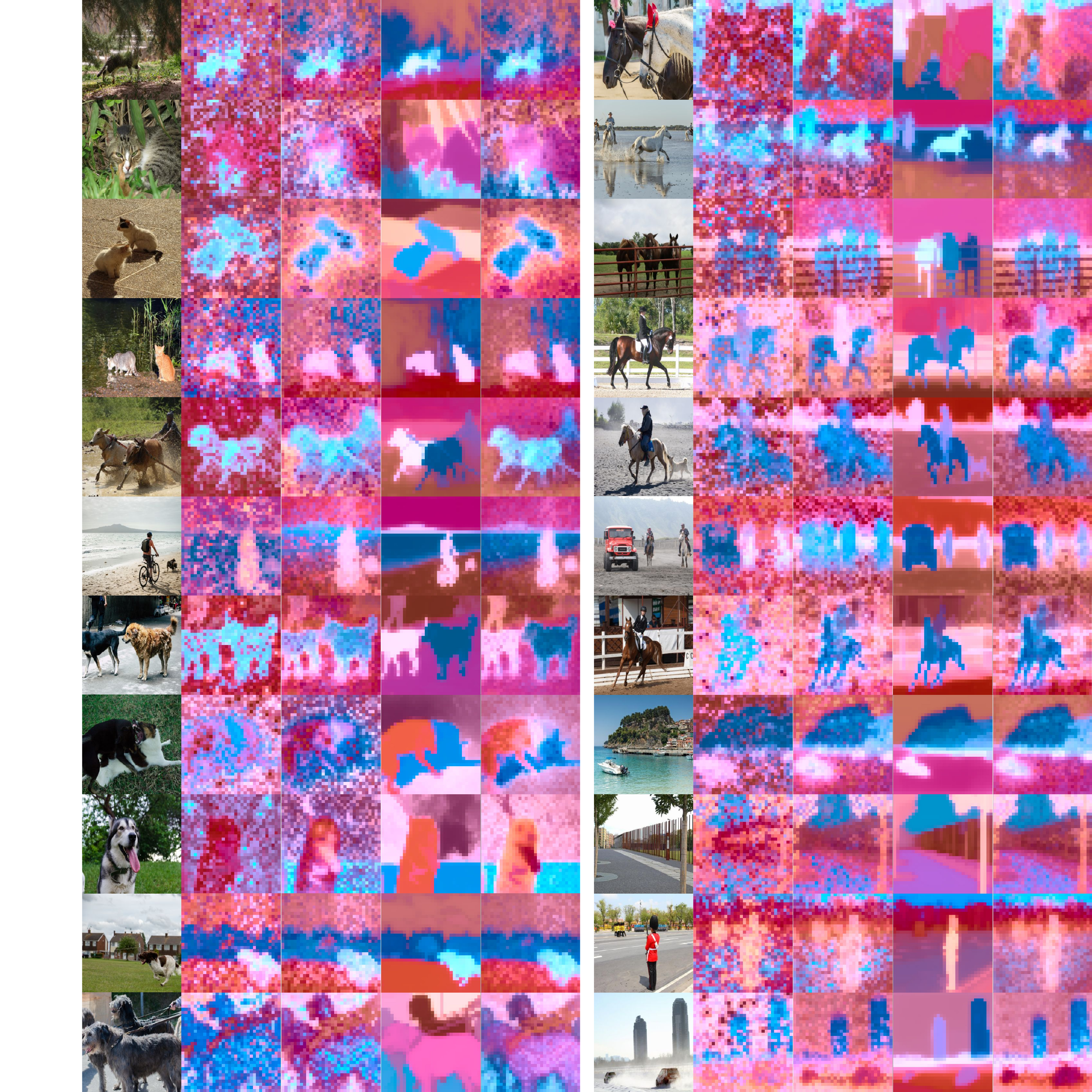}
    \caption{{\bf More Visualizations} of the feature space following Fig.~\ref{fig:feature_viz}. After the image itself, column 1 is \PEcore{G} last layer features, column 2 is \PEcore{G} aligned to its own layer 41, column 3 is \PEcore{G} aligned to SAM 2.1-L~\cite{sam2} mask logits, and column 4 is \PEcore{G} aligned to both, denoted \PEspat{G}. See \S\ref{appx:feature_viz} for visualization method.}
    \label{fig:more_feature_viz}
\end{figure}

To visualize the features in Fig.~\ref{fig:feature_viz} and Fig.~\ref{fig:more_feature_viz}, our goal is to map a 1536-dimensional space down to 3 dimensions to view how the model encodes each token in relation to each other. One naive approach would be to apply PCA with 3 dimensions across all token in the image. However, we find this alone can be misleading.

Specifically, if the model has rich semantics, it should be the case that most of those 1536 features have some useful information in them. Some of that information could be spatially contiguous, some of it not. We want PCA to only select the \textit{spatially contiguous} information, since we are trying to evaluate the spatial quality of the features. However, naively applying PCA will not necessarily do that, especially for models with information aggregated in ``global tokens'' (\S\ref{sec:core_feature_analysis}). Despite these tokens carrying important information, they are not spatially contiguous. Thus, if PCA dedicates a large portion of its 3 dimensions to global tokens, the features will \textit{look} like their spatial quality is bad, despite the features containing good spatial information.

\begin{figure}[b!]
    \centering
    \includegraphics[width=0.5\linewidth, trim=4.45in 0in 0in 15in, clip]{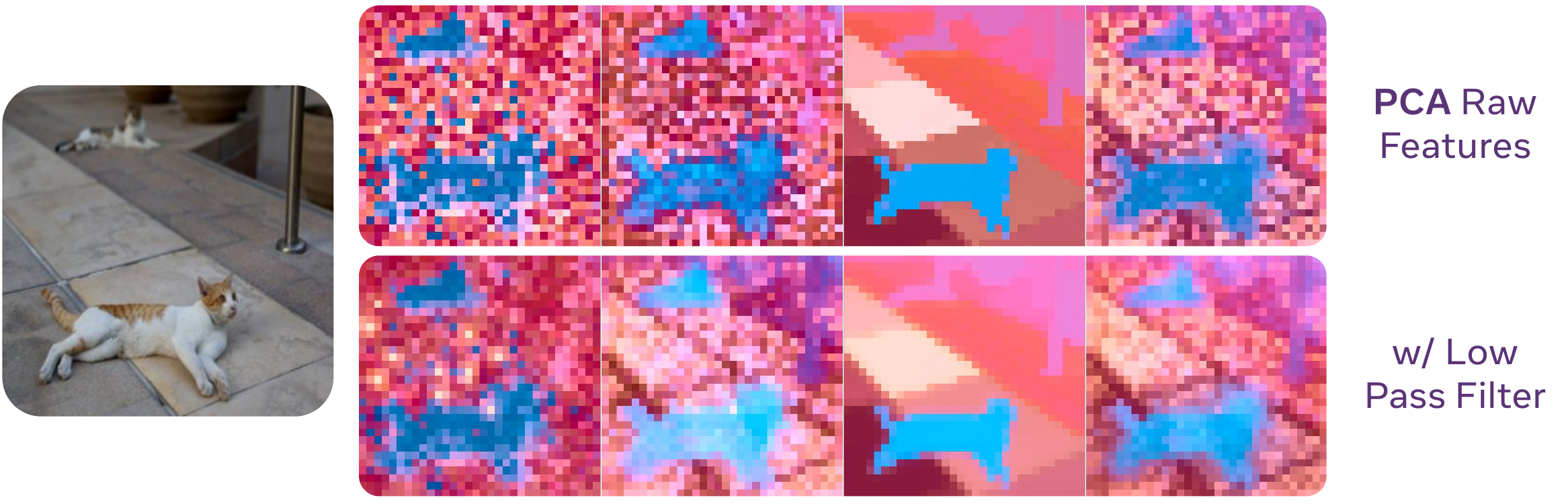}
    \caption{{\bf Feature Visualization Ablation.} With raw features (top row), PCA misses spatially contiguous parts of the feature space and instead focuses on global tokens (which carry information but are not spatially coherent). By applying a simple low pass filter (bottom row), we can reveal spatial information that PCA originally missed (see column 2: with raw features, the background looks like a mess, with the low pass filter the tiles become visible).}
    \label{fig:feature_viz_comparison}
\end{figure}

So, how do we select for only the \textit{spatially contiguous} information to visualize? The answer is simple: by definition, the spatially contiguous information will be\ldots\ spatially contiguous. To keep the spatially contiguous information while lowering the impact of the global tokens, we can simply apply a low pass filter to the features (specifically, a gaussian blur with kernel size 3 and a $\sigma$ of 1). To retain the detail of the original features, we can average the two together. Thus, to visualize features, we use the 3D PCA of the of the following. $x$ denotes the model's output features, and $g(x)$ denotes gaussian blur.
\begin{equation}
    0.5 x + 0.5g(x,k=3,\sigma=1)
\end{equation}
We show the impact of this in Fig.~\ref{fig:feature_viz_comparison}. Blurring the features make them appear more detailed! In reality, that information was always there, just PCA did not show it. Thus, great care must be taken when visualizing high dimensional feature spaces. If they were easy to map to 3 dimensions---you wouldn't need 1536 of them!

Then, to map the PCA dimensions to RBG pixel values, we map each PCA component to a corresponding channel in LCh color space, then convert those LCh colors to RGB to get the final image. Note that we use LCh instead of RGB directly for aesthetic reasons, and also because LCh is a cylindrical color space---where smooth changes to the values look like smooth changes in colors to humans---and thus is easier to discern.

\subsubsection{Frozen Feature Dense Prediction }
\label{appx:dense_pred}

We discuss the detailed settings of the results for dense prediction with frozen features in Tab.~\ref{tbl:dense_prediction}. Each model is evaluated with its native resolution up to 448 or 448 (whichever is optimal).

\paragraph{Zero-Shot Tracking.} We evaluate our pretrained models on label propagation task using the protocols in ~\cite{jabri2020space, vgpt} on DAVIS dataset~\cite{davis2017}. This evaluation does not require any finetuning or probing, therefore preserves the spatial features in the model. Following Toto~\cite{vgpt}, we use the features from the last n\,$=$\,7 frames to find the nearest neighbor patch in the current frame, and then propagate the masks from the previous frames to the current frame. Note that this evaluation method does not require any training.

\paragraph{Semantic Segmentation.} For semantic segmentation, we evaluate our pretrained models on ADE20K~\cite{ade20k} semantic segmentation task. We use a linear layer and convolutional layer to map intermediate spatial features to segmentation masks following~\cite{dinov2}. The models are evaluated and then features are resized to 518\,$\times$\,518. We only use features from single layer. The probing layers are finetuned with AdamW~\cite{adamw} with a learning rate of 0.001.

\paragraph{Depth Estimation.} For depth estimation on NYUv2~\cite{nyu_depth}, we follow~\cite{li2024binsformer,dinov2}.
We use a DPT-head~\cite{dpt} on top of our frozen pretrained model and use only single layer features. We scale the size of the DPT-head for each models based on the hidden size for each architecture. Because NYU is a small dataset and the models we evaluate are large, we observe the results for most models are noisy and prone to overfitting. Thus, for fair comparison we train \textit{all models} for 20 epochs and for \textit{all models} take the lowest validation loss over all epochs.

\paragraph{Frozen Detection.} For the frozen feature detection results presented in \S\ref{sec:layerfinder}, we evaluated using Mask R-CNN~\cite{maskrcnn} as a probe. We used a resolution of 1024 for Fig.~\ref{fig:layerfinder} and 768 for the remainining experiments in \S\ref{sec:layerfinder}. Because the backbones were frozen, we did not add any global attention and instead simply tiled the input image with a window size of 32 for the 1024px experiments and 24 for the 768px experiments. All models were interpolated to patch 16. Finally, the backbones were frozen and only the FPN and R-CNN heads trained for 15 epochs on COCO with a stepwise decay LR without drop path.

\subsubsection{End-to-End Finetuning Detection and Segmentation}
\label{appx:det}
We provide a detailed discussion of settings of end-to-end finetuning on detection and segmentation presented in Tab.~\ref{tbl:det_SFT}. The hyperparameters can be found in Tab.~\ref{tbl_app:coco}. We find that the default 100-epoch protocol in ViTDet~\cite{vitdet,d2} causes overfitting problems in COCO experiments especially for billion-level parameter vision encoders, so we tune the training epochs, learning rate, drop path and learning rate decay accordingly.

The LVIS experiment setting is the same as COCO except all L-size models use learning rate of 2e-4 and all g-size and G-size models use 75 epochs.

\begin{table}[!h]
\centering

\tablestyle{5pt}{1.1}
\begin{tabular}[t]{l | c }
config  & values \\
\hline
optimizer & AdamW \\ 
optimizer momentum & (0.9, 0.999)\\
weight decay & 0.1 \\
learning rate &  $\rightarrow$ \\
learning rate schedule & Step-wise decay\\
learning rate decay &  $\rightarrow$ \\
batch size & 64 \\
image size & 1024$\times$1024 \\
augmentation & LSJ [0.1, 2.0] \\
epochs & $\rightarrow$ \\
drop path & $\rightarrow$ \\
postional embedding & abswin~\cite{abswin} \\
patch size & 16 \\
window size & $\rightarrow$ \\
global window index & $\rightarrow$ \\ 
\end{tabular}
\quad\quad\quad\quad
\begin{tabular}[t]{l | c c c c c c c}
model & lr & epochs & drop path & lr decay & layers & global window index & window size \\
\hline
OpenAI CLIP-L & 1e-4 & 100 & 0.4 & 0.8 & 24 & (5, 11, 17, 23) & 14 \\
MetaCLIP-L & 1e-4 & 100 & 0.4 & 0.8 & 24 & (5, 11, 17, 23) & 14\\
MetaCLIP-G & 5e-5 & 75 & 0.5 & 0.9 & 48 & (11, 23, 35, 47) & 14\\
SigLIP-so & 1e-4 & 100 & 0.4 & 0.8 & 27 & (2, 10, 18, 26) & 14 \\
EVA02-L & 1e-4 & 100 & 0.4 & 0.8 & 24 & (5, 11, 17, 23) & 14 \\
MAE-L & 1e-4 & 100 & 0.4 & 0.8 & 24 & (5, 11, 17, 23) & 14 \\
SigLIP2-so & 1e-4 & 100 & 0.4 & 0.8 & 27 & (2, 10, 18, 26) & 14 \\
SigLIP2-g & 5e-5 & 75 & 0.5 & 0.9 & 40 & (9, 19, 29, 39) & 14 \\
DINOv2-L & 1e-4 & 100 & 0.4 & 0.8 & 24 & (5, 11, 17, 23) & 32 \\
DINOv2-g & 5e-5 & 36 & 0.5 & 0.9 & 40 & (9, 19, 29, 39) & 32 \\
{\bf \PEcore{}G} & 5e-5 & 75 & 0.5 & 0.9 & 50 & (12, 24, 36, 49) & 32 \\
{\bf \PEspat{}G} & 5e-5 & 36 & 0.5 & 0.9 & 50 & (12, 24, 36, 49) & 32 \\
\end{tabular}
\captionsetup{justification=centering}
\caption{\bf Settings for End-to-End Finetuning Detection and Segmentation.}
\label{tbl_app:coco}
\end{table}

\clearpage

\subsubsection{System-Level Comparison on Detection}
\label{appx:det_sota_setting}

\begin{wraptable}{r}{0.35\textwidth}
\vspace{-30pt}
\centering
\tablestyle{3pt}{1.1} 
\begin{tabular}{y{80}x{25}}
    \shline
    Test-Time Aug & \ct[c7]{AP$_\text{box}$}{}\\
    \hline
    \addpadding{}
    No TTA & 65.2 \\
    + More Queries & 65.3 \\
    + SoftNMS~\cite{softnms} & 65.8 \\
    + Flip Aug &  65.8 \\
    + Multiscale Aug & \textbf{66.0} \\
    \shline
\end{tabular}
\caption{{\bf Test-Time Aug} for system-level comparison on COCO in Tab.~\ref{tbl:det_coco}.}
\label{tab:det_tta}
\vspace{-10pt}
\end{wraptable}

We describe our implementation for system-level comparison to the state-of-the-arts on COCO object detection in Tab~\ref{tbl:det_coco}. Our implementation is based on the DETA repository\footnote{\url{https://github.com/jozhang97/DETA}}. We replace the vision encoder with our \PEspat{} and maintain the same hyperparameters as in the end-to-end finetuning settings, while keeping the detector unchanged. The training process consists of three stages:

\begin{enumerate}
    \item {\bf Initial Training}: Train on Objects365 for 12 epochs with an image resolution of 1024\,$\times$\,1024, a total batch size of 256, and a learning rate of 2e-4, which is divided by 10 at the 10th epoch.
    
    \item {\bf Increasing Resolution}: Continue training on Objects365 for 6 epochs with a resolution of 1536\,$\times$\,1536, a total batch size of 128, and a learning rate of 5e-5, which is divided by 10 at the 5th epoch.

    \item {\bf Finetuning}: Finetune on COCO dataset for 12 epochs with an image resolution of 1728\,$\times$\,1728, a total batch size of 64, and a learning rate of 5e-5, which is divided by 10 at the 8th epoch.

    \item {\bf Further Increasing Resolution}: Further finetune on COCO dataset for 3 epochs with a resolution of 1824\,$\times$\,1824, a total batch size of 64. To save GPU memory, we use SGD optimizer instead of Adam, with a learning rate of 5e-3, which is divided by 10 at the 2th epoch.
\end{enumerate}

We apply a series of test-time augmentation techniques to further improve the performance, detailed in Tab.~\ref{tab:det_tta}.

\section{Additional Results}

\subsection{\PEcore{}: Robust Image Pretraining}
\label{appx:core_img_pt}

In Tab.~\ref{tab:core_img_pretraing_raw}, we present the raw data for the robustness metrics in Fig.~\ref{fig:core_pt_ablations}. Across the board, each change improved almost all metrics (with the exception of progressive resolution slightly hurting the average and mask regularization slightly hurting ImageNet Adversarial). The fact that there were no tradeoffs to these changes, indicate that their improvements to the features are general. This could be why most of these changes improved performance for downstream tasks as well.

Note that in \S\ref{sec:core_image_pt}, we only discuss changes that we know to work. There are several changes that we have tried that do not work (i.e., do not improve performance or lower performance). For instance: average pooling instead of using a class token, increasing the text tower size, using hue or contrast jitter, and maintaining the same resolution throughout training but dropping tokens instead of progressive resolution (FLIP-style).

We also find increasing batch size and increasing training iterations for an L scale model to have equivalent effects. This is in contrast to the batch size scaling observed by~\cite{siglip}, but it is possible that this difference is down to a hyperparameter issue.

\begin{table*}[!h]
    \centering
    \makebox[\linewidth][c]{
    \tablestyle{0pt}{1.15} 
    \begin{tabular}{wy{90} awwwwww}
        \shline
         & \multirow{2}{*}{\vspace{-2.2cm} Step} & \multicolumn{7}{c}{\ct[c1]{\it Zero-Shot Classification}} \\
            & 
            & \cb[c1]{\textit{\textbf{Avg Class.}}}{}
            & \cb[c1]{ImageNet}{val~\cite{imagenet}}
            & \cb[c1]{ImageNet}{v2~\cite{imagenetv2}}
            & \cb[c1]{ObjectNet}{IN Classes~\cite{objectnet}}
            & \cb[c1]{ImageNet}{Adversarial~\cite{imagenet-a}}
            & \cb[c1]{ImageNet}{Renditions~\cite{imagenet-r}}
            & \cb[c1]{ImageNet}{Sketch~\cite{imagenet-sketch}}
            \\
        \hline
        \addpadding
        1 & Baseline               & \ca{75.3} & 78.9 & 71.9 & 73.7 & 68.3 & 91.1 & 67.8 \\
        2 & Progressive Resolution & \ca{75.1} & 78.9 & 71.8 & 72.4 & 69.9 & 90.5 & 67.0 \\
        3 & High Batch Size        & \ca{76.2} & 79.5 & 72.8 & 74.1 & 71.8 & 91.0 & 68.1 \\
        4 & LAMB and High LR       & \ca{76.9} & 79.9 & 73.3 & 74.3 & 73.5 & 91.5 & 68.6 \\
        5 & High Resolution (336)  & \ca{78.3} & 80.4 & 73.8 & 75.6 & 79.2 & 92.0 & 68.8 \\
        6 & 2D RoPE                & \ca{79.2} & 80.7 & 74.1 & 77.4 & 80.9 & 92.7 & 69.4 \\
        7 & Attention Pooling      & \ca{80.1} & 81.0 & 74.8 & 78.4 & 82.9 & 93.4 & 69.9 \\
        8 & Data Augmentation      & \ca{80.8} & 81.1 & 75.2 & 80.8 & 83.1 & 93.5 & 71.2 \\
        9 & Mask Regularization    & \ca{80.9} & 81.3 & 75.3 & 80.9 & 82.8 & 93.8 & 71.2 \\
        \shline
    \end{tabular}
    }
    \caption{{\bf Robust Image Pretraining Full Results.} Raw results for the robustness metrics metrics in Fig.~\ref{fig:core_pt_ablations}. Almost every change improves every metric, but some metrics are improved more than others  (e.g., ObjectNet and ImageNet-A).
    }
    \label{tab:core_img_pretraing_raw}
\end{table*}

\subsection{\PEcore{}: Video Data Scaling}

\begin{table}[!h]
    \centering
    { 
    \tablestyle{0pt}{1.05} 
    \begin{tabular}{x{20} awwwwww awwwwwww}
        \shline
          & \multicolumn{7}{c}{\ct[c1]{\it Image Zero-Shot}} & \multicolumn{8}{c}{\ct[c3]{\it Video Zero-Shot}} \\
              \cb{Video Data Size}{}
            & \cb[c1]{\textit{\textbf{Average Image}}}{}
            & \cb[c1]{ImageNet}{val~\cite{imagenet}}
            & \cb[c1]{ImageNet}{v2~\cite{imagenetv2}}
            & \cb[c1]{ObjectNet}{IN Classes~\cite{objectnet}}
            & \cb[c1]{ImageNet}{Adversarial~\cite{imagenet-a}}
            & \cb[c2]{MS-COCO}{txt$\rightarrow$img~\cite{coco}}
            & \cb[c2]{MS-COCO}{img$\rightarrow$txt~\cite{coco}}
            & \cb[c3]{\textit{\textbf{Average Video}}}{}
            & \cb[c3]{Kinetics}{400~\cite{kay2017kinetics}}
            & \cb[c3]{Kinetics}{600~\cite{kay2017kinetics}}
            & \cb[c3]{Kinetics}{700~\cite{kay2017kinetics}}
            & \cb[c3]{UCF 101}{\cite{soomro2012ucf101}}
            & \cb[c3]{HMDB 51}{\cite{kuehne2011hmdb}}
            & \cb[c3]{MSR-VTT}{txt$\rightarrow$vid~\cite{vtt}}
            & \cb[c3]{MSR-VTT}{vid$\rightarrow$txt~\cite{vtt}}
            \\
            
        \hline
        \addpadding
        0M              
        & \cat{77.0} & 83.9 & 78.6 & 86.6 & 90.3 & 52.1 & 70.3
        & \cat{57.0} & 70.3 & 69.4 & 61.6 & 78.5 & 47.4 & 40.5 & 31.4
        \\        
        3M              
        & \ca{77.7} & 84.1 & 78.8 & 86.6 & 90.9 & 53.3 & 74.2
        & \ca{61.6} & 72.4 & 72.2 & 64.2 & 88.5 & 53.8 & 42.8 & 37.6
        \\
        6M
        & \ca{78.0} & 84.2 & 79.0 & 86.7 & 91.1 & 54.0 & 72.7
        & \ca{63.6} & 73.5 & 73.4 & 66.0 & 88.9 & 54.6 & 44.9 & 43.6 
        \\
        8M           
        & \ca{78.4} & 84.2 & 79.2 & 87.0 & 91.6 & 54.9 & 73.6
        & \ca{64.8} & 74.5 & 74.5 & 67.7 & 89.5 & 55.3 & 46.9 & 45.5
        \\
        11M 
        & \ca{78.6} & 84.2 & 79.2 & 87.2 & 91.8 & 55.4 & 73.8
        & \ca{65.2} & 75.1 & 75.0 & 67.6 & 89.7 & 55.6 & 47.7 & 45.8
        \\
        14M
        & \ca{78.8} & 84.2 & 79.2 & 87.5 & 91.9 & 55.7 & 74.3
        & \ca{65.5} & 75.4 & 75.3 & 67.9 & 89.9 & 55.8 & 47.8 & 46.3
        \\
        17M
        & \ca{78.9} & 84.2 & 79.2 & 87.7 & 92.0  & 55.8 & 74.3 
        & \ca{65.8} & 75.7 & 75.5 & 68.2 & 90.2 & 56.0 & 48.3 & 46.7
        \\      
        \shline
    \end{tabular}
    }
    \caption{{\bf Scaling Video Data.} Increasing the number of synthetic video data generated by our proposed video data engine consistently enhances the performance of image and video classification and retrieval tasks.
    }
    \label{tab:video-ft-ablation-details}
\end{table}

The detailed video data scaling results are presented in Tab.~\ref{tab:video-ft-ablation-details}. Our experiments demonstrate that increasing the number of synthetic video data generated by the proposed video data engine enhances the performance of classification and retrieval on both image and video benchmarks. On image benchmarks, while improvements on ImageNet val and v2 plateaued earlier compared to ObjectNet and ImageNet Adversarial, MS-COCO retrieval performance continued to show gains. On video benchmarks, scaling synthetic video data consistently yields better performance for both classification and retrieval tasks. We expect that further scaling up the video data with our video data engine will continue to drive performance improvements.

\subsection{\PEcore{}: Smaller Models}
\label{appx:core_smaller_models}

\begin{table*}[!h]
    \centering
    \makebox[\linewidth][c]{
    \tablestyle{0pt}{1.15} 
    \begin{tabular}{y{105} ww awwwwww}
        \shline
        \multirow{2}{*}{\vspace{-2.2cm} Model}  &&& \multicolumn{7}{c}{\ct[c1]{\it Zero-Shot Classification}} \\
            & \cb{Teacher's Temp}{}
            & \cb{Model Scale}{}
            & \cb[c1]{\textit{\textbf{Avg Class.}}}{}
            & \cb[c1]{ImageNet}{val~\cite{imagenet}}
            & \cb[c1]{ImageNet}{v2~\cite{imagenetv2}}
            & \cb[c1]{ObjectNet}{IN Classes~\cite{objectnet}}
            & \cb[c1]{ImageNet}{Adversarial~\cite{imagenet-a}}
            & \cb[c1]{ImageNet}{Renditions~\cite{imagenet-r}}
            & \cb[c1]{ImageNet}{Sketch~\cite{imagenet-sketch}}
            \\
        \hline
        \addpadding
        vanilla pretrained model & -  & B  & \ca{66.2} & 74.2 & 67.4 & 62.5 & 50.2 & 83.0 & 59.8 \\ 
        \hline
        \addpadding
        \multirow{4}{*}{distillation} & $\times$2 & B & \ca{65.2} & 71.8 & 65.5 & 61.4 & 50.2 & 83.6 & 58.6 \\
         & $\times$1 & B & \ca{68.0} & 74.9 & 68.1 & 64.7 & 54.1 & 85.3 & 61.1\\
         & $\times$0.7 & B & \ca{68.2} & 75.1 & 68.2 & 65.3 & 54.4 & 85.1 & 61.3 \\
         & $\times$0.5 & B & \ca{\textbf{68.3}} & 75.2 & 68.2 & 65.3 & 54.2 & 85.2 & 61.4 \\
        \hline
    \end{tabular}
    }
    \caption{{\bf Ablation Study on Teacher's Distribution Temperature.} We evaluate the effect of varying temperatures on the teacher's distribution, using a pretrained vanilla CLIP model (ViT-B/14, resolution 224) as a baseline (details in \S\ref{sec:core_image_pt}). The models are finetuned via distillation with a short schedule of 50K steps.
    }
    \label{tab:distillation-temperatrue-ablation}
\end{table*}

\paragraph{Ablation: Distillation Temperature.} 
To optimize the performance of smaller models (B and L-scales in Tab.~\ref{tab:pe2b}), we utilize a distillation finetuning approach with \PEcore{G} as the teacher model. During this process, both student and teacher models encode image and text inputs to compute image-to-text and text-to-image similarity distributions, similar to CLIP training~\cite{clip}. The student's distributions are then optimized to match those of the teacher by minimizing KL-divergence loss on both image-to-text and text-to-image similarity distributions.

We find that using a fixed and smaller temperature (i.e., higher logit scale), which controls the range of logits in the softmax, significantly enhances the effectiveness of distillation. This results in a sharper distribution for the teacher's distributions. In contrast, the student's temperature remains learnable, consistent with our pretraining procedure and CLIP training.

In Tab.~\ref{tab:distillation-temperatrue-ablation}, we present an ablation study examining the impact of temperature on the teacher's distribution. For this analysis, we utilize a pretrained \textit{vanilla} CLIP model (ViT-B/14, resolution 224), which serves as a baseline for comparison (see \S\ref{sec:core_image_pt} for details). The models are finetuned using distillation with a concise schedule of 50K steps. Notably, our results show that employing a smaller temperature for the teacher's distributions yields improved performance on zero-shot ImageNet benchmarks.

\begin{table}[!h]
    \centering
    {
    \tablestyle{0pt}{1.05} 
    \begin{tabular}{x{70} y{90} awwwwww awwwwwww}
        \shline
          && \multicolumn{7}{c}{\ct[c1]{\it Image Zero-Shot}} 
           & \multicolumn{8}{c}{\ct[c3]{\it Video Zero-Shot}} \\
              Model
           & Stage & \cb[c1]{\textit{\textbf{Average Image}}}{}
            & \cb[c1]{ImageNet}{val~\cite{imagenet}}
            & \cb[c1]{ImageNet}{v2~\cite{imagenetv2}}
            & \cb[c1]{ObjectNet}{IN Classes~\cite{objectnet}}
            & \cb[c1]{ImageNet}{Adversarial~\cite{imagenet-a}}
            & \cb[c2]{MS-COCO}{txt$\rightarrow$img~\cite{coco}}
            & \cb[c2]{MS-COCO}{img$\rightarrow$txt~\cite{coco}}
            & \cb[c3]{\textit{\textbf{Average Video}}}{}
            & \cb[c3]{Kinetics}{400~\cite{kay2017kinetics}}
            & \cb[c3]{Kinetics}{600~\cite{kay2017kinetics}}
            & \cb[c3]{Kinetics}{700~\cite{kay2017kinetics}}
            & \cb[c3]{UCF 101}{\cite{soomro2012ucf101}}
            & \cb[c3]{HMDB 51}{\cite{kuehne2011hmdb}}
            & \cb[c3]{MSR-VTT}{txt$\rightarrow$vid~\cite{vtt}}
            & \cb[c3]{MSR-VTT}{vid$\rightarrow$txt~\cite{vtt}}
            \\
            
        \hline
        SigLIP2-L/16~\cite{siglip2} & -
        & \cat{76.0} & 83.1 & 77.4 & 84.4 & 84.3 & 55.3 & 71.4
        & \cat{56.2} & 65.3 & 62.5 & 56.8 & 86.7 & 49.3 & 41.5 & 31.4 \\        
        \PEcore{L} & image pretraining
        & \cat{75.1} & 82.9 & 76.8 & 81.8 & 85.6 & 53.0 & 70.4
        & \cat{59.0} & 68.0 & 67.7 & 58.5 & 85.5 & 57.7 & 42.0 & 33.4 \\
        \PEcore{L} & +image distillation from \PEcore{G}
        & \cat{77.6} & \textbf{83.6} & \textbf{78.1} & 84.4 & 88.9 & 56.0 & 74.7
        & \cat{64.5} & 73.0 & 72.6 & 64.8 & 86.5 & 58.0 & 47.9 & 48.4 \\  
        \PEcore{L} & +video finetuning
        & \cat{\textbf{78.0}} & 83.5 & 77.9 & \textbf{84.7} & \textbf{89.0} & \textbf{57.1} & \textbf{75.9}
        & \cat{\textbf{65.3}} & \textbf{73.4} & \textbf{72.7} & \textbf{65.3} & \textbf{87.1} & \textbf{58.5} & \textbf{50.3} & \textbf{50.1} \\              
        \shline
    \end{tabular}
    }
    \caption{{\bf Building Strong Smaller Models.} This table illustrates the step-by-step process of developing the \PEcore{L} 336px model, as outlined in \S\ref{sec:unified-encoder}. Starting with the pretrained \PEcore{L}, both image distillation, along with video finetuning, enhance performance across image and video benchmarks, resulting in a unified L-scale model.}
    \label{tab:smaller-models}
\end{table}

\paragraph{Building strong smaller models.} In Tab.~\ref{tab:smaller-models}, we demonstrate our step-by-step training strategy for building strong smaller models at the L scale, as discussed in \S\ref{sec:unified-encoder}. Specifically, we outline our approach to image pretraining, image distillation, and video finetuning, and distillation. Leveraging the robust foundation established by our pretraining techniques 
(\S\ref{sec:core_image_pt}), we show that distilling from \PEcore{G}, our strongest unified perception encoder, yields improvements on both image and video benchmarks. Furthermore, a short-scheduled video finetuning provides an additional boost in performance on both benchmarks.

\subsection{\PElang{}: Additional Results}
\label{appx:mmlm_benchmark_results}

Analogous to Tab.~\ref{tab:lang_mllm_bench}, in Tab.~\ref{tab:lang_mllm_bench_tiling}, we compare \PEcore{} and \PElang{} with \textit{dynamic resolution} setting~\cite{liu2024llavanext,llama3}. More specifically, we use up to 4 tiles, following after a \textit{thumbnail}, which is a whole image resized into $448\times 448$. 
With the maximum number of tiles of 4, the model can cover $\{1\times 1, 1\times 2, 1\times 3, 1\times 4, 2\times 1, 2\times 2, 3 \times 1, 4\times 1\}$ tile ratios. 
Similar to the Tab.~\ref{tab:lang_mllm_bench},~\ref{tab:lang_mllm_bench_qwen},~\ref{tab:lang_mllm_system_level} in the main paper, we show that \PElang{} largely outperforms the baseline vision encoders by large margins across all categories of MLLM tasks. 
Note that \PElang{} has been alignment-tuned with native resolution input, as opposed to \textit{e.g.}, InternViT 2.5, which has been midtrained with dynamic tiling, which shows \PElang{}'s strong generality for different input formats. 

Next, in Tab.~\ref{tab:lang_refcoco},~\ref{tab:lang_refcoco_qwen2},~\ref{tab:lang_refcoco_tiling}, we show the breakdowns of RefCOCO/+/g~\cite{kazemzadeh2014referitgame} with Llama 3.1-instruct 8B as language model, Qwen2.5 LM 7B as language model, and with Llama 3.1-instruct 8B and dynamic tiling ($4+1$), respectively.
In our SFT data, we have VisualGenome~\cite{krishna2017visual}, DCI~\cite{Urbanek_2024_CVPR}, and Flickr30K~\cite{plummer2015flickr30k} as grounding datasets, and RefCOCO/+/g are unseen. We therefore report zeroshot performance of the MLLMs to evaluate spatial understanding capability of the vision encoders.
Overall, \PElang{} L or G show the best performance across all RefCOCO splits, except with Qwen2.5 LM. 
This is because (1) InternViT 2.5 6B is midtrained with Qwen2 LM, and (2) during pre/mid-training the training data of RefCOCO/+/g are seen.

\begin{table*}[ht]
    \centering
    \makebox[\linewidth][c]{
    \tablestyle{0pt}{1.05} 
        \begin{tabular}{y{50}wx{20} awwww awwww awww a awwwwww}
        \shline
        \multirow{2}{*}{\vspace{-2.2cm} Model}  &&& \multicolumn{5}{c}{\ct[c3]{\it OCR / Chart / Doc. Q\&A}} %
        & \multicolumn{5}{c}{\ct[c4]{\it Visual Q\&A}} & \multicolumn{4}{c}{\ct[c5]{\it Captioning}} & \multicolumn{1}{c}{\ct[c6]{}} & \multicolumn{7}{c}{\ct[c7]{\it Video}}\\
            & \cb{Encoder Params}{}
            & \cb{Resolution}{Patch Size}
            & \cb[c3]{\textit{\textbf{Avg. OCR QA}}}{}
            & \cb[c3]{ChartQA}{Acc.~\cite{zheng2024chartqa}}
            & \cb[c3]{DocVQA}{Acc.~\cite{mathew2021docvqa}}
            & \cb[c3]{Info. QA}{Acc.~\cite{mathew2022infographicvqa}}
            & \cb[c3]{AI2D}{Acc.~\cite{kembhavi2016ai2d}}
            & \cb[c4]{\textit{\textbf{Avg. VQA}}}{}
            & \cb[c4]{TextVQA}{Acc.~\cite{singh2019textvqa}}
            & \cb[c4]{OK-VQA}{Acc.~\cite{schwenk2022okvqa}}
            & \cb[c4]{POPE}{Acc. ~\cite{li2023popebenchmark}}
            & \cb[c4]{VQAv2}{Acc.~\cite{goyal2017vqav2}}
            & \cb[c5]{\textit{\textbf{Avg. Cap.}}}{}
            & \cb[c5]{Flicker}{CIDEr~\cite{flickr}}
            & \cb[c5]{COCO}{CIDEr ~\cite{coco}}
            & \cb[c5]{No Cap}{CIDEr~\cite{agrawal2019nocaps}}
            & \cb[c6]{\textit{\textbf{Avg. Ground.}}}{RefCOCO/g/+~\cite{kazemzadeh2014referitgame}} 
            & \cb[c7]{\textit{\textbf{Avg. Video}}}{}
            & \cb[c7]{VideoMME}{Acc.~\cite{fu2024videomme}}
            & \cb[c7]{STAR}{Acc.~\cite{wu2021star}}
            & \cb[c7]{TGIF-QA}{Acc.~\cite{jang2017tgif}}
            & \cb[c7]{EgoSchema}{Acc.~\cite{mangalam2024egoschema}}
            & \cb[c7]{MVBench}{Acc.~\cite{li2024mvbench}}
            & \cb[c7]{PerceptionTest}{Acc.~\cite{patraucean2024perceptiontest}}            \\
        \hline
\multicolumn{1}{l}{{\textit{256 Tokens per Tile}}}                & & & \ca{} &&&&& \ca{} &&&&& \ca{} &&&& \cat{} & \cat{} &&&&&& \\

MetaCLIP-L~\cite{metaclip}                  & 0.3B           & \rp{224}{14} & \ca{61.8} & 71.1 & 62.5 & 40.2 & 73.3 & \ca{74.6} & 65.3 & 64.9 & 88.5 & 79.8 & \ca{113.4} & 90.4 & 133.5 & 116.2 & \ca{67.1} & \ca{48.0} & 44.8 & 47.1 & 62.7 & 39.0 & 46.0 & 48.3 \\
MetaCLIP-G~\cite{metaclip}                  & 1.8B             & \rp{224}{14} & \ca{60.3} & 68.1 & 61.3 & 39.1 & 72.8 & \ca{74.9} & 65.4 & 65.9 & 88.2 & 80.1 & \ca{114.2} & 91.8 & 134.4 & 116.5 & \ca{66.0} & \ca{49.0} & 46.5 & 46.5 & 62.5 & 45.0 & 44.7 & 48.9 \\
\textbf{\PElang{} G}$^\dagger$                 & \,\,\,1.7B$^*$ & \rp{224}{14} & \ca{70.2} & 79.8 & 79.1 & 47.5 & 74.6 & \ca{76.0} & 70.6 & 64.3 & 88.3 & 80.6 & \ca{116.3} & 92.0 & 136.4 & 120.5 & \ca{69.5} & \ca{56.6} & 49.0 & 55.9 & 69.9 & 61.2 & 50.0 & 53.6 \\
 \hline
\multicolumn{1}{l}{{\textit{576 Tokens per Tile}}}                & & & \ca{} &&&&& \ca{} &&&&& \ca{} &&&& \cat{} & \cat{} &&&&&& \\
CLIP~\cite{clip}                          & 0.3B           & \rp{336}{14} & \ca{69.6} & 76.8 & 78.2 & 50.3 & 72.9 & \ca{76.3} & 71.8 & 64.9 & 88.0 & 80.4 & \ca{114.0} & 90.9 & 134.4 & 116.6 &  \ca{68.5} & \ca{50.8} & 46.6 & 52.2 & 65.0 & 44.6 & 46.3 & 49.9 \\
AIMv2-L~\cite{aimv2}                        & 0.3B           & \rp{336}{14} & \ca{66.7} & 74.1 & 74.9 & 45.2 & 72.4 & \ca{77.4} & 73.5 & 65.6 & 89.0 & 81.7 & \ca{116.4} & 92.5 & 137.1 & 119.5 & \ca{66.6} &  \ca{54.1} & 43.4 & 54.3 & 70.6 & 56.0 & 47.3 & 52.7 \\
SigLIP2-so~\cite{siglip2}                 & 0.4B           & \rp{384}{16} & \ca{55.5} & 61.4 & 54.9 & 33.3 & 72.3 & \ca{76.5} & 70.1 & 66.0 & 88.6 & 81.2 & \ca{118.0} & \textbf{95.8} & 138.3 & 119.8 & \ca{66.5} &  \ca{54.3} & 44.9 & 52.8 & 66.8 & 58.6 & 49.6 & 53.3 \\
SigLIP2-g-opt~\cite{siglip2}                 & 1.1B             & \rp{384}{16} & \ca{56.2} & 63.1 & 55.3 & 34.0 & 72.4 & \ca{77.0} & 70.3 & \textbf{66.7} & 89.6 & 81.6 & \ca{117.7} & 94.9 & 137.8 & 120.3 & \ca{66.5} &  \ca{53.9} & 46.2 & 53.9 & 66.6 & 53.8 & 48.5 & 54.7 \\
\textbf{\PElang{} G}$^\dagger$                 & \,\,\,1.7B$^*$ & \rp{336}{14} & \ca{77.5} & 82.1 & 88.5 & 61.8 & 77.4 & \ca{79.7} & 80.2 & 66.4 & 89.8 & 82.5 & \ca{120.3} & 97.4 & 140.2 & 123.2 & \ca{71.9} & \ca{59.8} & 49.4 & 62.7 & 74.1 & 64.0 & 53.1 & 55.6 \\
\hline
\multicolumn{1}{l}{{\textit{1024 Tokens per Tile}}}                & & & \ca{} &&&&& \ca{} &&&&& \ca{} &&&& \cat{} & \cat{} &&&&&& \\
SigLIP2-so~\cite{siglip2}                 & 0.4B           & \rp{512}{16} & \ca{56.9} & 66.0 & 56.5 & 34.3 & 70.9 & \ca{76.4} & 69.9 & 66.2 & 88.4 & 81.2 & \ca{117.8} & 94.7 & 137.8 & 120.9 & \ca{67.8} &   \ca{46.2} & 47.0 & 44.9 & 66.7 & 39.2 & 34.5 & 45.1 \\
\textbf{\PEcore{L}} & 0.3B & \rp{448}{14}                                & \ca{67.1} & 72.4 & 78.3 & 46.4 & 71.2 & \ca{76.4} & 74.0 & 63.7 & 88.8 & 79.0 & \ca{113.9} & 91.5 & 134.5 & 115.7 & \ca{62.9} &   \ca{51.4} & 47.0 & 51.2 & 62.7 & 49.6 & 47.8 & 50.1 \\
\textbf{\PElang{L}} & 0.3B & \rp{448}{14}                                & \ca{78.3} & 82.8 & 89.3 & 65.2 & 75.9 & \ca{78.5} & 78.8 & 64.4 & 89.6 & 81.3 & \ca{117.8} & 94.7 & 138.1 & 120.7 & \ca{71.6} & \ca{56.5} & 47.0 & 57.2 & 68.0 & 59.8 & 52.3 & 54.7 \\
\hline
AIMv2 3B~\cite{aimv2}                     & 2.7B             & \rp{448}{14} & \ca{67.5} & 73.0 & 78.2 & 46.5 & 72.2 & \ca{78.8} & 79.2 & 66.2 & 88.3 & 81.7 & \ca{119.0} & \textbf{95.8} & 139.7 & 121.5 &  \ca{65.1} & \ca{54.0} & \textbf{49.6} & 55.4 & 67.3 & 49.6 & 49.9 & 52.5 \\
InternViT2.5 6B~\cite{chen2024internvit2p5}  & 5.5B             & \rp{448}{14} & \ca{67.4} & 74.6 & 74.3 & 47.6 & 72.9 & \ca{75.9} & 71.3 & 64.8 & 87.7 & 79.7 & \ca{110.4} & 85.3 & 132.5 & 113.5 &  \ca{56.8} & \ca{52.0} & 46.0 & 49.6 & 65.0 & 50.6 & 49.6 & 51.3 \\
\textbf{\PEcore{G}}                       & 1.9B           & \rp{448}{14} & \ca{68.0} & 73.4 & 81.2 & 47.6 & 69.7 & \ca{76.4} & 74.3 & 62.5 & 89.1 & 79.6 & \ca{113.0} & 91.6 & 134.5 & 112.9 &  \ca{67.6} & \ca{53.2} & 46.0 & 54.3 & 67.0 & 51.2 & 48.7 & 52.0 \\
\textbf{\PElang{G}}                       & \,\,\,1.7B$^*$ & \rp{448}{14} & \ca{\textbf{78.6}} & \textbf{81.8} & \textbf{89.8} & \textbf{67.8} & \textbf{75.0} & \ca{\textbf{80.3}} & \textbf{82.3} & \textbf{66.7} & \textbf{89.6} & \textbf{82.8} & \ca{\textbf{119.6}} & 95.2 & \textbf{140.3} & \textbf{123.4} &  \ca{\textbf{71.8}} & \ca{\textbf{59.0}} & \textbf{49.6} & \textbf{61.8} & \textbf{73.9} & \textbf{60.0} & \textbf{52.6} & \textbf{56.3} \\
        \shline
    \end{tabular}
    }
    \caption{{\bf 4+1 Tile Llama 8B MLLM Results.} Llama 3.1-instruct 8B~\cite{llama3} is used as a language model.  $^*$\PElang{} has 1.7B parameters since we discard the last 3 layers during language alignment. 
    All MLLMs are trained with dynamic tiling for different image sizes and aspect ratio. 
    We use up to 4 image tiles of $448 \times 448$ (or the corresponding resolution for each encoder).
    The image tiles follow after a \textit{thumbnail} input, similar to prior work~\cite{liu2024llavanext}. $^\dagger$Evaluation on an model that was interpolated without additional training (i.e., \textit{zero-shot} resolution).}
    \label{tab:lang_mllm_bench_tiling}
\end{table*}

\begin{table}[ht]
    \centering
    \makebox[\linewidth][c]{
    \tablestyle{0pt}{1.05} 
    \begin{tabular}{y{50}www awwwwww awwwwww awwwwwwww}
        \shline
        \multirow{2}{*}{\vspace{-2.2cm} Model}  &&& \multicolumn{9}{c}{\ct[c6]{\it Grounding}}\\
            & \cb{Encoder Params}{}
            & \cb{Resolution}{Patch Size}
            & \cb[c6]{\textit{\textbf{Avg. Ground.}}}{}
            & \cb[c6]{RefCOCO}{val~\cite{kazemzadeh2014referitgame}}
            & \cb[c6]{RefCOCO}{testA~\cite{kazemzadeh2014referitgame}}
            & \cb[c6]{RefCOCO}{testB~\cite{kazemzadeh2014referitgame}}
            & \cb[c6]{RefCOCO+}{val~\cite{kazemzadeh2014referitgame}}
            & \cb[c6]{RefCOCO+}{testA~\cite{kazemzadeh2014referitgame}}
            & \cb[c6]{RefCOCO+}{testB~\cite{kazemzadeh2014referitgame}}
            & \cb[c6]{RefCOCOg}{val~\cite{kazemzadeh2014referitgame}}
            & \cb[c6]{RefCOCOg}{test~\cite{kazemzadeh2014referitgame}}
            \\
        \hline
\multicolumn{1}{l}{{\textit{256 Tokens per Image}}}                & & & \cat{} &&& &&& &&  \\
MetaCLIP-L~\cite{metaclip}                  & 0.3B & \rp{224}{14} & \ca{60.6} & 63.6 & 56.7 & 67.5 & 54.1 & 58.9 & 48.8 & 67.2 & 67.8 \\
MetaCLIP-G~\cite{metaclip}                  & 1.8B & \rp{224}{14} & \ca{60.5} & 62.0 & 56.5 & 67.8 & 53.5 & 58.7 & 49.2 & 68.2 & 68.3 \\
\textbf{\PElang{} G}$^\dagger$                 & \,\,\,1.7B$^*$ & \rp{224}{14} & \ca{65.7} & 67.7 & 64.4 & 70.9 & 58.3 & 62.0 & 56.6 & 73.2 & 74.4 \\
\hline
\multicolumn{1}{l}{{\textit{576 Tokens per Image}}}                & & & \cat{} &&& &&& &&  \\
CLIP~\cite{clip}                            & 0.3B & \rp{336}{14} & \ca{65.0} & 66.7 & 61.4 & 71.6 & 57.6 & 62.5 & 54.5 & 73.2 & 72.8 \\
AIMv2-L~\cite{aimv2}                        & 0.3B & \rp{336}{14} & \ca{63.3} & 65.4 & 61.6 & 69.6 & 55.0 & 60.0 & 52.0 & 71.1 & 71.5 \\
AIMv2-L Dist. ~\cite{aimv2}                 & 0.3B & \rp{336}{14} & \ca{62.6} & 64.8 & 61.0 & 69.4 & 54.4 & 59.0 & 51.3 & 70.8 & 70.0 \\
SigLIP2-so~\cite{siglip2}                   & 0.4B & \rp{384}{16} & \ca{67.4} & 68.8 & 66.5 & 71.0 & 60.3 & 61.8 & 58.5 & 76.2 & 76.0 \\
SigLIP2-g-opt~\cite{siglip2}                & 1.1B & \rp{384}{16} & \ca{66.5} & 67.9 & 66.1 & 70.1 & 58.8 & 61.7 & 57.1 & 75.5 & 75.0 \\
\textbf{\PElang{} G}$^\dagger$                 & \,\,\,1.7B$^*$ & \rp{336}{14} & \ca{68.9} & 69.8 & 67.5 & 73.2 & 61.5 & 64.0 & 60.8 & 77.3 & 77.7 \\
\hline
\multicolumn{1}{l}{{\textit{1024 Tokens per Image}}}                & & & \cat{} &&& &&& &&  \\
InternViT2.5 L~\cite{chen2024internvit2p5}  & 0.3B & \rp{448}{14} & \ca{66.9} & 69.3 & 66.7 & 72.6 & 58.3 & 63.1 & 57.2 & 74.2 & 74.0 \\
SigLIP2-so~\cite{siglip2}                   & 0.4B & \rp{512}{16} & \ca{69.6} & 71.4 & 69.2 & 74.4 & 61.3 & 64.8 & 60.3 & 77.9 & 77.2 \\
\textbf{\PEcore{} L}                        & 0.3B & \rp{448}{14} & \ca{59.7} & 61.7 & 55.3 & 66.9 & 53.1 & 58.8 & 48.0 & 68.5 & 67.5 \\
\textbf{\PElang{} L}                        & 0.3B & \rp{448}{14} & \ca{70.5} & 71.8 & \textbf{70.2} & 73.0 & 63.7 & 66.1 & 62.7 & 78.8 & 78.9 \\
\hline
    \addpadding
DINOv2~\cite{dinov2}                        & 1.1B & \rp{448}{14} & \ca{64.9} & 67.2 & 62.5 & 70.5 & 57.0 & 61.0 & 54.5 & 73.1 & 73.1 \\
AIMv2 3B~\cite{aimv2}                       & 2.7B & \rp{448}{14} & \ca{36.1} & 37.6 & 34.1 & 40.7 & 32.7 & 36.2 & 32.0 & 36.9 & 38.6 \\
InternViT2.5 6B~\cite{chen2024internvit2p5} & 5.5B & \rp{448}{14} & \ca{68.0} & 70.2 & 67.6 & 72.2 & 60.6 & 64.0 & 58.7 & 75.3 & 75.2 \\
\textbf{\PEcore{} G}                        & 1.9B & \rp{448}{14} & \ca{66.6} & 68.3 & 64.4 & 72.3 & 58.7 & 62.7 & 56.0 & 75.1 & 75.0 \\
\textbf{\PElang{} G}                        & \,\,\,1.7B$^*$ & 448/14 & \ca{\textbf{71.3}} & \textbf{71.9} & 69.9 & \textbf{75.1} & \textbf{64.2} & \textbf{67.3} & \textbf{63.0} & \textbf{79.4} & \textbf{79.2} \\
        \shline
    \end{tabular}
    }
    \caption{{\bf Llama MLLM-Based Zeroshot RefCOCO.} Llama 3.1-instruct 8B~\cite{llama3} is used for zeroshot RefCOCO/+/g grounding. }
    \label{tab:lang_refcoco}
\end{table}

\begin{table}[ht]
        \centering
        \makebox[\linewidth][c]{
        \tablestyle{0pt}{1.05} 
        \begin{tabular}{y{55}www awwwwww awwwwww awwwwwwww}
            \shline
            \multirow{2}{*}{\vspace{-2.2cm} Model}  &&& \multicolumn{9}{c}{\ct[c6]{\it Grounding}}\\
                & \cb{Encoder Params}{}
                & \cb{Resolution}{Patch Size}
                & \cb[c6]{\textit{\textbf{Avg. Ground.}}}{}
                & \cb[c6]{RefCOCO}{val~\cite{kazemzadeh2014referitgame}}
                & \cb[c6]{RefCOCO}{testA~\cite{kazemzadeh2014referitgame}}
                & \cb[c6]{RefCOCO}{testB~\cite{kazemzadeh2014referitgame}}
                & \cb[c6]{RefCOCO+}{val~\cite{kazemzadeh2014referitgame}}
                & \cb[c6]{RefCOCO+}{testA~\cite{kazemzadeh2014referitgame}}
                & \cb[c6]{RefCOCO+}{testB~\cite{kazemzadeh2014referitgame}}
                & \cb[c6]{RefCOCOg}{val~\cite{kazemzadeh2014referitgame}}
                & \cb[c6]{RefCOCOg}{test~\cite{kazemzadeh2014referitgame}}
                \\
            \hline
    \multicolumn{1}{l}{{\textit{576 Tokens per Image}}}                & & & \cat{} &&& &&& &&  \\
    SigLIP2-so~\cite{siglip2}                   & 0.4B & \rp{384}{16} & \ca{70.0} & 73.6 & 73.0 & 74.3 & 60.9 & 62.7 & 59.9 & 78.4 & 77.2 \\
    SigLIP2-g-opt~\cite{siglip2}                & 1.1B & \rp{384}{16} & \ca{69.9} & 73.3 & 72.4 & 73.6 & 60.5 & 62.3 & 60.7 & 78.4 & 78.2 \\
    \textbf{\PElang{} G}$^\dagger$                 & \,\,\,1.7B$^*$ & \rp{336}{14} & \ca{70.1} & 73.4 & 72.0 & 75.3 & 62.0 & 64.2 & 61.2 & 78.4 & 77.7 \\
    \hline
    \multicolumn{1}{l}{{\textit{1024 Tokens per Image}}}                & & & \cat{} &&& &&& &&  \\
    InternViT2.5 L~\cite{chen2024internvit2p5}  & 0.3B & \rp{448}{14} & \ca{68.1} & 72.4 & 69.1 & 74.1 & 59.3 & 62.4 & 56.6 & 75.2 & 75.5 \\
    SigLIP2-so~\cite{siglip2}                   & 0.4B & \rp{512}{16} & \ca{70.5} & 74.1 & 73.7 & 74.4 & 61.7 & 62.9 & 61.0 & 78.6 & 77.9 \\
    \textbf{\PEcore{L}}                         & 0.3B & \rp{448}{14} & \ca{66.5} & 70.4 & 67.8 & 71.5 & 57.7 & 61.1 & 56.2 & 75.8 & 75.3 \\
    \textbf{\PElang{L}}                         & 0.3B & \rp{448}{14} & \ca{70.4} & 74.4 & 72.6 & 74.6 & 62.2 & 64.0 & 62.0 & 79.0 & 78.7 \\
    \hline
    \addpadding 
    DINOv2~\cite{dinov2}                        & 1.1B & \rp{448}{14} & \ca{69.3} & 73.4 & 71.1 & 73.9 & 60.0 & 63.9 & 59.0 & 76.4 & 76.7 \\
    AIMv2 3B~\cite{aimv2}                       & 2.7B & \rp{448}{14} & \ca{67.6} & 71.4 & 67.7 & 72.3 & 59.2 & 61.2 & 56.3 & 76.4 & 76.4 \\
    InternViT2.5 6B$^\ddagger$~\cite{chen2024internvit2p5} & 5.5B   & \rp{448}{14} & \ca{\textbf{72.8}} & \textbf{77.7} & \textbf{76.5} & \textbf{77.1} & 63.6 & \textbf{66.0} & 62.2 & \textbf{80.0} & 79.5 \\
    \textbf{\PEcore{G}}                         & 1.9B & \rp{448}{14} & \ca{70.5} & 74.0 & 71.8 & 75.8 & 61.5 & 64.8 & 60.1 & 78.5 & 77.3 \\
    \textbf{\PElang{G}}                         & \,\,\,1.7B$^*$ & \rp{448}{14} & \ca{72.1} & 75.4 & 72.9 & 76.3 & \textbf{64.2} & 65.9 & \textbf{62.9} & 79.7 & \textbf{79.7} \\
            \shline
        \end{tabular}
        }
        \caption{{\bf Qwen MLLM-Based Zeroshot RefCOCO.} QwenLM 2.5 7B~\cite{qwen2.5} is used as a language model. All MLLMs report zeroshot results on RefCOCO/+/g datasets. $^\ddagger$Trained with RefCOCO/+/g beforehand. }
        \label{tab:lang_refcoco_qwen2}
    \end{table}

\begin{table}[ht]
        \centering
        \makebox[\linewidth][c]{
        \tablestyle{0pt}{1.05} 
        \begin{tabular}{y{55}www awwwwww awwwwww awwwwwwww}
            \shline
            \multirow{2}{*}{\vspace{-2.2cm} Model}  &&& \multicolumn{9}{c}{\ct[c6]{\it Grounding}}\\
                & \cb{Encoder Params}{}
                & \cb{Resolution}{Patch Size}
                & \cb[c6]{\textit{\textbf{Avg. Ground.}}}{}
                & \cb[c6]{RefCOCO}{val~\cite{kazemzadeh2014referitgame}}
                & \cb[c6]{RefCOCO}{testA~\cite{kazemzadeh2014referitgame}}
                & \cb[c6]{RefCOCO}{testB~\cite{kazemzadeh2014referitgame}}
                & \cb[c6]{RefCOCO+}{val~\cite{kazemzadeh2014referitgame}}
                & \cb[c6]{RefCOCO+}{testA~\cite{kazemzadeh2014referitgame}}
                & \cb[c6]{RefCOCO+}{testB~\cite{kazemzadeh2014referitgame}}
                & \cb[c6]{RefCOCOg}{val~\cite{kazemzadeh2014referitgame}}
                & \cb[c6]{RefCOCOg}{test~\cite{kazemzadeh2014referitgame}}
                \\
            \hline
    \multicolumn{1}{l}{{\textit{256 Tokens per Tile}}}                & & & \cat{} &&& &&& &&  \\
    MetaCLIP-L~\cite{metaclip}                  & 0.3B & \rp{224}{14} & \ca{67.1} & 69.3 & 65.0 & 73.2 & 60.5 & 64.9 & 56.5 & 74.3 & 73.4 \\ 
    MetaCLIP-G~\cite{metaclip}                  & 1.8B & \rp{224}{14} & \ca{66.0} & 67.9 & 63.2 & 71.9 & 59.2 & 62.9 & 55.8 & 73.8 & 73.1 \\ 
    \textbf{\PElang{} G}$^\dagger$                 & \,\,\,1.7B$^*$ & \rp{224}{14} & \ca{70.3} & 71.6 & 69.6 & 73.7 & 63.3 & 66.2 & 62.6 & 78.6 & 78.2 \\

    \hline
    \multicolumn{1}{l}{{\textit{576 Tokens per Tile}}}                & & & \cat{} &&& &&& &&  \\
    CLIP~\cite{clip}                            & 0.3B & \rp{336}{14} & \ca{68.5} & 70.7 & 66.6 & 74.1 & 61.1 & 65.9 & 58.1 & 76.0 & 75.1 \\ 
    AIMv2-L~\cite{aimv2}                        & 0.3B & \rp{336}{14} & \ca{66.6} & 68.4 & 65.5 & 71.4 & 59.3 & 63.4 & 56.5 & 74.2 & 74.2 \\ 
    SigLIP2-so~\cite{siglip2}                   & 0.4B & \rp{384}{16} & \ca{66.5} & 67.9 & 66.1 & 70.1 & 58.8 & 61.7 & 57.1 & 75.5 & 75.0 \\ 
    SigLIP2-g-opt~\cite{siglip2}                & 1.1B & \rp{384}{16} & \ca{66.5} & 68.2 & 65.6 & 70.1 & 59.0 & 62.3 & 58.0 & 74.8 & 74.0 \\ 
    \textbf{\PElang{} G}$^\dagger$                 & \,\,\,1.7B$^*$ & \rp{336}{14} & \ca{71.9} & 73.6 & 71.5 & 74.9 & 64.8 & 67.3 & 63.9 & 80.4 & 80.6 \\
    \hline
    \multicolumn{1}{l}{{\textit{1024 Tokens per Tile}}}                & & & \cat{} &&& &&& &&  \\
    SigLIP2-so~\cite{siglip2}                   & 0.4B & \rp{512}{16} & \ca{67.8} & 69.2 & 67.8 & 71.2 & 59.9 & 62.5 & 59.0 & 76.9 & 76.0 \\ 
    \textbf{\PEcore{L}}                         & 0.3B & \rp{448}{14} & \ca{62.9} & 65.3 & 59.9 & 69.2 & 56.6 & 62.2 & 52.0 & 70.1 & 70.0 \\ 
    \textbf{\PElang{L}}                         & 0.3B & \rp{448}{14} & \ca{71.6}          & \textbf{73.0} & \textbf{70.8} & 74.3          & \textbf{65.2} & \textbf{67.2} & 62.9          & 79.7          & 79.7 \\ 
                                                                    								
    \hline
    \addpadding
    AIMv2 3B~\cite{aimv2}                       & 2.7B & \rp{448}{14} & \ca{65.1} & 66.9 & 62.9 & 71.1 & 58.1 & 62.4 & 55.6 & 71.8 & 72.2 \\ 
    InternViT2.5 6B$^\ddagger$~\cite{chen2024internvit2p5} & 5.5B  & \rp{448}{14} & \ca{56.8} & 61.0 & 56.4 & 65.8 & 51.0 & 57.0 & 46.1 & 58.0 & 58.9 \\ 
    \textbf{\PEcore{G}}                        & 1.9B & \rp{448}{14} & \ca{67.6} & 69.2 & 65.8 & 72.4 & 59.9 & 64.1 & 58.3 & 75.1 & 75.6 \\ 
    \textbf{\PElang{G}}                        & \,\,\,1.7B$^*$ & \rp{448}{14} & \ca{\textbf{71.8}} & 72.6          & 70.7          & \textbf{74.6} & 64.8          & 66.6          & \textbf{64.6} & \textbf{80.4} & \textbf{80.3} \\ 
            \shline
        \end{tabular}
        }
        \caption{{\bf 4+1 Tile Llama 8B MLLM-Based Zeroshot RefCOCO.} Llama 3.1-instruct 8B~\cite{llama3} is used as a language model.
    All trained with dynamic tiling for different image sizes and aspect ratio. 
    We use up to 4 image tiles of the encoder's native resolution, with a \textit{thumbnail} image in front, similar to prior work~\cite{liu2024llavanext}. 
    $^\ddagger$Trained with RefCOCO/+/g beforehand. }
        \label{tab:lang_refcoco_tiling}
    \end{table}

\clearpage

\subsection{\PEspat{}: Additional Qualitative Results}
\label{appx:more_feature_viz}

\clearpage

{
    \small
    \bibliographystyle{ieeenat_fullname}
    \bibliography{reference}
}

\end{document}

%% file: preamble.tex
\def\bf{\bfseries\sffamily}

\newcommand{\expandref}[2]{\ref{#1}\hyperref[#1]{#2}}

\newcommand{\kindatiny}{\fontsize{6pt}{7.2pt}\selectfont}

\newlength\savewidth\newcommand\shline{\noalign{\global\savewidth\arrayrulewidth
  \global\arrayrulewidth 0.5pt}\hline\noalign{\global\arrayrulewidth\savewidth}}

\newcommand{\tablestyle}[2]{%
    \fontfamily{ptm}\selectfont%
    \let\itold\it%
    \def\it{\itold \fontfamily{ptm}\selectfont}%
    \setlength{\tabcolsep}{#1}\renewcommand{\arraystretch}{#2}\centering\kindatiny%
    \let\citeold\cite%
    \renewcommand{\cite}[1]{\normalfont\fontfamily{ptm}\selectfont\tiny\citeold{##1}}%
}
\newcommand{\bigtablestyle}[2]{%
    \fontfamily{ptm}\selectfont%
    \let\itold\it%
    \def\it{\itold \fontfamily{ptm}\selectfont}%
    \setlength{\tabcolsep}{#1}\renewcommand{\arraystretch}{#2}\centering\footnotesize%
    \let\citeold\cite%
    \renewcommand{\cite}[1]{\normalfont\fontfamily{ptm}\selectfont\footnotesize\citeold{##1}}%
}

\renewcommand{\paragraph}[1]{\vspace{1.25mm}\noindent\textbf{#1}}

\newcolumntype{x}[1]{>{\centering\arraybackslash}p{#1pt}}
\newcolumntype{y}[1]{>{\raggedright\arraybackslash}p{#1pt}}
\newcolumntype{z}[1]{>{\raggedleft\arraybackslash}p{#1pt}}
\newcolumntype{w}{>{\centering\arraybackslash}p{18pt}}
\newcolumntype{a}{>{\centering\arraybackslash}p{16pt}}

\newcommand{\uu}[1]{{#1}}
\newcommand{\bb}[1]{\textbf{#1}}

\newcommand{\rp}[2]{{#1\textcolor{c0-item-text}{\tiny /#2}}}

\definecolor{c0-title-bkg}{HTML}{ffffff}
\definecolor{c0-title-text}{HTML}{000000}
\definecolor{c0-item-bkg}{HTML}{ffffff}
\definecolor{c0-item-text}{HTML}{676767}

\definecolor{c1-title-bkg}{HTML}{d1e2dd}
\definecolor{c1-title-text}{HTML}{005953}
\definecolor{c1-item-bkg}{HTML}{e6efec}
\definecolor{c1-item-text}{HTML}{2d7b6d}

\definecolor{c2-title-bkg}{HTML}{cfe1e1}
\definecolor{c2-title-text}{HTML}{005760}
\definecolor{c2-item-bkg}{HTML}{e4eeed}
\definecolor{c2-item-text}{HTML}{24797b}

\definecolor{c3-title-bkg}{HTML}{cddfe5}
\definecolor{c3-title-text}{HTML}{00536b}
\definecolor{c3-item-bkg}{HTML}{e2ecef}
\definecolor{c3-item-text}{HTML}{287687}

\definecolor{c4-title-bkg}{HTML}{cedce8}
\definecolor{c4-title-text}{HTML}{124e74}
\definecolor{c4-item-bkg}{HTML}{e1eaf1}
\definecolor{c4-item-text}{HTML}{3a7190}

\definecolor{c5-title-bkg}{HTML}{d0d9eb}
\definecolor{c5-title-text}{HTML}{324779}
\definecolor{c5-item-bkg}{HTML}{e1e8f3}
\definecolor{c5-item-text}{HTML}{4d6b97}

\definecolor{c6-title-bkg}{HTML}{d3d5ed}
\definecolor{c6-title-text}{HTML}{493e7b}
\definecolor{c6-item-bkg}{HTML}{e3e5f5}
\definecolor{c6-item-text}{HTML}{61639b}

\definecolor{c7-title-bkg}{HTML}{dad1ed}
\definecolor{c7-title-text}{HTML}{5a3477}
\definecolor{c7-item-bkg}{HTML}{e5e1f5}
\definecolor{c7-item-text}{HTML}{725b99}

\definecolor{c8-title-bkg}{HTML}{ded1ec}
\definecolor{c8-title-text}{HTML}{633273}
\definecolor{c8-item-bkg}{HTML}{ebe2f6}
\definecolor{c8-item-text}{HTML}{7c5997}

\definecolor{c9-title-bkg}{HTML}{e5d1eb}
\definecolor{c9-title-text}{HTML}{6c2f6b}
\definecolor{c9-item-bkg}{HTML}{f0e0f6}
\definecolor{c9-item-text}{HTML}{885591}

\definecolor{c10-title-bkg}{HTML}{ebd1e7}
\definecolor{c10-title-text}{HTML}{722e5f}
\definecolor{c10-item-bkg}{HTML}{f5e2f3}
\definecolor{c10-item-text}{HTML}{915487}

\definecolor{avg-title-bkg}{HTML}{f3f3f3}
\definecolor{avg-title-text}{HTML}{000000}
\definecolor{avg-item-bkg}{HTML}{f3f3f3}
\definecolor{avg-item-text}{HTML}{000000}

\newcommand{\addpadding}{%
  \rule{0pt}{\dimexpr\normalbaselineskip-1pt\relax}%
}

\newcommand{\cc}[2][c0]{\cellcolor{#1-item-bkg}\textcolor{#1-item-text}{\it \tiny{#2}}}
\newcommand{\ct}[2][c0]{\addpadding{\cellcolor{#1-item-bkg}\textcolor{#1-title-text}{#2}}}

\definecolor{promptcolor}{HTML}{D1D0F2}
\definecolor{promptcolorheader}{HTML}{bdbcec}
\newcommand{\promptbox}[2]{
    \begin{tcolorbox}[
        top=0.3em,bottom=0.3em,left=0.5em,right=0.5em,
        toptitle=0.3em,bottomtitle=0.2em,boxsep=0pt,
        colframe=promptcolorheader,colback=promptcolor!50,boxrule=0.5pt,
        title={\footnotesize \fontfamily{zi4}\selectfont #1}
    ]
        \footnotesize
        {\fontfamily{zi4}\selectfont #2}
    \end{tcolorbox}
}

\ExplSyntaxOn

\cs_generate_variant:Nn \coffin_typeset:Nnnnn {Nffff}

\NewDocumentCommand\rotbox{ O{l,H} D<>{0pt,0pt} m m}{
    \hcoffin_set:Nn \l_tmpa_coffin {#4}
    \coffin_rotate:Nn \l_tmpa_coffin {#3}
    \coffin_typeset:Nffff \l_tmpa_coffin 
        {\clist_item:nn{#1}{1}}
        {\clist_item:nn{#1}{2}}
        {\clist_item:nn{#2}{1}}
        {\clist_item:nn{#2}{2}}
}
\ExplSyntaxOff

\DeclareRobustCommand{\PEcore}{PE$_\text{core}$}
\DeclareRobustCommand{\PElang}{PE$_\text{lang}$}
\DeclareRobustCommand{\PEspat}{PE$_\text{spatial}$}

\newlength{\ccustomlen}
\newcommand{\ccustom}[3][c0]{%
    \cellcolor{#1-item-bkg}{%
        \rotbox[l,t]{90}{%
            \parbox[t]{\ccustomlen}{%
                \ifthenelse{\isempty{#3}}{%
                    \mbox{%
                        \kindatiny\textcolor{#1-title-text}{#2}%
                    }%
                }{%
                    \kindatiny\textcolor{#1-title-text}{#2} \\%
                    \tiny{\textcolor{#1-item-text}{\it #3}}%
                }%
            }%
        }%
    }%
}

\newcommand{\cb}[3][c0]{%
    \setlength{\ccustomlen}{1.2cm}%
    \ccustom[#1]{#2}{#3}%
}

\newcommand{\ca}[1]{\cellcolor{avg-item-bkg}{#1}}
\newcommand{\cat}[1]{\addpadding{\cellcolor{avg-item-bkg}{#1}}}

\newcommand{\cmark}{\ding{51}}%
\newcommand{\xmark}{\ding{55}}%

\def\vs{\emph{vs.}\xspace}

\makeatother

%% file: arxiv.bbl
\begin{thebibliography}{169}
\providecommand{\natexlab}[1]{#1}
\providecommand{\url}[1]{\texttt{#1}}
\expandafter\ifx\csname urlstyle\endcsname\relax
  \providecommand{\doi}[1]{doi: #1}\else
  \providecommand{\doi}{doi: \begingroup \urlstyle{rm}\Url}\fi

\bibitem[Agrawal et~al.(2019)Agrawal, Desai, Wang, Chen, Jain, Johnson, Batra, Parikh, Lee, and Anderson]{agrawal2019nocaps}
Harsh Agrawal, Karan Desai, Yufei Wang, Xinlei Chen, Rishabh Jain, Mark Johnson, Dhruv Batra, Devi Parikh, Stefan Lee, and Peter Anderson.
\newblock Nocaps: Novel object captioning at scale.
\newblock In \emph{ICCV}, 2019.

\bibitem[Agrawal et~al.(2024)Agrawal, Antoniak, Hanna, Bout, Chaplot, Chudnovsky, Costa, Monicault, Garg, Gervet, Ghosh, Héliou, Jacob, Jiang, Khandelwal, Lacroix, Lample, Casas, Lavril, Scao, Lo, Marshall, Martin, Mensch, Muddireddy, Nemychnikova, Pellat, Platen, Raghuraman, Rozière, Sablayrolles, Saulnier, Sauvestre, Shang, Soletskyi, Stewart, Stock, Studnia, Subramanian, Vaze, Wang, and Yang]{agrawal2024pixtral}
Pravesh Agrawal, Szymon Antoniak, Emma~Bou Hanna, Baptiste Bout, Devendra Chaplot, Jessica Chudnovsky, Diogo Costa, Baudouin~De Monicault, Saurabh Garg, Theophile Gervet, Soham Ghosh, Amélie Héliou, Paul Jacob, Albert~Q. Jiang, Kartik Khandelwal, Timothée Lacroix, Guillaume Lample, Diego~Las Casas, Thibaut Lavril, Teven~Le Scao, Andy Lo, William Marshall, Louis Martin, Arthur Mensch, Pavankumar Muddireddy, Valera Nemychnikova, Marie Pellat, Patrick~Von Platen, Nikhil Raghuraman, Baptiste Rozière, Alexandre Sablayrolles, Lucile Saulnier, Romain Sauvestre, Wendy Shang, Roman Soletskyi, Lawrence Stewart, Pierre Stock, Joachim Studnia, Sandeep Subramanian, Sagar Vaze, Thomas Wang, and Sophia Yang.
\newblock Pixtral 12b.
\newblock \emph{arXiv:2410.07073}, 2024.

\bibitem[Bai et~al.(2023)Bai, Bai, Yang, Wang, Tan, Wang, Lin, Zhou, and Zhou]{qwen-vl}
Jinze Bai, Shuai Bai, Shusheng Yang, Shijie Wang, Sinan Tan, Peng Wang, Junyang Lin, Chang Zhou, and Jingren Zhou.
\newblock {Qwen-VL}: A versatile vision-language model for understanding, localization, text reading, and beyond.
\newblock \emph{arXiv:2308.12966}, 2023.

\bibitem[Barbu et~al.(2019)Barbu, Mayo, Alverio, Luo, Wang, Gutfreund, Tenenbaum, and Katz]{objectnet}
Andrei Barbu, David Mayo, Julian Alverio, William Luo, Christopher Wang, Dan Gutfreund, Josh Tenenbaum, and Boris Katz.
\newblock {ObjectNet}: A large-scale bias-controlled dataset for pushing the limits of object recognition models.
\newblock In \emph{NeurIPS}, 2019.

\bibitem[Beyer et~al.(2024)Beyer, Steiner, Pinto, Kolesnikov, Wang, Salz, Neumann, Alabdulmohsin, Tschannen, Bugliarello, Unterthiner, Keysers, Koppula, Liu, Grycner, Gritsenko, Houlsby, Kumar, Rong, Eisenschlos, Kabra, Bauer, Bosnjak, Chen, Minderer, Voigtlaender, Bica, Balazevic, Puigcerver, Papalampidi, H{\'{e}}naff, Xiong, Soricut, Harmsen, and Zhai]{paligemma}
Lucas Beyer, Andreas Steiner, Andr{\'{e}}~Susano Pinto, Alexander Kolesnikov, Xiao Wang, Daniel Salz, Maxim Neumann, Ibrahim Alabdulmohsin, Michael Tschannen, Emanuele Bugliarello, Thomas Unterthiner, Daniel Keysers, Skanda Koppula, Fangyu Liu, Adam Grycner, Alexey~A. Gritsenko, Neil Houlsby, Manoj Kumar, Keran Rong, Julian Eisenschlos, Rishabh Kabra, Matthias Bauer, Matko Bosnjak, Xi Chen, Matthias Minderer, Paul Voigtlaender, Ioana Bica, Ivana Balazevic, Joan Puigcerver, Pinelopi Papalampidi, Olivier~J. H{\'{e}}naff, Xi Xiong, Radu Soricut, Jeremiah Harmsen, and Xiaohua Zhai.
\newblock {PaliGemma}: {A} versatile 3b {VLM} for transfer.
\newblock \emph{arXiv:2407.07726}, 2024.

\bibitem[Bodla et~al.(2017)Bodla, Singh, Chellappa, and Davis]{softnms}
Navaneeth Bodla, Bharat Singh, Rama Chellappa, and Larry~S Davis.
\newblock {Soft-NMS}--{I}mproving object detection with one line of code.
\newblock In \emph{ICCV}, 2017.

\bibitem[Bolya et~al.(2023)Bolya, Ryali, Hoffman, and Feichtenhofer]{abswin}
Daniel Bolya, Chaitanya Ryali, Judy Hoffman, and Christoph Feichtenhofer.
\newblock Window attention is bugged: how not to interpolate position embeddings.
\newblock In \emph{ICLR}, 2023.

\bibitem[Bordes et~al.(2022)Bordes, Balestriero, Garrido, Bardes, and Vincent]{bordes2022guillotine}
Florian Bordes, Randall Balestriero, Quentin Garrido, Adrien Bardes, and Pascal Vincent.
\newblock Guillotine regularization: Why removing layers is needed to improve generalization in self-supervised learning.
\newblock \emph{arXiv:2206.13378}, 2022.

\bibitem[Bossard et~al.(2014)Bossard, Guillaumin, and Van~Gool]{food101}
Lukas Bossard, Matthieu Guillaumin, and Luc Van~Gool.
\newblock Food-101 -- {M}ining discriminative components with random forests.
\newblock In \emph{ECCV}, 2014.

\bibitem[Bradski(2000)]{opencv}
Gary Bradski.
\newblock The {OpenCV} library.
\newblock \emph{Dr. Dobb's Journal: Software Tools for the Professional Programmer}, 2000.

\bibitem[Cai and Vasconcelos(2018)]{cascadercnn}
Zhaowei Cai and Nuno Vasconcelos.
\newblock Cascade {R-CNN}: Delving into high quality object detection.
\newblock In \emph{CVPR}, 2018.

\bibitem[Carion et~al.(2020)Carion, Massa, Synnaeve, Usunier, Kirillov, and Zagoruyko]{carion2020detr}
Nicolas Carion, Francisco Massa, Gabriel Synnaeve, Nicolas Usunier, Alexander Kirillov, and Sergey Zagoruyko.
\newblock End-to-end object detection with transformers.
\newblock In \emph{ECCV}, 2020.

\bibitem[Chai et~al.(2025)Chai, Song, Du, Meng, Madhavan, Bar-Tal, Hwang, Xie, and Manning]{auroracap}
Wenhao Chai, Enxin Song, Yilun Du, Chenlin Meng, Vashisht Madhavan, Omer Bar-Tal, Jeng-Neng Hwang, Saining Xie, and Christopher~D. Manning.
\newblock {AuroraCap}: Efficient, performant video detailed captioning and a new benchmark.
\newblock In \emph{ICLR}, 2025.

\bibitem[Chen et~al.(2019)Chen, Pang, Wang, Xiong, Li, Sun, Feng, Liu, Shi, Ouyang, Loy, and Lin]{htc}
Kai Chen, Jiangmiao Pang, Jiaqi Wang, Yu Xiong, Xiaoxiao Li, Shuyang Sun, Wansen Feng, Ziwei Liu, Jianping Shi, Wanli Ouyang, Chen~Change Loy, and Dahua Lin.
\newblock Hybrid task cascade for instance segmentation.
\newblock In \emph{CVPR}, 2019.

\bibitem[Chen et~al.(2020{\natexlab{a}})Chen, Radford, Child, Wu, Jun, Luan, and Sutskever]{igpt}
Mark Chen, Alec Radford, Rewon Child, Jeffrey Wu, Heewoo Jun, David Luan, and Ilya Sutskever.
\newblock Generative pretraining from pixels.
\newblock In \emph{ICML}, 2020{\natexlab{a}}.

\bibitem[Chen et~al.(2020{\natexlab{b}})Chen, Kornblith, Norouzi, and Hinton]{chen2020simclr}
Ting Chen, Simon Kornblith, Mohammad Norouzi, and Geoffrey Hinton.
\newblock A simple framework for contrastive learning of visual representations.
\newblock In \emph{ICML}, 2020{\natexlab{b}}.

\bibitem[Chen et~al.(2023)Chen, Wang, Changpinyo, Piergiovanni, Padlewski, Salz, Goodman, Grycner, Mustafa, Beyer, Kolesnikov, Puigcerver, Ding, Rong, Akbari, Mishra, Xue, Thapliyal, Bradbury, Kuo, Seyedhosseini, Jia, Ayan, Riquelme, Steiner, Angelova, Zhai, Houlsby, and Soricut]{pali}
Xi Chen, Xiao Wang, Soravit Changpinyo, AJ Piergiovanni, Piotr Padlewski, Daniel Salz, Sebastian Goodman, Adam Grycner, Basil Mustafa, Lucas Beyer, Alexander Kolesnikov, Joan Puigcerver, Nan Ding, Keran Rong, Hassan Akbari, Gaurav Mishra, Linting Xue, Ashish Thapliyal, James Bradbury, Weicheng Kuo, Mojtaba Seyedhosseini, Chao Jia, Burcu~Karagol Ayan, Carlos Riquelme, Andreas Steiner, Anelia Angelova, Xiaohua Zhai, Neil Houlsby, and Radu Soricut.
\newblock Pali: A jointly-scaled multilingual language-image model.
\newblock In \emph{ICLR}, 2023.

\bibitem[Chen et~al.(2024{\natexlab{a}})Chen, Wang, Cao, Liu, Gao, Cui, Zhu, Ye, Tian, Liu, Gu, Wang, Li, Ren, Chen, Luo, Wang, Jiang, Wang, He, Shi, Zhang, Lv, Wang, Shao, Chu, Tu, He, Wu, Deng, Ge, Chen, Zhang, Wang, Dou, Lu, Zhu, Lu, Lin, Qiao, Dai, and Wang]{chen2024internvit2p5}
Zhe Chen, Weiyun Wang, Yue Cao, Yangzhou Liu, Zhangwei Gao, Erfei Cui, Jinguo Zhu, Shenglong Ye, Hao Tian, Zhaoyang Liu, Lixin Gu, Xuehui Wang, Qingyun Li, Yimin Ren, Zixuan Chen, Jiapeng Luo, Jiahao Wang, Tan Jiang, Bo Wang, Conghui He, Botian Shi, Xingcheng Zhang, Han Lv, Yi Wang, Wenqi Shao, Pei Chu, Zhongying Tu, Tong He, Zhiyong Wu, Huipeng Deng, Jiaye Ge, Kai Chen, Kaipeng Zhang, Limin Wang, Min Dou, Lewei Lu, Xizhou Zhu, Tong Lu, Dahua Lin, Yu Qiao, Jifeng Dai, and Wenhai Wang.
\newblock Expanding performance boundaries of open-source multimodal models with model, data, and test-time scaling.
\newblock \emph{arXiv:2412.05271}, 2024{\natexlab{a}}.

\bibitem[Chen et~al.(2024{\natexlab{b}})Chen, Wu, Wang, Su, Chen, Xing, Zhong, Zhang, Zhu, Lu, Li, Luo, Lu, Qiao, and Dai]{internvl}
Zhe Chen, Jiannan Wu, Wenhai Wang, Weijie Su, Guo Chen, Sen Xing, Muyan Zhong, Qinglong Zhang, Xizhou Zhu, Lewei Lu, Bin Li, Ping Luo, Tong Lu, Yu Qiao, and Jifeng Dai.
\newblock {InternVL}: Scaling up vision foundation models and aligning for generic visual-linguistic tasks.
\newblock In \emph{CVPR}, 2024{\natexlab{b}}.

\bibitem[Cheng et~al.(2017)Cheng, Han, and Lu]{cheng_2017_resisc}
Gong Cheng, Junwei Han, and Xiaoqiang Lu.
\newblock Remote sensing image scene classification: Benchmark and state of the art.
\newblock \emph{Proceedings of the IEEE}, 2017.

\bibitem[Cho et~al.(2025)Cho, Madotto, Mavroudi, Afouras, Nagarajan, Maaz, Song, Ma, Hu, Rasheed, Sun, Huang, Bolya, Jain, Martin, Wang, Ravi, Jain, Stark, Moon, Damavandi, Lee, Westbury, Khan, Kr\"{a}henb\"{u}hl, Doll{\'a}r, Torresani, Grauman, and Feichtenhofer]{PLM}
Jang~Hyun Cho, Andrea Madotto, Effrosyni Mavroudi, Triantafyllos Afouras, Tushar Nagarajan, Muhammad Maaz, Yale Song, Tengyu Ma, Shuming Hu, Hanoona Rasheed, Peize Sun, Po-Yao Huang, Daniel Bolya, Suyog Jain, Miguel Martin, Huiyu Wang, Nikhila Ravi, Shashank Jain, Temmy Stark, Shane Moon, Babak Damavandi, Vivian Lee, Andrew Westbury, Salman Khan, Philipp Kr\"{a}henb\"{u}hl, Piotr Doll{\'a}r, Lorenzo Torresani, Kristen Grauman, and Christoph Feichtenhofer.
\newblock Perceptionlm: Open-access data and models for detailed visual understanding.
\newblock \emph{arXiv:2504.13180}, 2025.

\bibitem[Cho et~al.(2024)Cho, Shin, Hong, Arnab, Seo, and Kim]{cho2024catseg}
Seokju Cho, Heeseong Shin, Sunghwan Hong, Anurag Arnab, Paul~Hongsuck Seo, and Seungryong Kim.
\newblock {CAT-Seg}: Cost aggregation for open-vocabulary semantic segmentation.
\newblock In \emph{CVPR}, 2024.

\bibitem[Darcet et~al.(2024)Darcet, Oquab, Mairal, and Bojanowski]{vitsneedregisters}
Timoth{\'e}e Darcet, Maxime Oquab, Julien Mairal, and Piotr Bojanowski.
\newblock Vision transformers need registers.
\newblock In \emph{ICLR}, 2024.

\bibitem[Dehghani et~al.(2023)Dehghani, Djolonga, Mustafa, Padlewski, Heek, Gilmer, Steiner, Caron, Geirhos, Alabdulmohsin, Jenatton, Beyer, Tschannen, Arnab, Wang, Riquelme, Minderer, Puigcerver, Evci, Kumar, van Steenkiste, Elsayed, Mahendran, Yu, Oliver, Huot, Bastings, Collier, Gritsenko, Birodkar, Vasconcelos, Tay, Mensink, Kolesnikov, Pavetić, Tran, Kipf, Lučić, Zhai, Keysers, Harmsen, and Houlsby]{vit22b}
Mostafa Dehghani, Josip Djolonga, Basil Mustafa, Piotr Padlewski, Jonathan Heek, Justin Gilmer, Andreas Steiner, Mathilde Caron, Robert Geirhos, Ibrahim Alabdulmohsin, Rodolphe Jenatton, Lucas Beyer, Michael Tschannen, Anurag Arnab, Xiao Wang, Carlos Riquelme, Matthias Minderer, Joan Puigcerver, Utku Evci, Manoj Kumar, Sjoerd van Steenkiste, Gamaleldin~F. Elsayed, Aravindh Mahendran, Fisher Yu, Avital Oliver, Fantine Huot, Jasmijn Bastings, Mark~Patrick Collier, Alexey Gritsenko, Vighnesh Birodkar, Cristina Vasconcelos, Yi Tay, Thomas Mensink, Alexander Kolesnikov, Filip Pavetić, Dustin Tran, Thomas Kipf, Mario Lučić, Xiaohua Zhai, Daniel Keysers, Jeremiah Harmsen, and Neil Houlsby.
\newblock Scaling vision transformers to 22 billion parameters.
\newblock In \emph{ICML}, 2023.

\bibitem[Deitke et~al.(2024)Deitke, Clark, Lee, Tripathi, Yang, Park, Salehi, Muennighoff, Lo, Soldaini, Lu, Anderson, Bransom, Ehsani, Ngo, Chen, Patel, Yatskar, Callison-Burch, Head, Hendrix, Bastani, VanderBilt, Lambert, Chou, Chheda, Sparks, Skjonsberg, Schmitz, Sarnat, Bischoff, Walsh, Newell, Wolters, Gupta, Zeng, Borchardt, Groeneveld, Nam, Lebrecht, Wittlif, Schoenick, Michel, Krishna, Weihs, Smith, Hajishirzi, Girshick, Farhadi, and Kembhavi]{molmo}
Matt Deitke, Christopher Clark, Sangho Lee, Rohun Tripathi, Yue Yang, Jae~Sung Park, Mohammadreza Salehi, Niklas Muennighoff, Kyle Lo, Luca Soldaini, Jiasen Lu, Taira Anderson, Erin Bransom, Kiana Ehsani, Huong Ngo, YenSung Chen, Ajay Patel, Mark Yatskar, Chris Callison-Burch, Andrew Head, Rose Hendrix, Favyen Bastani, Eli VanderBilt, Nathan Lambert, Yvonne Chou, Arnavi Chheda, Jenna Sparks, Sam Skjonsberg, Michael Schmitz, Aaron Sarnat, Byron Bischoff, Pete Walsh, Chris Newell, Piper Wolters, Tanmay Gupta, Kuo-Hao Zeng, Jon Borchardt, Dirk Groeneveld, Crystal Nam, Sophie Lebrecht, Caitlin Wittlif, Carissa Schoenick, Oscar Michel, Ranjay Krishna, Luca Weihs, Noah~A. Smith, Hannaneh Hajishirzi, Ross Girshick, Ali Farhadi, and Aniruddha Kembhavi.
\newblock Molmo and pixmo: Open weights and open data for state-of-the-art multimodal models.
\newblock \emph{arXiv:2409.17146}, 2024.

\bibitem[Deng et~al.(2009)Deng, Dong, Socher, Li, Li, and Fei-Fei]{imagenet}
Jia Deng, Wei Dong, Richard Socher, Li-Jia Li, Kai Li, and Li Fei-Fei.
\newblock {ImageNet}: A large-scale hierarchical image database.
\newblock In \emph{CVPR}, 2009.

\bibitem[Desai and Johnson(2021)]{desai2021virtex}
Karan Desai and Justin Johnson.
\newblock {VirTex}: Learning visual representations from textual annotations.
\newblock In \emph{CVPR}, 2021.

\bibitem[Ding et~al.(2022)Ding, Xue, Xia, and Dai]{ding2023zss}
Jian Ding, Nan Xue, Gui{-}Song Xia, and Dengxin Dai.
\newblock Decoupling zero-shot semantic segmentation.
\newblock In \emph{CVPR}, 2022.

\bibitem[Dosovitskiy et~al.(2020)Dosovitskiy, Beyer, Kolesnikov, Weissenborn, Zhai, Unterthiner, Dehghani, Minderer, Heigold, Gelly, Uszkoreit, and Houlsby]{vit}
Alexey Dosovitskiy, Lucas Beyer, Alexander Kolesnikov, Dirk Weissenborn, Xiaohua Zhai, Thomas Unterthiner, Mostafa Dehghani, Matthias Minderer, Georg Heigold, Sylvain Gelly, Jakob Uszkoreit, and Neil Houlsby.
\newblock An image is worth 16x16 words: Transformers for image recognition at scale.
\newblock In \emph{ICLR}, 2020.

\bibitem[El-Nouby et~al.(2024)El-Nouby, Klein, Zhai, Bautista, Toshev, Shankar, Susskind, and Joulin]{aimv1}
Alaaeldin El-Nouby, Michal Klein, Shuangfei Zhai, Miguel~Angel Bautista, Alexander Toshev, Vaishaal Shankar, Joshua~M Susskind, and Armand Joulin.
\newblock Scalable pre-training of large autoregressive image models.
\newblock In \emph{ICML}, 2024.

\bibitem[Fan et~al.(2025)Fan, Tong, Zhu, Sinha, Liu, Chen, Rabbat, Ballas, LeCun, Bar, and Xie]{webdino}
David Fan, Shengbang Tong, Jiachen Zhu, Koustuv Sinha, Zhuang Liu, Xinlei Chen, Michael Rabbat, Nicolas Ballas, Yann LeCun, Amir Bar, and Saining Xie.
\newblock Scaling language-free visual representation learning.
\newblock \emph{arXiv:2504.01017}, 2025.

\bibitem[Fan et~al.(2023)Fan, Krishnan, Isola, Katabi, and Tian]{rewrite}
Lijie Fan, Dilip Krishnan, Phillip Isola, Dina Katabi, and Yonglong Tian.
\newblock Improving {CLIP} training with language rewrites.
\newblock In \emph{NeurIPS}, 2023.

\bibitem[Fang et~al.(2024{\natexlab{a}})Fang, Jose, Jain, Schmidt, Toshev, and Shankar]{dfn}
Alex Fang, Albin~Madappally Jose, Amit Jain, Ludwig Schmidt, Alexander Toshev, and Vaishaal Shankar.
\newblock Data filtering networks.
\newblock In \emph{ICLR}, 2024{\natexlab{a}}.

\bibitem[Fang et~al.(2023)Fang, Wang, Xie, Sun, Wu, Wang, Huang, Wang, and Cao]{eva}
Yuxin Fang, Wen Wang, Binhui Xie, Quan Sun, Ledell Wu, Xinggang Wang, Tiejun Huang, Xinlong Wang, and Yue Cao.
\newblock {EVA}: Exploring the limits of masked visual representation learning at scale.
\newblock In \emph{CVPR}, 2023.

\bibitem[Fang et~al.(2024{\natexlab{b}})Fang, Sun, Wang, Huang, Wang, and Cao]{eva2}
Yuxin Fang, Quan Sun, Xinggang Wang, Tiejun Huang, Xinlong Wang, and Yue Cao.
\newblock {EVA-02}: A visual representation for neon genesis.
\newblock \emph{Image and Vision Computing}, 2024{\natexlab{b}}.

\bibitem[Feichtenhofer(2020)]{feichtenhofer2020x3d}
Christoph Feichtenhofer.
\newblock {X3D}: Expanding architectures for efficient video recognition.
\newblock In \emph{CVPR}, 2020.

\bibitem[Fini et~al.(2025)Fini, Shukor, Li, Dufter, Klein, Haldimann, Aitharaju, da~Costa, B{\'{e}}thune, Gan, Toshev, Eichner, Nabi, Yang, Susskind, and El{-}Nouby]{aimv2}
Enrico Fini, Mustafa Shukor, Xiujun Li, Philipp Dufter, Michal Klein, David Haldimann, Sai Aitharaju, Victor Guilherme~Turrisi da Costa, Louis B{\'{e}}thune, Zhe Gan, Alexander~T. Toshev, Marcin Eichner, Moin Nabi, Yinfei Yang, Joshua~M. Susskind, and Alaaeldin El{-}Nouby.
\newblock Multimodal autoregressive pre-training of large vision encoders.
\newblock In \emph{CVPR}, 2025.

\bibitem[Fu et~al.(2024)Fu, Dai, Luo, Li, Ren, Zhang, Wang, Zhou, Shen, Zhang, Chen, Li, Lin, Zhao, Li, Xu, Zheng, Chen, Ji, and Sun]{fu2024videomme}
Chaoyou Fu, Yuhan Dai, Yongdong Luo, Lei Li, Shuhuai Ren, Renrui Zhang, Zihan Wang, Chenyu Zhou, Yunhang Shen, Mengdan Zhang, Peixian Chen, Yanwei Li, Shaohui Lin, Sirui Zhao, Ke Li, Tong Xu, Xiawu Zheng, Enhong Chen, Rongrong Ji, and Xing Sun.
\newblock {Video-MME}: The first-ever comprehensive evaluation benchmark of multi-modal llms in video analysis.
\newblock \emph{arXiv:2405.21075}, 2024.

\bibitem[Gadre et~al.(2023)Gadre, Ilharco, Fang, Hayase, Smyrnis, Nguyen, Marten, Wortsman, Ghosh, Zhang, Orgad, Entezari, Daras, Pratt, Ramanujan, Bitton, Marathe, Mussmann, Vencu, Cherti, Krishna, Koh, Saukh, Ratner, Song, Hajishirzi, Farhadi, Beaumont, Oh, Dimakis, Jitsev, Carmon, Shankar, and Schmidt]{datacomp}
Samir~Yitzhak Gadre, Gabriel Ilharco, Alex Fang, Jonathan Hayase, Georgios Smyrnis, Thao Nguyen, Ryan Marten, Mitchell Wortsman, Dhruba Ghosh, Jieyu Zhang, Eyal Orgad, Rahim Entezari, Giannis Daras, Sarah Pratt, Vivek Ramanujan, Yonatan Bitton, Kalyani Marathe, Stephen Mussmann, Richard Vencu, Mehdi Cherti, Ranjay Krishna, Pang~Wei Koh, Olga Saukh, Alexander Ratner, Shuran Song, Hannaneh Hajishirzi, Ali Farhadi, Romain Beaumont, Sewoong Oh, Alex Dimakis, Jenia Jitsev, Yair Carmon, Vaishaal Shankar, and Ludwig Schmidt.
\newblock {DataComp}: In search of the next generation of multimodal datasets.
\newblock In \emph{NeurIPS}, 2023.

\bibitem[Goyal et~al.(2017)Goyal, Khot, Summers-Stay, Batra, and Parikh]{goyal2017vqav2}
Yash Goyal, Tejas Khot, Douglas Summers-Stay, Dhruv Batra, and Devi Parikh.
\newblock Making the v in {VQA} matter: Elevating the role of image understanding in visual question answering.
\newblock In \emph{CVPR}, 2017.

\bibitem[Gupta et~al.(2019)Gupta, Dollar, and Girshick]{lvis}
Agrim Gupta, Piotr Dollar, and Ross Girshick.
\newblock {LVIS}: A dataset for large vocabulary instance segmentation.
\newblock In \emph{CVPR}, 2019.

\bibitem[He et~al.(2016)He, Zhang, Ren, and Sun]{resnet}
Kaiming He, Xiangyu Zhang, Shaoqing Ren, and Jian Sun.
\newblock Deep residual learning for image recognition.
\newblock In \emph{CVPR}, 2016.

\bibitem[He et~al.(2017)He, Gkioxari, Doll{\'a}r, and Girshick]{maskrcnn}
Kaiming He, Georgia Gkioxari, Piotr Doll{\'a}r, and Ross Girshick.
\newblock {Mask R-CNN}.
\newblock In \emph{ICCV}, 2017.

\bibitem[He et~al.(2022)He, Chen, Xie, Li, Doll{\'a}r, and Girshick]{mae}
Kaiming He, Xinlei Chen, Saining Xie, Yanghao Li, Piotr Doll{\'a}r, and Ross Girshick.
\newblock Masked autoencoders are scalable vision learners.
\newblock In \emph{CVPR}, 2022.

\bibitem[Heinrich et~al.(2025)Heinrich, Ranzinger, Hongxu, Yin, Lu, Kautz, Tao, Catanzaro, and Molchanov]{heinrich2024radio2.5}
Greg Heinrich, Mike Ranzinger, Hongxu, Yin, Yao Lu, Jan Kautz, Andrew Tao, Bryan Catanzaro, and Pavlo Molchanov.
\newblock {RADIOv2.5}: Improved baselines for agglomerative vision foundation models.
\newblock In \emph{CVPR}, 2025.

\bibitem[Hendrycks et~al.(2021{\natexlab{a}})Hendrycks, Basart, Mu, Kadavath, Wang, Dorundo, Desai, Zhu, Parajuli, Guo, Song, Steinhardt, and Gilmer]{imagenet-r}
Dan Hendrycks, Steven Basart, Norman Mu, Saurav Kadavath, Frank Wang, Evan Dorundo, Rahul Desai, Tyler Zhu, Samyak Parajuli, Mike Guo, Dawn Song, Jacob Steinhardt, and Justin Gilmer.
\newblock The many faces of robustness: A critical analysis of out-of-distribution generalization.
\newblock In \emph{ICCV}, 2021{\natexlab{a}}.

\bibitem[Hendrycks et~al.(2021{\natexlab{b}})Hendrycks, Zhao, Basart, Steinhardt, and Song]{imagenet-a}
Dan Hendrycks, Kevin Zhao, Steven Basart, Jacob Steinhardt, and Dawn Song.
\newblock Natural adversarial examples.
\newblock In \emph{CVPR}, 2021{\natexlab{b}}.

\bibitem[Heo et~al.(2024)Heo, Park, Han, and Yun]{heo2024rotary}
Byeongho Heo, Song Park, Dongyoon Han, and Sangdoo Yun.
\newblock Rotary position embedding for vision transformer.
\newblock In \emph{ECCV}, 2024.

\bibitem[Hinton et~al.(2015)Hinton, Vinyals, and Dean]{distillation}
Geoffrey Hinton, Oriol Vinyals, and Jeff Dean.
\newblock Distilling the knowledge in a neural network.
\newblock In \emph{NeurIPS Deep Learning Workshop}, 2015.

\bibitem[Huang et~al.(2016)Huang, Sun, Liu, Sedra, and Weinberger]{droppath}
Gao Huang, Yu Sun, Zhuang Liu, Daniel Sedra, and Kilian~Q Weinberger.
\newblock Deep networks with stochastic depth.
\newblock In \emph{ECCV}, 2016.

\bibitem[Ilharco et~al.(2021)Ilharco, Wortsman, Wightman, Gordon, Carlini, Taori, Dave, Shankar, Namkoong, Miller, Hajishirzi, Farhadi, and Schmidt]{openclip}
Gabriel Ilharco, Mitchell Wortsman, Ross Wightman, Cade Gordon, Nicholas Carlini, Rohan Taori, Achal Dave, Vaishaal Shankar, Hongseok Namkoong, John Miller, Hannaneh Hajishirzi, Ali Farhadi, and Ludwig Schmidt.
\newblock {OpenCLIP}, 2021.

\bibitem[Jabri et~al.(2020)Jabri, Owens, and Efros]{jabri2020space}
Allan Jabri, Andrew Owens, and Alexei Efros.
\newblock Space-time correspondence as a contrastive random walk.
\newblock In \emph{NeurIPS}, 2020.

\bibitem[Jang et~al.(2017)Jang, Song, Yu, Kim, and Kim]{jang2017tgif}
Yunseok Jang, Yale Song, Youngjae Yu, Youngjin Kim, and Gunhee Kim.
\newblock {TGIF-QA}: Toward spatio-temporal reasoning in visual question answering.
\newblock In \emph{CVPR}, 2017.

\bibitem[Jia et~al.(2021)Jia, Yang, Xia, Chen, Parekh, Pham, Le, Sung, Li, and Duerig]{align}
Chao Jia, Yinfei Yang, Ye Xia, Yi-Ting Chen, Zarana Parekh, Hieu Pham, Quoc Le, Yun-Hsuan Sung, Zhen Li, and Tom Duerig.
\newblock Scaling up visual and vision-language representation learning with noisy text supervision.
\newblock In \emph{ICML}, 2021.

\bibitem[Kay et~al.(2017)Kay, Carreira, Simonyan, Zhang, Hillier, Vijayanarasimhan, Viola, Green, Back, Natsev, Suleyman, and Zisserman]{kay2017kinetics}
Will Kay, Joao Carreira, Karen Simonyan, Brian Zhang, Chloe Hillier, Sudheendra Vijayanarasimhan, Fabio Viola, Tim Green, Trevor Back, Paul Natsev, Mustafa Suleyman, and Andrew Zisserman.
\newblock The kinetics human action video dataset.
\newblock \emph{arXiv:1705.06950}, 2017.

\bibitem[Kazemzadeh et~al.(2014)Kazemzadeh, Ordonez, Matten, and Berg]{kazemzadeh2014referitgame}
Sahar Kazemzadeh, Vicente Ordonez, Mark Matten, and Tamara Berg.
\newblock Referitgame: Referring to objects in photographs of natural scenes.
\newblock In \emph{EMNLP}, 2014.

\bibitem[Kembhavi et~al.(2016)Kembhavi, Salvato, Kolve, Seo, Hajishirzi, and Farhadi]{kembhavi2016ai2d}
Aniruddha Kembhavi, Mike Salvato, Eric Kolve, Minjoon Seo, Hannaneh Hajishirzi, and Ali Farhadi.
\newblock A diagram is worth a dozen images.
\newblock In \emph{ECCV}, 2016.

\bibitem[Kirillov et~al.(2023)Kirillov, Mintun, Ravi, Mao, Rolland, Gustafson, Xiao, Whitehead, Berg, Lo, Dollár, and Girshick]{sam}
Alexander Kirillov, Eric Mintun, Nikhila Ravi, Hanzi Mao, Chloe Rolland, Laura Gustafson, Tete Xiao, Spencer Whitehead, Alexander~C. Berg, Wan-Yen Lo, Piotr Dollár, and Ross Girshick.
\newblock Segment anything.
\newblock In \emph{ICCV}, 2023.

\bibitem[Krause et~al.(2013)Krause, Stark, Deng, and Fei-Fei]{cars}
Jonathan Krause, Michael Stark, Jia Deng, and Li Fei-Fei.
\newblock 3d object representations for fine-grained categorization.
\newblock In \emph{ICCV Workshop}, 2013.

\bibitem[Krishna et~al.(2017)Krishna, Zhu, Groth, Johnson, Hata, Kravitz, Chen, Kalantidis, Li, Shamma, Bernstein, and Li]{krishna2017visual}
Ranjay Krishna, Yuke Zhu, Oliver Groth, Justin Johnson, Kenji Hata, Joshua Kravitz, Stephanie Chen, Yannis Kalantidis, Li-Jia Li, David~A. Shamma, Michael~S. Bernstein, and Fei-Fei Li.
\newblock Visual genome: Connecting language and vision using crowdsourced dense image annotations.
\newblock \emph{IJCV}, 2017.

\bibitem[Krizhevsky et~al.(2012)Krizhevsky, Sutskever, and Hinton]{alexnet}
Alex Krizhevsky, Ilya Sutskever, and Geoffrey~E Hinton.
\newblock Imagenet classification with deep convolutional neural networks.
\newblock In \emph{NeurIPS}, 2012.

\bibitem[Kuehne et~al.(2011)Kuehne, Jhuang, Garrote, Poggio, and Serre]{kuehne2011hmdb}
Hildegard Kuehne, Hueihan Jhuang, Est{\'\i}baliz Garrote, Tomaso Poggio, and Thomas Serre.
\newblock {HMDB}: a large video database for human motion recognition.
\newblock In \emph{ICCV}, 2011.

\bibitem[Kuo et~al.(2023)Kuo, Cui, Gu, Piergiovanni, and Angelova]{fvlm}
Weicheng Kuo, Yin Cui, Xiuye Gu, A.~J. Piergiovanni, and Anelia Angelova.
\newblock {F-VLM:} open-vocabulary object detection upon frozen vision and language models.
\newblock In \emph{ICLR}, 2023.

\bibitem[Lai et~al.(2024)Lai, Zhang, Zhang, Wu, Bai, Timofeev, Du, Gan, Shan, Chuah, Yang, and Cao]{veclip}
Zhengfeng Lai, Haotian Zhang, Bowen Zhang, Wentao Wu, Haoping Bai, Aleksei Timofeev, Xianzhi Du, Zhe Gan, Jiulong Shan, Chen-Nee Chuah, Yinfei Yang, and Meng Cao.
\newblock {VeCLIP}: Improving {CLIP} training via visual-enriched captions.
\newblock In \emph{ECCV}, 2024.

\bibitem[Laurençon et~al.(2024)Laurençon, Tronchon, Cord, and Sanh]{laurençon2024matters}
Hugo Laurençon, Léo Tronchon, Matthieu Cord, and Victor Sanh.
\newblock What matters when building vision-language models?
\newblock In \emph{NeurIPS}, 2024.

\bibitem[Li et~al.(2025)Li, Zhang, Guo, Zhang, Li, Zhang, Zhang, Li, Liu, and Li]{llava-onevision}
Bo Li, Yuanhan Zhang, Dong Guo, Renrui Zhang, Feng Li, Hao Zhang, Kaichen Zhang, Yanwei Li, Ziwei Liu, and Chunyuan Li.
\newblock {LLaVA-OneVision}: Easy visual task transfer.
\newblock \emph{TMLR}, 2025.

\bibitem[Li et~al.(2023{\natexlab{a}})Li, Wang, Li, Wang, He, Wang, and Qiao]{umt}
Kunchang Li, Yali Wang, Yizhuo Li, Yi Wang, Yinan He, Limin Wang, and Yu Qiao.
\newblock Unmasked teacher: Towards training-efficient video foundation models.
\newblock In \emph{ICCV}, 2023{\natexlab{a}}.

\bibitem[Li et~al.(2024{\natexlab{a}})Li, Wang, He, Li, Wang, Liu, Wang, Xu, Chen, Luo, Wang, and Qiao]{li2024mvbench}
Kunchang Li, Yali Wang, Yinan He, Yizhuo Li, Yi Wang, Yi Liu, Zun Wang, Jilan Xu, Guo Chen, Ping Luo, Limin Wang, and Yu Qiao.
\newblock {MVBench}: A comprehensive multi-modal video understanding benchmark.
\newblock In \emph{CVPR}, 2024{\natexlab{a}}.

\bibitem[Li et~al.(2022{\natexlab{a}})Li, Zhang, Zhang, Yang, Li, Zhong, Wang, Yuan, Zhang, Hwang, Chang, and Gao]{glip}
Liunian~Harold Li, Pengchuan Zhang, Haotian Zhang, Jianwei Yang, Chunyuan Li, Yiwu Zhong, Lijuan Wang, Lu Yuan, Lei Zhang, Jenq-Neng Hwang, Kai-Wei Chang, and Jianfeng Gao.
\newblock Grounded language-image pre-training.
\newblock In \emph{CVPR}, 2022{\natexlab{a}}.

\bibitem[Li et~al.(2023{\natexlab{b}})Li, Wang, and Xie]{li2023clipa}
Xianhang Li, Zeyu Wang, and Cihang Xie.
\newblock An inverse scaling law for {CLIP} training.
\newblock In \emph{NeurIPS}, 2023{\natexlab{b}}.

\bibitem[Li et~al.(2023{\natexlab{c}})Li, Wang, and Xie]{li2023clipav2}
Xianhang Li, Zeyu Wang, and Cihang Xie.
\newblock {CLIPA}-v2: Scaling {CLIP} training with 81.1\% zero-shot imagenet accuracy within a {\$}10,000 budget; an extra {\$}4,000 unlocks 81.8{\%} accuracy.
\newblock \emph{arXiv:2306.15658}, 2023{\natexlab{c}}.

\bibitem[Li et~al.(2022{\natexlab{b}})Li, Mao, Girshick, and He]{vitdet}
Yanghao Li, Hanzi Mao, Ross Girshick, and Kaiming He.
\newblock Exploring plain vision transformer backbones for object detection.
\newblock In \emph{ECCV}, 2022{\natexlab{b}}.

\bibitem[Li et~al.(2023{\natexlab{d}})Li, Du, Zhou, Wang, Zhao, and Wen]{li2023popebenchmark}
Yifan Li, Yifan Du, Kun Zhou, Jinpeng Wang, Wayne~Xin Zhao, and Ji-Rong Wen.
\newblock Evaluating object hallucination in large vision-language models.
\newblock In \emph{EMNLP}, 2023{\natexlab{d}}.

\bibitem[Li et~al.(2023{\natexlab{e}})Li, Fan, Hu, Feichtenhofer, and He]{flip}
Yanghao Li, Haoqi Fan, Ronghang Hu, Christoph Feichtenhofer, and Kaiming He.
\newblock Scaling language-image pre-training via masking.
\newblock In \emph{CVPR}, 2023{\natexlab{e}}.

\bibitem[Li et~al.(2024{\natexlab{b}})Li, Wang, Liu, and Jiang]{li2024binsformer}
Zhenyu Li, Xuyang Wang, Xianming Liu, and Junjun Jiang.
\newblock Binsformer: Revisiting adaptive bins for monocular depth estimation.
\newblock \emph{TIP}, 2024{\natexlab{b}}.

\bibitem[Lin et~al.(2014)Lin, Maire, Belongie, Hays, Perona, Ramanan, Doll{\'a}r, and Zitnick]{coco}
Tsung-Yi Lin, Michael Maire, Serge Belongie, James Hays, Pietro Perona, Deva Ramanan, Piotr Doll{\'a}r, and C~Lawrence Zitnick.
\newblock Microsoft {COCO}: Common objects in context.
\newblock In \emph{ECCV}, 2014.

\bibitem[Liu et~al.(2024{\natexlab{a}})Liu, Li, Li, Li, Zhang, Shen, and Lee]{liu2024llavanext}
Haotian Liu, Chunyuan Li, Yuheng Li, Bo Li, Yuanhan Zhang, Sheng Shen, and Yong~Jae Lee.
\newblock {LLaVA-NeXT}: Improved reasoning, ocr, and world knowledge, 2024{\natexlab{a}}.

\bibitem[Liu et~al.(2024{\natexlab{b}})Liu, Li, Wu, and Lee]{llava}
Haotian Liu, Chunyuan Li, Qingyang Wu, and Yong~Jae Lee.
\newblock Visual instruction tuning.
\newblock \emph{NeurIPS}, 2024{\natexlab{b}}.

\bibitem[Liu et~al.(2021)Liu, Lin, Cao, Hu, Wei, Zhang, Lin, and Guo]{swin}
Ze Liu, Yutong Lin, Yue Cao, Han Hu, Yixuan Wei, Zheng Zhang, Stephen Lin, and Baining Guo.
\newblock Swin transformer: Hierarchical vision transformer using shifted windows.
\newblock In \emph{ICCV}, 2021.

\bibitem[Liu et~al.(2022{\natexlab{a}})Liu, Hu, Lin, Yao, Xie, Wei, Ning, Cao, Zhang, Dong, Wei, and Guo]{swin2}
Ze Liu, Han Hu, Yutong Lin, Zhuliang Yao, Zhenda Xie, Yixuan Wei, Jia Ning, Yue Cao, Zheng Zhang, Li Dong, Furu Wei, and Baining Guo.
\newblock Swin transformer v2: Scaling up capacity and resolution.
\newblock In \emph{CVPR}, 2022{\natexlab{a}}.

\bibitem[Liu et~al.(2022{\natexlab{b}})Liu, Mao, Wu, Feichtenhofer, Darrell, and Xie]{convnext}
Zhuang Liu, Hanzi Mao, Chao{-}Yuan Wu, Christoph Feichtenhofer, Trevor Darrell, and Saining Xie.
\newblock A {ConvNet} for the 2020s.
\newblock In \emph{CVPR}, 2022{\natexlab{b}}.

\bibitem[Llama~Team(2024)]{llama3}
AI~@~Meta Llama~Team.
\newblock The llama 3 herd of models.
\newblock \emph{arXiv:2407.21783}, 2024.

\bibitem[Loshchilov and Hutter(2019)]{adamw}
Ilya Loshchilov and Frank Hutter.
\newblock Decoupled weight decay regularization.
\newblock \emph{ICLR}, 2019.

\bibitem[Luo et~al.(2021)Luo, Ji, Zhong, Chen, Lei, Duan, and Li]{clip4clip}
Huaishao Luo, Lei Ji, Ming Zhong, Yang Chen, Wen Lei, Nan Duan, and Tianrui Li.
\newblock {CLIP4Clip}: An empirical study of clip for end to end video clip retrieval.
\newblock \emph{Neurocomputing}, 2021.

\bibitem[Ma et~al.(2024)Ma, Goldstein, Albergo, Boffi, Vanden-Eijnden, and Xie]{ma2024sit}
Nanye Ma, Mark Goldstein, Michael~S Albergo, Nicholas~M Boffi, Eric Vanden-Eijnden, and Saining Xie.
\newblock {SiT}: Exploring flow and diffusion-based generative models with scalable interpolant transformers.
\newblock In \emph{ECCV}, 2024.

\bibitem[Maaz et~al.(2024{\natexlab{a}})Maaz, Rasheed, Khan, and Khan]{Maaz2023VideoChatGPT}
Muhammad Maaz, Hanoona Rasheed, Salman Khan, and Fahad~Shahbaz Khan.
\newblock Video-{ChatGPT}: Towards detailed video understanding via large vision and language models.
\newblock In \emph{ACL}, 2024{\natexlab{a}}.

\bibitem[Maaz et~al.(2024{\natexlab{b}})Maaz, Rasheed, Khan, and Khan]{Maaz2024VideoGPT+}
Muhammad Maaz, Hanoona Rasheed, Salman Khan, and Fahad~Shahbaz Khan.
\newblock {VideoGPT+}: Integrating image and video encoders for enhanced video understanding.
\newblock \emph{arXiv:2406.09418}, 2024{\natexlab{b}}.

\bibitem[Maji et~al.(2013)Maji, Rahtu, Kannala, Blaschko, and Vedaldi]{aircraft}
Subhransu Maji, Esa Rahtu, Juho Kannala, Matthew Blaschko, and Andrea Vedaldi.
\newblock Fine-grained visual classification of aircraft.
\newblock \emph{arxiv:1306.5151}, 2013.

\bibitem[Mangalam et~al.(2024)Mangalam, Akshulakov, and Malik]{mangalam2024egoschema}
Karttikeya Mangalam, Raiymbek Akshulakov, and Jitendra Malik.
\newblock Egoschema: A diagnostic benchmark for very long-form video language understanding.
\newblock \emph{NeurIPS}, 2024.

\bibitem[Maninis et~al.(2024)Maninis, Chen, Ghosh, Karpur, Chen, Xia, Cao, Salz, Han, Dlabal, et~al.]{maninis2024tips}
Kevis-Kokitsi Maninis, Kaifeng Chen, Soham Ghosh, Arjun Karpur, Koert Chen, Ye Xia, Bingyi Cao, Daniel Salz, Guangxing Han, Jan Dlabal, et~al.
\newblock Tips: Text-image pretraining with spatial awareness.
\newblock \emph{arXiv:2410.16512}, 2024.

\bibitem[Mathew et~al.(2021)Mathew, Karatzas, and Jawahar]{mathew2021docvqa}
Minesh Mathew, Dimosthenis Karatzas, and CV Jawahar.
\newblock {DocVQA}: A dataset for vqa on document images.
\newblock In \emph{WACV}, 2021.

\bibitem[Mathew et~al.(2022)Mathew, Bagal, Tito, Karatzas, Valveny, and Jawahar]{mathew2022infographicvqa}
Minesh Mathew, Viraj Bagal, Rub{\`e}n Tito, Dimosthenis Karatzas, Ernest Valveny, and CV Jawahar.
\newblock Infographicvqa.
\newblock In \emph{WACV}, 2022.

\bibitem[McKinzie et~al.(2024)McKinzie, Gan, Fauconnier, Dodge, Zhang, Dufter, Shah, Du, Peng, Weers, Belyi, Zhang, Singh, Kang, Jain, Hè, Schwarzer, Gunter, Kong, Zhang, Wang, Wang, Du, Lei, Wiseman, Yin, Lee, Wang, Pang, Grasch, Toshev, and Yang]{mm1}
Brandon McKinzie, Zhe Gan, Jean-Philippe Fauconnier, Sam Dodge, Bowen Zhang, Philipp Dufter, Dhruti Shah, Xianzhi Du, Futang Peng, Floris Weers, Anton Belyi, Haotian Zhang, Karanjeet Singh, Doug Kang, Ankur Jain, Hongyu Hè, Max Schwarzer, Tom Gunter, Xiang Kong, Aonan Zhang, Jianyu Wang, Chong Wang, Nan Du, Tao Lei, Sam Wiseman, Guoli Yin, Mark Lee, Zirui Wang, Ruoming Pang, Peter Grasch, Alexander Toshev, and Yinfei Yang.
\newblock {MM1}: methods, analysis and insights from multimodal {LLM} pre-training.
\newblock In \emph{ECCV}, 2024.

\bibitem[Minderer et~al.(2022)Minderer, Gritsenko, Stone, Neumann, Weissenborn, Dosovitskiy, Mahendran, Arnab, Dehghani, Shen, Wang, Zhai, Kipf, and Houlsby]{owlv1}
Matthias Minderer, Alexey~A. Gritsenko, Austin Stone, Maxim Neumann, Dirk Weissenborn, Alexey Dosovitskiy, Aravindh Mahendran, Anurag Arnab, Mostafa Dehghani, Zhuoran Shen, Xiao Wang, Xiaohua Zhai, Thomas Kipf, and Neil Houlsby.
\newblock Simple open-vocabulary object detection with vision transformers.
\newblock In \emph{ECCV}, 2022.

\bibitem[Minderer et~al.(2023)Minderer, Gritsenko, and Houlsby]{owlv2}
Matthias Minderer, Alexey Gritsenko, and Neil Houlsby.
\newblock Scaling open-vocabulary object detection.
\newblock In \emph{NeurIPS}, 2023.

\bibitem[Nguyen et~al.(2023)Nguyen, Gadre, Ilharco, Oh, and Schmidt]{Nguyen2023recap}
Thao Nguyen, Samir~Yitzhak Gadre, Gabriel Ilharco, Sewoong Oh, and Ludwig Schmidt.
\newblock Improving multimodal datasets with image captioning.
\newblock In \emph{NeurIPS}, 2023.

\bibitem[Nilsback and Zisserman(2008)]{flower102}
Maria-Elena Nilsback and Andrew Zisserman.
\newblock Automated flower classification over a large number of classes.
\newblock In \emph{ICVGIP}, 2008.

\bibitem[Oquab et~al.(2024)Oquab, Darcet, Moutakanni, Vo, Szafraniec, Khalidov, Fernandez, Haziza, Massa, El{-}Nouby, Assran, Ballas, Galuba, Howes, Huang, Li, Misra, Rabbat, Sharma, Synnaeve, Xu, J{\'{e}}gou, Mairal, Labatut, Joulin, and Bojanowski]{dinov2}
Maxime Oquab, Timoth{\'{e}}e Darcet, Th{\'{e}}o Moutakanni, Huy~V. Vo, Marc Szafraniec, Vasil Khalidov, Pierre Fernandez, Daniel Haziza, Francisco Massa, Alaaeldin El{-}Nouby, Mido Assran, Nicolas Ballas, Wojciech Galuba, Russell Howes, Po{-}Yao Huang, Shang{-}Wen Li, Ishan Misra, Michael Rabbat, Vasu Sharma, Gabriel Synnaeve, Hu Xu, Herv{\'{e}} J{\'{e}}gou, Julien Mairal, Patrick Labatut, Armand Joulin, and Piotr Bojanowski.
\newblock {DINOv2}: Learning robust visual features without supervision.
\newblock \emph{TMLR}, 2024.

\bibitem[Ouyang-Zhang et~al.(2022)Ouyang-Zhang, Cho, Zhou, and Kr{\"a}henb{\"u}hl]{deta}
Jeffrey Ouyang-Zhang, Jang~Hyun Cho, Xingyi Zhou, and Philipp Kr{\"a}henb{\"u}hl.
\newblock {NMS}strikes back.
\newblock \emph{arXiv:2212.06137}, 2022.

\bibitem[Parkhi et~al.(2012)Parkhi, Vedaldi, Zisserman, and Jawahar]{pets}
Omkar~M. Parkhi, Andrea Vedaldi, Andrew Zisserman, and C.~V. Jawahar.
\newblock Cats and dogs.
\newblock In \emph{CVPR}, 2012.

\bibitem[Peng et~al.(2023)Peng, Wang, Dong, Hao, Huang, Ma, and Wei]{kosmos-2}
Zhiliang Peng, Wenhui Wang, Li Dong, Yaru Hao, Shaohan Huang, Shuming Ma, and Furu Wei.
\newblock Kosmos-2: Grounding multimodal large language models to the world.
\newblock \emph{arXiv:2306.14824}, 2023.

\bibitem[Pham et~al.(2023)Pham, Dai, Ghiasi, Kawaguchi, Liu, Yu, Yu, Chen, Luong, Wu, Tan, and Le]{basic}
Hieu Pham, Zihang Dai, Golnaz Ghiasi, Kenji Kawaguchi, Hanxiao Liu, Adams~Wei Yu, Jiahui Yu, Yi-Ting Chen, Minh-Thang Luong, Yonghui Wu, Mingxing Tan, and Quoc~V. Le.
\newblock Combined scaling for zero-shot transfer learning.
\newblock \emph{Neurocomputing}, 2023.

\bibitem[Plummer et~al.(2015)Plummer, Wang, Cervantes, Caicedo, Hockenmaier, and Lazebnik]{plummer2015flickr30k}
Bryan~A Plummer, Liwei Wang, Chris~M Cervantes, Juan~C Caicedo, Julia Hockenmaier, and Svetlana Lazebnik.
\newblock Flickr30k entities: Collecting region-to-phrase correspondences for richer image-to-sentence models.
\newblock In \emph{ICCV}, 2015.

\bibitem[Pont-Tuset et~al.(2017)Pont-Tuset, Perazzi, Caelles, Arbel{\'a}ez, Sorkine-Hornung, and Van~Gool]{davis2017}
Jordi Pont-Tuset, Federico Perazzi, Sergi Caelles, Pablo Arbel{\'a}ez, Alex Sorkine-Hornung, and Luc Van~Gool.
\newblock The 2017 {DAVIS} challenge on video object segmentation.
\newblock \emph{arXiv:1704.00675}, 2017.

\bibitem[Pătrăucean et~al.(2024)Pătrăucean, Smaira, Gupta, Continente, Markeeva, Banarse, Koppula, Heyward, Malinowski, Yang, Doersch, Matejovicova, Sulsky, Miech, Frechette, Klimczak, Koster, Zhang, Winkler, Aytar, Osindero, Damen, Zisserman, and Carreira]{patraucean2024perceptiontest}
Viorica Pătrăucean, Lucas Smaira, Ankush Gupta, Adrià~Recasens Continente, Larisa Markeeva, Dylan Banarse, Skanda Koppula, Joseph Heyward, Mateusz Malinowski, Yi Yang, Carl Doersch, Tatiana Matejovicova, Yury Sulsky, Antoine Miech, Alex Frechette, Hanna Klimczak, Raphael Koster, Junlin Zhang, Stephanie Winkler, Yusuf Aytar, Simon Osindero, Dima Damen, Andrew Zisserman, and João Carreira.
\newblock Perception test: A diagnostic benchmark for multimodal video models.
\newblock In \emph{NeurIPS}, 2024.

\bibitem[Radford et~al.(2021)Radford, Kim, Hallacy, Ramesh, Goh, Agarwal, Sastry, Askell, Mishkin, Clark, Krueger, and Sutskever]{clip}
Alec Radford, Jong~Wook Kim, Chris Hallacy, Aditya Ramesh, Gabriel Goh, Sandhini Agarwal, Girish Sastry, Amanda Askell, Pamela Mishkin, Jack Clark, Gretchen Krueger, and Ilya Sutskever.
\newblock Learning transferable visual models from natural language supervision.
\newblock In \emph{ICML}, 2021.

\bibitem[Rajasegaran et~al.(2025)Rajasegaran, Radosavovic, Ravishankar, Gandelsman, Feichtenhofer, and Malik]{vgpt}
Jathushan Rajasegaran, Ilija Radosavovic, Rahul Ravishankar, Yossi Gandelsman, Christoph Feichtenhofer, and Jitendra Malik.
\newblock An empirical study of autoregressive pre-training from videos.
\newblock \emph{arXiv:2501.05453}, 2025.

\bibitem[Ramesh et~al.(2022)Ramesh, Dhariwal, Nichol, Chu, and Chen]{dalle2}
Aditya Ramesh, Prafulla Dhariwal, Alex Nichol, Casey Chu, and Mark Chen.
\newblock Hierarchical text-conditional image generation with {CLIP} latents.
\newblock \emph{arXiv:2204.06125}, 2022.

\bibitem[Ranftl et~al.(2021)Ranftl, Bochkovskiy, and Koltun]{dpt}
Ren{\'e} Ranftl, Alexey Bochkovskiy, and Vladlen Koltun.
\newblock Vision transformers for dense prediction.
\newblock In \emph{ICCV}, 2021.

\bibitem[Ranzinger et~al.(2024)Ranzinger, Heinrich, Kautz, and Molchanov]{ranzinger2023radio}
Mike Ranzinger, Greg Heinrich, Jan Kautz, and Pavlo Molchanov.
\newblock {AM-RADIO}: Agglomerative vision foundation model--reduce all domains into one.
\newblock In \emph{CVPR}, 2024.

\bibitem[Ravi et~al.(2024)Ravi, Gabeur, Hu, Hu, Ryali, Ma, Khedr, R{\"a}dle, Rolland, Gustafson, Mintun, Pan, Alwala, Carion, Wu, Girshick, Doll{\'a}r, and Feichtenhofer]{sam2}
Nikhila Ravi, Valentin Gabeur, Yuan-Ting Hu, Ronghang Hu, Chaitanya Ryali, Tengyu Ma, Haitham Khedr, Roman R{\"a}dle, Chloe Rolland, Laura Gustafson, Eric Mintun, Junting Pan, Kalyan~Vasudev Alwala, Nicolas Carion, Chao-Yuan Wu, Ross Girshick, Piotr Doll{\'a}r, and Christoph Feichtenhofer.
\newblock {SAM 2}: Segment anything in images and videos.
\newblock In \emph{ICLR}, 2024.

\bibitem[Recht et~al.(2019)Recht, Roelofs, Schmidt, and Shankar]{imagenetv2}
Benjamin Recht, Rebecca Roelofs, Ludwig Schmidt, and Vaishaal Shankar.
\newblock Do imagenet classifiers generalize to imagenet?
\newblock In \emph{ICML}, 2019.

\bibitem[Rojas et~al.(2022)Rojas, Diamos, Kini, Kanter, Reddi, and Coleman]{dollar_st}
William A.~Gaviria Rojas, Sudnya Diamos, Keertan~Ranjan Kini, David Kanter, Vijay~Janapa Reddi, and Cody Coleman.
\newblock The dollar street dataset: images representing the geographic and socioeconomic diversity of the world.
\newblock In \emph{NeurIPS Datasets and Benchmarks}, 2022.

\bibitem[Rombach et~al.(2022)Rombach, Blattmann, Lorenz, Esser, and Ommer]{ldm}
Robin Rombach, Andreas Blattmann, Dominik Lorenz, Patrick Esser, and Bj{\"o}rn Ommer.
\newblock High-resolution image synthesis with latent diffusion models.
\newblock In \emph{CVPR}, 2022.

\bibitem[Sariyildiz et~al.(2020)Sariyildiz, Perez, and Larlus]{sariyildiz2020icmlm}
Mert~Bulent Sariyildiz, Julien Perez, and Diane Larlus.
\newblock Learning visual representations with caption annotations.
\newblock In \emph{ECCV}, 2020.

\bibitem[Sariyildiz et~al.(2024)Sariyildiz, Weinzaepfel, Lucas, Larlus, and Kalantidis]{sariyildiz2024unic}
Mert~Bulent Sariyildiz, Philippe Weinzaepfel, Thomas Lucas, Diane Larlus, and Yannis Kalantidis.
\newblock {UNIC}: Universal classification models via multi-teacher distillation.
\newblock In \emph{ECCV}, 2024.

\bibitem[Schuhmann et~al.(2022)Schuhmann, Beaumont, Vencu, Gordon, Wightman, Cherti, Coombes, Katta, Mullis, Wortsman, Schramowski, Kundurthy, Crowson, Schmidt, Kaczmarczyk, and Jitsev]{laion}
Christoph Schuhmann, Romain Beaumont, Richard Vencu, Cade~W Gordon, Ross Wightman, Mehdi Cherti, Theo Coombes, Aarush Katta, Clayton Mullis, Mitchell Wortsman, Patrick Schramowski, Srivatsa~R Kundurthy, Katherine Crowson, Ludwig Schmidt, Robert Kaczmarczyk, and Jenia Jitsev.
\newblock {LAION}-5b: An open large-scale dataset for training next generation image-text models.
\newblock In \emph{NeurIPS Datasets and Benchmarks}, 2022.

\bibitem[Schwenk et~al.(2022)Schwenk, Khandelwal, Clark, Marino, and Mottaghi]{schwenk2022okvqa}
Dustin Schwenk, Apoorv Khandelwal, Christopher Clark, Kenneth Marino, and Roozbeh Mottaghi.
\newblock {A-OKVQA}: A benchmark for visual question answering using world knowledge.
\newblock In \emph{ECCV}, 2022.

\bibitem[Shang et~al.(2024)Shang, Schmeckpeper, May, Minniti, Kelestemur, Watkins, and Herlant]{shang2024theia}
Jinghuan Shang, Karl Schmeckpeper, Brandon~B May, Maria~Vittoria Minniti, Tarik Kelestemur, David Watkins, and Laura Herlant.
\newblock Theia: Distilling diverse vision foundation models for robot learning.
\newblock In \emph{CoRL}, 2024.

\bibitem[Shao et~al.(2019)Shao, Li, Zhang, Peng, Yu, Zhang, Li, and Sun]{o365}
Shuai Shao, Zeming Li, Tianyuan Zhang, Chao Peng, Gang Yu, Xiangyu Zhang, Jing Li, and Jian Sun.
\newblock Objects365: A large-scale, high-quality dataset for object detection.
\newblock In \emph{ICCV}, 2019.

\bibitem[Shekhar et~al.(2023)Shekhar, Bordes, Vincent, and Morcos]{shekhar2023objectives}
Shashank Shekhar, Florian Bordes, Pascal Vincent, and Ari Morcos.
\newblock Objectives matter: Understanding the impact of self-supervised objectives on vision transformer representations.
\newblock \emph{arXiv:2304.13089}, 2023.

\bibitem[Sidorov et~al.(2020)Sidorov, Hu, Rohrbach, and Singh]{sidorov2020textcaps}
Oleksii Sidorov, Ronghang Hu, Marcus Rohrbach, and Amanpreet Singh.
\newblock Textcaps: a dataset for image captioning with reading comprehension.
\newblock In \emph{ECCV}, 2020.

\bibitem[Silberman et~al.(2012)Silberman, Hoiem, Kohli, and Fergus]{nyu_depth}
Nathan Silberman, Derek Hoiem, Pushmeet Kohli, and Rob Fergus.
\newblock Indoor segmentation and support inference from rgbd images.
\newblock In \emph{ECCV}, 2012.

\bibitem[Simonyan and Zisserman(2015)]{vgg}
Karen Simonyan and Andrew Zisserman.
\newblock Very deep convolutional networks for large-scale image recognition.
\newblock In \emph{ICLR}, 2015.

\bibitem[Singh et~al.(2019)Singh, Natarjan, Shah, Jiang, Chen, Parikh, and Rohrbach]{singh2019textvqa}
Amanpreet Singh, Vivek Natarjan, Meet Shah, Yu Jiang, Xinlei Chen, Devi Parikh, and Marcus Rohrbach.
\newblock Towards {VQA} models that can read.
\newblock In \emph{CVPR}, 2019.

\bibitem[Soomro et~al.(2012)Soomro, Zamir, and Shah]{soomro2012ucf101}
Khurram Soomro, Amir~Roshan Zamir, and Mubarak Shah.
\newblock {UCF101}: A dataset of 101 human actions classes from videos in the wild.
\newblock \emph{arXiv:1212.0402}, 2012.

\bibitem[Su et~al.(2024)Su, Lu, Pan, Wen, and Liu]{rope}
Jianlin Su, Yu Lu, Shengfeng Pan, Bo Wen, and Yunfeng Liu.
\newblock {RoFormer}: Enhanced transformer with rotary position embedding.
\newblock \emph{Neurocomputing}, 2024.

\bibitem[Sun et~al.(2024{\natexlab{a}})Sun, Cao, Xie, Jiang, and Pang]{sun2024cliper}
Lin Sun, Jiale Cao, Jin Xie, Xiaoheng Jiang, and Yanwei Pang.
\newblock {CLIPer}: Hierarchically improving spatial representation of {CLIP} for open-vocabulary semantic segmentation.
\newblock \emph{arXiv:2411.13836}, 2024{\natexlab{a}}.

\bibitem[Sun et~al.(2023)Sun, Fang, Wu, Wang, and Cao]{EVA-CLIP}
Quan Sun, Yuxin Fang, Ledell Wu, Xinlong Wang, and Yue Cao.
\newblock {EVA-CLIP}: Improved training techniques for clip at scale.
\newblock \emph{arXiv:2303.15389}, 2023.

\bibitem[Sun et~al.(2024{\natexlab{b}})Sun, Wang, Yu, Cui, Zhang, Zhang, and Wang]{eva18b}
Quan Sun, Jinsheng Wang, Qiying Yu, Yufeng Cui, Fan Zhang, Xiaosong Zhang, and Xinlong Wang.
\newblock {EVA-CLIP-18B}: Scaling clip to 18 billion parameters.
\newblock \emph{arXiv:2402.04252}, 2024{\natexlab{b}}.

\bibitem[Tan and Le(2019)]{efficientnet}
Mingxing Tan and Quoc Le.
\newblock {EfficientNet}: Rethinking model scaling for convolutional neural networks.
\newblock In \emph{ICML}, 2019.

\bibitem[Team(2025)]{gemma3}
Gemma Team.
\newblock Gemma 3 technical report.
\newblock \emph{arXiv:2503.19786}, 2025.

\bibitem[Thomee et~al.(2016)Thomee, Shamma, Friedland, Elizalde, Ni, Poland, Borth, and Li]{thomee2016yfcc100m}
Bart Thomee, David~A. Shamma, Gerald Friedland, Benjamin Elizalde, Karl Ni, Douglas Poland, Damian Borth, and Li-Jia Li.
\newblock {YFCC100M}: The new data in multimedia research.
\newblock \emph{Communications of the {ACM}}, 2016.

\bibitem[Tong et~al.(2024)Tong, Brown, Wu, Woo, Middepogu, Akula, Yang, Yang, Iyer, Pan, Wang, Fergus, LeCun, and Xie]{cambrian}
Shengbang Tong, Ellis Brown, Penghao Wu, Sanghyun Woo, Manoj Middepogu, Sai~Charitha Akula, Jihan Yang, Shusheng Yang, Adithya Iyer, Xichen Pan, Ziteng Wang, Rob Fergus, Yann LeCun, and Saining Xie.
\newblock Cambrian-1: A fully open, vision-centric exploration of multimodal llms.
\newblock In \emph{NeurIPS}, 2024.

\bibitem[Touvron et~al.(2021)Touvron, Cord, Sablayrolles, Synnaeve, and J{\'e}gou]{layerscale}
Hugo Touvron, Matthieu Cord, Alexandre Sablayrolles, Gabriel Synnaeve, and Herv{\'e} J{\'e}gou.
\newblock Going deeper with image transformers.
\newblock In \emph{ICCV}, 2021.

\bibitem[Touvron et~al.(2022)Touvron, Cord, and J{\'e}gou]{touvron2022deit}
Hugo Touvron, Matthieu Cord, and Herv{\'e} J{\'e}gou.
\newblock {DeiT} {III}: Revenge of the {ViT}.
\newblock In \emph{ECCV}, 2022.

\bibitem[Tschannen et~al.(2023)Tschannen, Kumar, Steiner, Zhai, Houlsby, and Beyer]{cappa}
Michael Tschannen, Manoj Kumar, Andreas Steiner, Xiaohua Zhai, Neil Houlsby, and Lucas Beyer.
\newblock Image captioners are scalable vision learners too.
\newblock In \emph{NeurIPS}, 2023.

\bibitem[Tschannen et~al.(2025)Tschannen, Gritsenko, Wang, Naeem, Alabdulmohsin, Parthasarathy, Evans, Beyer, Xia, Mustafa, H\'enaff, Harmsen, Steiner, and Zhai]{siglip2}
Michael Tschannen, Alexey Gritsenko, Xiao Wang, Muhammad~Ferjad Naeem, Ibrahim Alabdulmohsin, Nikhil Parthasarathy, Talfan Evans, Lucas Beyer, Ye Xia, Basil Mustafa, Olivier H\'enaff, Jeremiah Harmsen, Andreas Steiner, and Xiaohua Zhai.
\newblock {SigLIP} 2: Multilingual vision-language encoders with improved semantic understanding, localization, and dense features.
\newblock \emph{arXiv:2502.14786}, 2025.

\bibitem[Urbanek et~al.(2024)Urbanek, Bordes, Astolfi, Williamson, Sharma, and Romero-Soriano]{Urbanek_2024_CVPR}
Jack Urbanek, Florian Bordes, Pietro Astolfi, Mary Williamson, Vasu Sharma, and Adriana Romero-Soriano.
\newblock A picture is worth more than 77 text tokens: Evaluating {CLIP}-style models on dense captions.
\newblock In \emph{CVPR}, 2024.

\bibitem[Van~Horn et~al.(2018)Van~Horn, Mac~Aodha, Song, Cui, Sun, Shepard, Adam, Perona, and Belongie]{inat2017}
Grant Van~Horn, Oisin Mac~Aodha, Yang Song, Yin Cui, Chen Sun, Alex Shepard, Hartwig Adam, Pietro Perona, and Serge Belongie.
\newblock The inaturalist species classification and detection dataset.
\newblock In \emph{CVPR}, 2018.

\bibitem[Vaswani et~al.(2017)Vaswani, Shazeer, Parmar, Uszkoreit, Jones, Gomez, Kaiser, and Polosukhin]{transformer}
Ashish Vaswani, Noam Shazeer, Niki Parmar, Jakob Uszkoreit, Llion Jones, Aidan~N. Gomez, Lukasz Kaiser, and Illia Polosukhin.
\newblock Attention is all you need.
\newblock In \emph{NeurIPS}, 2017.

\bibitem[Walmer et~al.(2023)Walmer, Suri, Gupta, and Shrivastava]{walmer2023teaching}
Matthew Walmer, Saksham Suri, Kamal Gupta, and Abhinav Shrivastava.
\newblock Teaching matters: Investigating the role of supervision in vision transformers.
\newblock In \emph{CVPR}, 2023.

\bibitem[Wang et~al.(2019)Wang, Ge, Lipton, and Xing]{imagenet-sketch}
Haohan Wang, Songwei Ge, Zachary Lipton, and Eric~P Xing.
\newblock Learning robust global representations by penalizing local predictive power.
\newblock In \emph{NeurIPS}, 2019.

\bibitem[Wang et~al.(2024{\natexlab{a}})Wang, Bai, Tan, Wang, Fan, Bai, Chen, Liu, Wang, Ge, Fan, Dang, Du, Ren, Men, Liu, Zhou, Zhou, and Lin]{qwen2vl}
Peng Wang, Shuai Bai, Sinan Tan, Shijie Wang, Zhihao Fan, Jinze Bai, Keqin Chen, Xuejing Liu, Jialin Wang, Wenbin Ge, Yang Fan, Kai Dang, Mengfei Du, Xuancheng Ren, Rui Men, Dayiheng Liu, Chang Zhou, Jingren Zhou, and Junyang Lin.
\newblock Qwen2-{VL}: Enhancing vision-language model's perception of the world at any resolution.
\newblock \emph{arXiv:2409.12191}, 2024{\natexlab{a}}.

\bibitem[Wang et~al.(2023)Wang, Dai, Chen, Huang, Li, Zhu, Hu, Lu, Lu, Li, Wang, and Qiao]{internimage}
Wenhai Wang, Jifeng Dai, Zhe Chen, Zhenhang Huang, Zhiqi Li, Xizhou Zhu, Xiaowei Hu, Tong Lu, Lewei Lu, Hongsheng Li, Xiaogang Wang, and Yu Qiao.
\newblock {InternImage}: Exploring large-scale vision foundation models with deformable convolutions.
\newblock In \emph{CVPR}, 2023.

\bibitem[Wang et~al.(2024{\natexlab{b}})Wang, Li, Li, Yu, He, Chen, Pei, Zheng, Wang, Shi, Jiang, Li, Xu, Zhang, Huang, Qiao, Wang, and Wang]{internvideo2}
Yi Wang, Kunchang Li, Xinhao Li, Jiashuo Yu, Yinan He, Guo Chen, Baoqi Pei, Rongkun Zheng, Zun Wang, Yansong Shi, Tianxiang Jiang, Songze Li, Jilan Xu, Hongjie Zhang, Yifei Huang, Yu Qiao, Yali Wang, and Limin Wang.
\newblock {InternVideo2}: Scaling foundation models for multimodal video understanding.
\newblock In \emph{ECCV}, 2024{\natexlab{b}}.

\bibitem[Wei et~al.(2022)Wei, Fan, Xie, Wu, Yuille, and Feichtenhofer]{maskfeat}
Chen Wei, Haoqi Fan, Saining Xie, Chao{-}Yuan Wu, Alan~L. Yuille, and Christoph Feichtenhofer.
\newblock Masked feature prediction for self-supervised visual pre-training.
\newblock In \emph{CVPR}, 2022.

\bibitem[Wu et~al.(2021)Wu, Yu, Chen, Tenenbaum, and Gan]{wu2021star}
Bo Wu, Shoubin Yu, Zhenfang Chen, Joshua~B Tenenbaum, and Chuang Gan.
\newblock {STAR}: A benchmark for situated reasoning in real-world videos.
\newblock In \emph{NeurIPS}, 2021.

\bibitem[Wu et~al.(2019)Wu, Kirillov, Massa, Lo, and Girshick]{d2}
Yuxin Wu, Alexander Kirillov, Francisco Massa, Wan-Yen Lo, and Ross Girshick.
\newblock Detectron2, 2019.

\bibitem[Xiao et~al.(2014)Xiao, Ehinger, Hays, Torralba, and Oliva]{sun397}
Jianxiong Xiao, Krista~A. Ehinger, James Hays, Antonio Torralba, and Aude Oliva.
\newblock {SUN} database: Exploring a large collection of scene categories.
\newblock \emph{IJCV}, 2014.

\bibitem[Xu et~al.(2024{\natexlab{a}})Xu, Huang, Tan, Yeh, Kahn, Jou, Ghosh, Levy, Zettlemoyer, tau Yih, Li, Xie, and Feichtenhofer]{altogether}
Hu Xu, Po-Yao Huang, Xiaoqing~Ellen Tan, Ching-Feng Yeh, Jacob Kahn, Christine Jou, Gargi Ghosh, Omer Levy, Luke Zettlemoyer, Wen tau Yih, Shang-Wen Li, Saining Xie, and Christoph Feichtenhofer.
\newblock Altogether: Image captioning via re-aligning alt-text.
\newblock In \emph{EMNLP}, 2024{\natexlab{a}}.

\bibitem[Xu et~al.(2024{\natexlab{b}})Xu, Xie, Tan, Huang, Howes, Sharma, Li, Ghosh, Zettlemoyer, and Feichtenhofer]{metaclip}
Hu Xu, Saining Xie, Xiaoqing~Ellen Tan, Po-Yao Huang, Russell Howes, Vasu Sharma, Shang-Wen Li, Gargi Ghosh, Luke Zettlemoyer, and Christoph Feichtenhofer.
\newblock Demystifying clip data.
\newblock In \emph{ICLR}, 2024{\natexlab{b}}.

\bibitem[Xu et~al.(2016)Xu, Mei, Yao, and Rui]{vtt}
Jun Xu, Tao Mei, Ting Yao, and Yong Rui.
\newblock {MSR-VTT}: A large video description dataset for bridging video and language.
\newblock In \emph{CVPR}, 2016.

\bibitem[Yang et~al.(2024{\natexlab{a}})Yang, Yang, Hui, Zheng, Yu, Zhou, Li, Li, Liu, Huang, Dong, Wei, Lin, Tang, Wang, Yang, Tu, Zhang, Ma, Xu, Zhou, Bai, He, Lin, Dang, Lu, Chen, Yang, Li, Xue, Ni, Zhang, Wang, Peng, Men, Gao, Lin, Wang, Bai, Tan, Zhu, Li, Liu, Ge, Deng, Zhou, Ren, Zhang, Wei, Ren, Fan, Yao, Zhang, Wan, Chu, Liu, Cui, Zhang, and Fan]{qwen2}
An Yang, Baosong Yang, Binyuan Hui, Bo Zheng, Bowen Yu, Chang Zhou, Chengpeng Li, Chengyuan Li, Dayiheng Liu, Fei Huang, Guanting Dong, Haoran Wei, Huan Lin, Jialong Tang, Jialin Wang, Jian Yang, Jianhong Tu, Jianwei Zhang, Jianxin Ma, Jin Xu, Jingren Zhou, Jinze Bai, Jinzheng He, Junyang Lin, Kai Dang, Keming Lu, Keqin Chen, Kexin Yang, Mei Li, Mingfeng Xue, Na Ni, Pei Zhang, Peng Wang, Ru Peng, Rui Men, Ruize Gao, Runji Lin, Shijie Wang, Shuai Bai, Sinan Tan, Tianhang Zhu, Tianhao Li, Tianyu Liu, Wenbin Ge, Xiaodong Deng, Xiaohuan Zhou, Xingzhang Ren, Xinyu Zhang, Xipin Wei, Xuancheng Ren, Yang Fan, Yang Yao, Yichang Zhang, Yu Wan, Yunfei Chu, Yuqiong Liu, Zeyu Cui, Zhenru Zhang, and Zhihao Fan.
\newblock Qwen2 technical report.
\newblock \emph{arxiv:2407.10671}, 2024{\natexlab{a}}.

\bibitem[Yang et~al.(2024{\natexlab{b}})Yang, Yang, Zhang, Hui, Zheng, Yu, Li, Liu, Huang, Wei, Lin, Yang, Tu, Zhang, Yang, Yang, Zhou, Lin, Dang, Lu, Bao, Yang, Yu, Li, Xue, Zhang, Zhu, Men, Lin, Li, Xia, Ren, Ren, Fan, Su, Zhang, Wan, Liu, Cui, Zhang, and Qiu]{qwen2.5}
An Yang, Baosong Yang, Beichen Zhang, Binyuan Hui, Bo Zheng, Bowen Yu, Chengyuan Li, Dayiheng Liu, Fei Huang, Haoran Wei, Huan Lin, Jian Yang, Jianhong Tu, Jianwei Zhang, Jianxin Yang, Jiaxi Yang, Jingren Zhou, Junyang Lin, Kai Dang, Keming Lu, Keqin Bao, Kexin Yang, Le Yu, Mei Li, Mingfeng Xue, Pei Zhang, Qin Zhu, Rui Men, Runji Lin, Tianhao Li, Tingyu Xia, Xingzhang Ren, Xuancheng Ren, Yang Fan, Yang Su, Yichang Zhang, Yu Wan, Yuqiong Liu, Zeyu Cui, Zhenru Zhang, and Zihan Qiu.
\newblock Qwen2.5 technical report.
\newblock \emph{arXiv:2412.15115}, 2024{\natexlab{b}}.

\bibitem[You et~al.(2020)You, Li, Reddi, Hseu, Kumar, Bhojanapalli, Song, Demmel, Keutzer, and Hsieh]{lamb}
Yang You, Jing Li, Sashank~J. Reddi, Jonathan Hseu, Sanjiv Kumar, Srinadh Bhojanapalli, Xiaodan Song, James Demmel, Kurt Keutzer, and Cho{-}Jui Hsieh.
\newblock Large batch optimization for deep learning: Training {BERT} in 76 minutes.
\newblock In \emph{ICLR}, 2020.

\bibitem[Young et~al.(2014)Young, Lai, Hodosh, and Hockenmaier]{flickr}
Peter Young, Alice Lai, Micah Hodosh, and Julia Hockenmaier.
\newblock From image descriptions to visual denotations: New similarity metrics for semantic inference over event descriptions.
\newblock \emph{TACL}, 2014.

\bibitem[Yu et~al.(2022)Yu, Wang, Vasudevan, Yeung, Seyedhosseini, and Wu]{coca}
Jiahui Yu, Zirui Wang, Vijay Vasudevan, Legg Yeung, Mojtaba Seyedhosseini, and Yonghui Wu.
\newblock {CoCa}: Contrastive captioners are image-text foundation models.
\newblock \emph{TMLR}, 2022.

\bibitem[Yu et~al.(2025)Yu, Kwak, Jang, Jeong, Huang, Shin, and Xie]{repa}
Sihyun Yu, Sangkyung Kwak, Huiwon Jang, Jongheon Jeong, Jonathan Huang, Jinwoo Shin, and Saining Xie.
\newblock Representation alignment for generation: Training diffusion transformers is easier than you think.
\newblock In \emph{ICLR}, 2025.

\bibitem[Zhai et~al.(2023)Zhai, Mustafa, Kolesnikov, and Beyer]{siglip}
Xiaohua Zhai, Basil Mustafa, Alexander Kolesnikov, and Lucas Beyer.
\newblock Sigmoid loss for language image pre-training.
\newblock In \emph{ICCV}, 2023.

\bibitem[Zhang et~al.(2023)Zhang, Li, Liu, Zhang, Su, Zhu, Ni, and Shum]{dino_det}
Hao Zhang, Feng Li, Shilong Liu, Lei Zhang, Hang Su, Jun Zhu, Lionel~M Ni, and Heung-Yeung Shum.
\newblock {DINO}: {DETR} with improved denoising anchor boxes for end-to-end object detection.
\newblock In \emph{ICLR}, 2023.

\bibitem[Zhang et~al.(2016)Zhang, Isola, and Efros]{zhang2016colorful}
Richard Zhang, Phillip Isola, and Alexei~A Efros.
\newblock Colorful image colorization.
\newblock In \emph{ECCV}, 2016.

\bibitem[Zhang et~al.(2022)Zhang, Jiang, Miura, Manning, and Langlotz]{pmlr-v182-zhang22a}
Yuhao Zhang, Hang Jiang, Yasuhide Miura, Christopher~D. Manning, and Curtis~P. Langlotz.
\newblock Contrastive learning of medical visual representations from paired images and text.
\newblock In \emph{MLHC}, 2022.

\bibitem[Zhao et~al.(2024)Zhao, Gundavarapu, Yuan, Zhou, Yan, Sun, Friedman, Qian, Weyand, Zhao, Hornung, Schroff, Yang, Ross, Wang, Adam, Sirotenko, Liu, and Gong]{videoprism}
Long Zhao, Nitesh~Bharadwaj Gundavarapu, Liangzhe Yuan, Hao Zhou, Shen Yan, Jennifer~J. Sun, Luke Friedman, Rui Qian, Tobias Weyand, Yue Zhao, Rachel Hornung, Florian Schroff, Ming Yang, David~A. Ross, Huisheng Wang, Hartwig Adam, Mikhail Sirotenko, Ting Liu, and Boqing Gong.
\newblock {VideoPrism}: A foundational visual encoder for video understanding.
\newblock In \emph{ICML}, 2024.

\bibitem[Zheng et~al.(2025)Zheng, Wang, Thomas, and Huang]{zheng2024chartqa}
Hanwen Zheng, Sijia Wang, Chris Thomas, and Lifu Huang.
\newblock Advancing chart question answering with robust chart component recognition.
\newblock In \emph{WACV}, 2025.

\bibitem[Zheng et~al.(2016)Zheng, Zhao, Wang, Wang, and Tian]{zheng2016good}
Liang Zheng, Yali Zhao, Shengjin Wang, Jingdong Wang, and Qi Tian.
\newblock Good practice in cnn feature transfer.
\newblock \emph{arXiv:1604.00133}, 2016.

\bibitem[Zhou et~al.(2017)Zhou, Zhao, Puig, Fidler, Barriuso, and Torralba]{ade20k}
Bolei Zhou, Hang Zhao, Xavier Puig, Sanja Fidler, Adela Barriuso, and Antonio Torralba.
\newblock Scene parsing through {ADE20K} dataset.
\newblock In \emph{CVPR}, 2017.

\bibitem[Zhu et~al.(2025)Zhu, Wang, Chen, Liu, Ye, Gu, Duan, Tian, Su, Shao, Gao, Cui, Cao, Liu, Xu, Li, Wang, Lv, Chen, Li, He, Jiang, Luo, Wang, He, Shi, Zhang, Shao, He, Xiong, Qu, Sun, Jiao, Wu, Zhang, Deng, Ge, Chen, Wang, Dou, Lu, Zhu, Lu, Lin, Qiao, Dai, and Wang]{internvl3}
Jinguo Zhu, Weiyun Wang, Zhe Chen, Zhaoyang Liu, Shenglong Ye, Lixin Gu, Yuchen Duan, Hao Tian, Weijie Su, Jie Shao, Zhangwei Gao, Erfei Cui, Yue Cao, Yangzhou Liu, Weiye Xu, Hao Li, Jiahao Wang, Han Lv, Dengnian Chen, Songze Li, Yinan He, Tan Jiang, Jiapeng Luo, Yi Wang, Conghui He, Botian Shi, Xingcheng Zhang, Wenqi Shao, Junjun He, Yingtong Xiong, Wenwen Qu, Peng Sun, Penglong Jiao, Lijun Wu, Kaipeng Zhang, Huipeng Deng, Jiaye Ge, Kai Chen, Limin Wang, Min Dou, Lewei Lu, Xizhou Zhu, Tong Lu, Dahua Lin, Yu Qiao, Jifeng Dai, and Wenhai Wang.
\newblock {InternVL3}: Exploring advanced training and test-time recipes for open-source multimodal models.
\newblock \emph{arxiv:2504.10479}, 2025.

\bibitem[Zong et~al.(2023)Zong, Song, and Liu]{codetr}
Zhuofan Zong, Guanglu Song, and Yu Liu.
\newblock {DETRs} with collaborative hybrid assignments training.
\newblock In \emph{ICCV}, 2023.

\end{thebibliography}
